\documentclass[fleqn,12pt]{wlscirep}
\usepackage{tikz}
\usepackage{subfigure}
\usepackage{xcolor}
\usepackage{colortbl}
\usepackage{pgf}
\usepackage{pgfplots}
\usepackage[utf8]{inputenc}
\usepackage[T1]{fontenc}
\usepackage{textalpha}
\usepackage{graphicx}
\usepackage{multirow}
\usepackage{makecell}
\usepackage{booktabs}
\usepackage{amsfonts}
\usepackage{enumitem}
\usepackage{longtable}
\usepackage{tabularx}
\setlist[enumerate,itemize]{topsep=0pt, itemsep=0pt, leftmargin=*, after=\leavevmode}
\setlist[enumerate,1]{label=(\arabic*)}
\usepackage{subcaption}
\usepackage{hyperref}
\hypersetup{
    colorlinks=true,
    linkcolor=darkblue,   
    citecolor=darkgreen,  
    urlcolor=darkblue     
}
\usepackage{listings}
\usepackage{fancyvrb}
\usepackage{caption}
\usepackage{float}
\usepackage{indentfirst}
\usepackage[most]{tcolorbox}

\definecolor{RoseRed}{RGB}{245, 102, 98}
\definecolor{MyBlue}{RGB}{125, 164, 247}
\definecolor{MyGreen}{RGB}{152, 211, 90}
\definecolor{MyOrange}{RGB}{244, 177, 131}
\definecolor{LightBlue}{RGB}{173, 216, 230}

\definecolor{darkgreen}{rgb}{0,0.4,0}
\definecolor{darkblue}{rgb}{0,0.2,0.6}
\definecolor{chocolate}{HTML}{D2691E}
\definecolor{maroon}{HTML}{A00000}
\definecolor{indigo}{HTML}{4B0082}
\definecolor{violet}{HTML}{4B2E83}
\definecolor{lightblue}{rgb}{0.0, 0.0, 0.5}
\definecolor{cadmiumgreen}{rgb}{0.0, 0.42, 0.24}
\definecolor{forestgreen}{rgb}{0.13, 0.55, 0.13}
\definecolor{lightbluebg}{RGB}{235, 245, 255}  
\definecolor{blueframe}{RGB}{70, 130, 180}     
\usepackage[most]{tcolorbox}
\usepackage{fvextra}
\DefineVerbatimEnvironment{VerbatimWrap}{Verbatim}{
breaklines=true, 
breakindent=0pt, 
breaksymbol={},
fontsize=\scriptsize,
}

\newcommand{\textbothscript}[4]{#1 \ensuremath{^{\textcolor{gray}{\scalebox{.5}{±#2}}}_{\textcolor{gray}{\scalebox{.5}{[#3,\ #4]}}}}}

\newcommand{\std}[1]{\textcolor{gray}{\scalebox{.8}{$\pm$#1}}}

\definecolor{heatmapColor1}{RGB}{237,248,233}  
\definecolor{heatmapColor2}{RGB}{204,238,199}  
\definecolor{heatmapColor3}{RGB}{166,228,169}  
\definecolor{heatmapColor4}{RGB}{121,218,142}  
\definecolor{heatmapColor5}{RGB}{51,207,118} 

\newcommand{\genericHeatmapCell}[3]{%
  \pgfmathsetmacro{\valueIn}{#1}
  \pgfmathsetmacro{\minVal}{#2}%
  \pgfmathsetmacro{\maxVal}{#3}%
  \pgfmathsetmacro{\range}{\maxVal - \minVal}%
  \ifdim\range pt=0pt
      \def\normalizedvalue{50}
  \else
      \pgfmathsetmacro{\normalizedvalue}{min(100, max(0, (\valueIn - \minVal) / \range * 100))}%
  \fi

  \ifdim\normalizedvalue pt<25pt 
    \pgfmathparse{round(\normalizedvalue * 4)}\pgfmathtruncatemacro{\colorpercent}{\pgfmathresult}
    \edef\tempcellcolorstring{heatmapColor2!\colorpercent!heatmapColor1}%
  \else
    \ifdim\normalizedvalue pt<50pt 
      \pgfmathparse{round((\normalizedvalue - 25) * 4)}\pgfmathtruncatemacro{\colorpercent}{\pgfmathresult}
      \edef\tempcellcolorstring{heatmapColor3!\colorpercent!heatmapColor2}%
    \else
      \ifdim\normalizedvalue pt<75pt 
        \pgfmathparse{round((\normalizedvalue - 50) * 4)}\pgfmathtruncatemacro{\colorpercent}{\pgfmathresult}
        \edef\tempcellcolorstring{heatmapColor4!\colorpercent!heatmapColor3}%
      \else 
        \pgfmathparse{round((\normalizedvalue - 75) * 4)}\pgfmathtruncatemacro{\colorpercent}{\pgfmathresult}
        \edef\tempcellcolorstring{heatmapColor5!\colorpercent!heatmapColor4}%
      \fi
    \fi
  \fi

  \expandafter\cellcolor\expandafter{\tempcellcolorstring}%
  {#1}%
}

\newcommand{\reverseHeatmapCell}[3]{%
  \pgfmathsetmacro{\valueIn}{#1}
  \pgfmathsetmacro{\minVal}{#2}%
  \pgfmathsetmacro{\maxVal}{#3}%
  \pgfmathsetmacro{\range}{\maxVal - \minVal}%
  \ifdim\range pt=0pt
      \def\normalizedvalue{50}
  \else
      \pgfmathsetmacro{\normalizedvalue}{min(100, max(0, (\valueIn - \minVal) / \range * 100))}%
  \fi

  \ifdim\normalizedvalue pt<25pt 
    \pgfmathparse{round(\normalizedvalue * 4)}\pgfmathtruncatemacro{\colorpercent}{\pgfmathresult}
    \edef\tempcellcolorstring{heatmapColor4!\colorpercent!heatmapColor5}%
  \else
    \ifdim\normalizedvalue pt<50pt 
      \pgfmathparse{round((\normalizedvalue - 25) * 4)}\pgfmathtruncatemacro{\colorpercent}{\pgfmathresult}
      \edef\tempcellcolorstring{heatmapColor3!\colorpercent!heatmapColor4}%
    \else
      \ifdim\normalizedvalue pt<75pt 
        \pgfmathparse{round((\normalizedvalue - 50) * 4)}\pgfmathtruncatemacro{\colorpercent}{\pgfmathresult}
        \edef\tempcellcolorstring{heatmapColor2!\colorpercent!heatmapColor3}%
      \else 
        \pgfmathparse{round((\normalizedvalue - 75) * 4)}\pgfmathtruncatemacro{\colorpercent}{\pgfmathresult}
        \edef\tempcellcolorstring{heatmapColor1!\colorpercent!heatmapColor2}%
      \fi
    \fi
  \fi

  \expandafter\cellcolor\expandafter{\tempcellcolorstring}%
  {#1}%
}

\newcommand{\colortjhroc}[1]{\genericHeatmapCell{#1}{49}{100}} 
\newcommand{\colortjhprc}[1]{\genericHeatmapCell{#1}{46}{100}} 
\newcommand{\colortjhmae}[1]{\reverseHeatmapCell{#1}{2}{18}}
\newcommand{\colortjhmse}[1]{\reverseHeatmapCell{#1}{16}{380}} 
\newcommand{\colortjhrmse}[1]{\reverseHeatmapCell{#1}{3.9}{20}}

\newcommand{\colormimicoutroc}[1]{\genericHeatmapCell{#1}{50}{100}} 
\newcommand{\colormimicoutprc}[1]{\genericHeatmapCell{#1}{10}{100}} 

\newcommand{\colormimicreadroc}[1]{\genericHeatmapCell{#1}{50}{83}} 
\newcommand{\colormimicreadprc}[1]{\genericHeatmapCell{#1}{24}{70}} 

\newcommand{\coloroutroc}[1]{\genericHeatmapCell{#1}{56}{100}} 
\newcommand{\coloroutprc}[1]{\genericHeatmapCell{#1}{10}{75}} 
\newcommand{\colorreadroc}[1]{\genericHeatmapCell{#1}{50}{100}} 
\newcommand{\colorreadprc}[1]{\genericHeatmapCell{#1}{30}{80}} 

\newcommand{\colormmoutroc}[1]{\genericHeatmapCell{#1}{45}{100}}
\newcommand{\colormmoutprc}[1]{\genericHeatmapCell{#1}{5}{85}}
\newcommand{\colormmreadroc}[1]{\genericHeatmapCell{#1}{50}{85}}
\newcommand{\colormmreadprc}[1]{\genericHeatmapCell{#1}{20}{80}}

\title{ClinicRealm: Re-evaluating Large Language Models with Conventional Machine Learning for Non-Generative Clinical Prediction Tasks}
\author[5,7,+]{Yinghao Zhu}
\author[2,3,+]{Junyi Gao}
\author[5,+]{Zixiang Wang}
\author[4,5,+]{Weibin Liao}
\author[6]{Xiaochen Zheng}
\author[5]{Lifang Liang}
\author[2]{Miguel O. Bernabeu}
\author[5]{Yasha Wang}
\author[7]{Lequan Yu}
\author[1,*]{Chengwei Pan}
\author[2,*]{Ewen M. Harrison}
\author[5,*]{Liantao Ma}
\affil[1]{School of Artificial Intelligence, Beihang University, Beijing, China, 100191}
\affil[2]{Centre for Medical Informatics, The University of Edinburgh, Edinburgh, UK, EH8 9YL}
\affil[3]{Health Data Research UK, UK}
\affil[4]{School of Computer Science, Peking University, Beijing, China, 100871}
\affil[5]{National Engineering Research Center for Software Engineering, Peking University, Beijing, China, 100871}
\affil[6]{ETH Zurich, Zurich, Switzerland, 8092}
\affil[7]{School of Computing and Data Science, The University of Hong Kong, Hong Kong SAR, China, 999077}
\affil[*]{Correspond to \texttt{pancw@buaa.edu.cn}, \texttt{ewen.harrison@ed.ac.uk}, \texttt{malt@pku.edu.cn}}
\affil[+]{These authors contributed equally to this work}

\begin{abstract}
Large Language Models (LLMs) are increasingly deployed in medicine. However, their utility in non-generative clinical prediction, often presumed inferior to specialized models, remains under-evaluated, leading to ongoing debate within the field and potential for misuse, misunderstanding, or over-reliance due to a lack of systematic benchmarking. Our ClinicRealm study addresses this by benchmarking 15 GPT-style LLMs, 5 BERT-style models, and 11 traditional methods on unstructured clinical notes and structured Electronic Health Records (EHR), while also assessing their reasoning, reliability, and fairness. Key findings reveal a significant shift: for clinical note predictions, leading LLMs (e.g., DeepSeek-V3.1-Think, GPT-5) in zero-shot settings now decisively outperform finetuned BERT models. On structured EHRs, while specialized models excel with ample data, advanced LLMs (e.g., GPT-5, DeepSeek-V3.1-Think) show potent zero-shot capabilities, often surpassing conventional models in data-scarce settings. Notably, leading open-source LLMs can match or exceed proprietary counterparts. These results provide compelling evidence that modern LLMs are competitive tools for non-generative clinical prediction, particularly with unstructured text and offering data-efficient structured data options, thus necessitating a re-evaluation of model selection strategies. This research should serve as an important insight for medical informaticists, AI developers, and clinical researchers, potentially prompting a reassessment of current assumptions and inspiring new approaches to LLM application in predictive healthcare.
\end{abstract}

\begin{document}

\flushbottom
\maketitle

\section{Introduction}

In modern healthcare, the ability to accurately predict and understand patient outcomes using vast amounts of healthcare data is critical. A spectrum of data, from unstructured clinical notes to structured Electronic Health Records (EHR), presents a complex landscape for analysis. Clinicians increasingly utilize machine learning (ML) and deep learning (DL) models for informed decision-making, leveraging models trained specifically for tasks like mortality and hospital readmission prediction~\cite{harutyunyan2019multitask,ma2020concare,purushotham2018benchmarking}. More recently, the advent of generative artificial intelligence (GenAI), particularly Generative Pre-Trained Transformer-based Large Language Models (GPT-style LLMs), has shown remarkable capabilities in various generative tasks, including patient interaction~\cite{mesko2023impact}, resolving complex medical queries~\cite{qian2024liver}, and passing medical licensing exams~\cite{gilson2023does}, sparking interest in their broader clinical potential~\cite{chen2025benchmarking}.

\begin{figure}[!ht]
    \centering
    \includegraphics[width=1.0\linewidth]{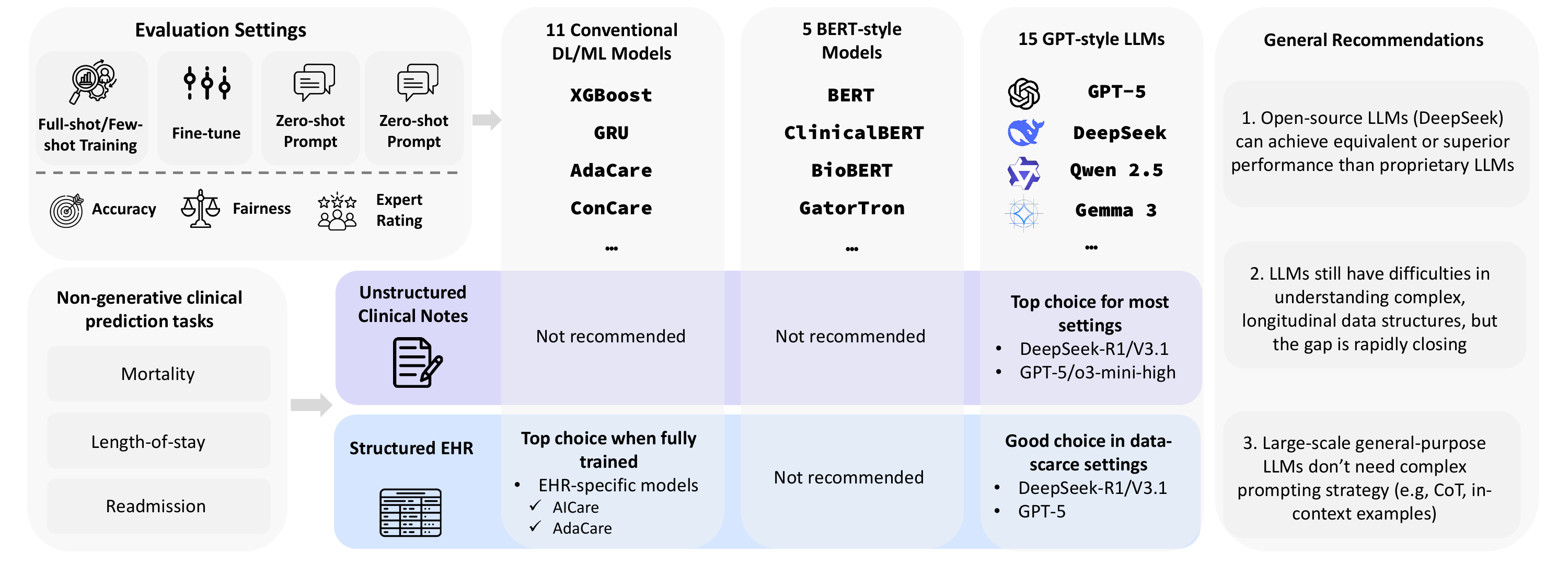}
    \caption{\textit{Comparative performance and recommendations for model selection in non-generative clinical tasks.} This figure summarizes the benchmarking results, offering guidance on selecting optimal models for different clinical scenarios. It compares conventional DL/ML, BERT-style, and Large Language Models (LLMs) across two categories of tasks: (i) prediction tasks using unstructured clinical notes data (e.g., mortality, readmission prediction), and (ii) prediction tasks using structured Electronic Health Record (EHR) data (e.g., mortality, length-of-stay prediction).}
    \label{fig:framework}
\end{figure}

While LLMs have excelled in generative tasks, their effectiveness in non-generative clinical prediction tasks such as risk probability assessment, crucial for direct clinical decision support-has been less clear and often underestimated. \textbf{Indeed, a prevailing assumption has been that LLMs generally underperform these specialized, locally trained models in such clinical prediction scenarios~\cite{chen2024clinicalbench,lehman2023we,brown2025large}. However, our comprehensive benchmarking of state-of-the-art LLMs, including reasoning-enhanced and open-source models, challenges this view and reveals a rapidly evolving landscape.} Our findings highlight several key aspects of this evolution:

\begin{itemize}
    \item For unstructured clinical notes, where locally finetuned BERT variants were widely considered optimal for prediction~\cite{lehman2023we,brown2025large,bucher2024fine}, our results indicate a significant shift. Recent zero-shot LLMs (e.g., GPT-5, DeepSeek-R1/V3.1) now substantially outperform these specialized models, suggesting that the extensive efforts for BERT finetuning in this domain may be less critical than previously thought.
    \item Regarding structured EHR data, conventional ML/DL models perform strongly when trained with ample data~\cite{purushotham2018benchmarking,chen2024clinicalbench}. In contrast, advanced LLMs (e.g., GPT-5, DeepSeek-R1/V3.1) show impressive zero-shot capabilities; even when compared to simpler models trained on full datasets, their performance can be within a 10\% margin. In data-scarce scenarios, these LLMs can surpass most conventional models. This makes them particularly promising for applications with limited data, such as in the context of emerging diseases or rare conditions.
    \item Furthermore, the performance of medically domain-specific finetuned LLMs in these predictive tasks warrants careful consideration. While often optimized and finetuned for question-answering, our experiments show they may not offer an advantage over general-purpose LLMs for non-generative, discriminative clinical predictions. Notably, leading open-source models (e.g., DeepSeek) demonstrate performance in these tasks that is comparable, and at times superior, to proprietary counterparts, further broadening the accessibility of high-performing models in data-sensitive settings.
\end{itemize}
The evidence for these observations is drawn from our ClinicRealm benchmark, which evaluates 15 state-of-the-art Large Language Models (LLMs), including the latest high-performing open-source and proprietary releases, alongside 5 BERT-style language models representing previously leading text encoders. These are further benchmarked against 11 established conventional machine learning/deep learning models, including 4 specialized predictive models designed for EHR data, ensuring a comprehensive assessment across the spectrum of current and foundational methodologies. Furthermore, we extend our analysis to multimodal settings, evaluating models that integrate both structured EHR data and unstructured clinical notes. This study aims to rectify existing potential misconceptions and provide clarity across representative non-generative medical prediction tasks (Figure~\ref{fig:framework}). By offering actionable insights into optimal model selection for various clinical data types and tasks-from structured EHR to clinical notes, we identify key research gaps for future clinical LLM development. Ultimately, this work seeks to enhance the quality of patient care by facilitating more accurate and appropriately chosen predictive analytics. We provide the up-to-date benchmark results online (\url{https://yhzhu99.github.io/ehr-llm-benchmark/}).

\section{Methods}

\subsection{Data Sources and Processing}

Our study utilizes three publicly available datasets: MIMIC-IV~\cite{mimic4}, MIMIC-III~\cite{mimic3}, and a dataset from Tongji Hospital (TJH)\cite{tjh}. For MIMIC-III and MIMIC-IV, each unique hospital admission (identified by subject and admission IDs) was treated as a distinct sample. To ensure full reproducibility, all preprocessing scripts are available at our GitHub repository.

\begin{itemize}
\item \textit{Clinical Notes (MIMIC-IV-Note v2.2 \& MIMIC-III v1.4):} From MIMIC-IV, we used discharge summaries, which inherently summarize the entire hospitalization, for retrospective mortality and prospective readmission prediction. To enable a truly prospective mortality prediction task, we incorporated MIMIC-III and extracted all ``Physician'', ``Nursing'', and ``Nursing/other'' notes recorded within the first 24 hours of admission, concatenating them chronologically.
\item \textit{Structured EHR (MIMIC-IV v3.1 \& TJH):} For MIMIC-IV, we used a cohort of adult patient admissions, extracting a time series of 17 physiologic variables as defined in the established benchmark pipeline\cite{harutyunyan2019multitask}. The TJH dataset comprises data from 485 COVID-19 patients, including 73 lab tests and vital signs along with demographic features. For both datasets, to handle sparsity, we aggregated measurements on a daily basis. For MIMIC-IV admissions longer than seven days, we retained daily records from the last seven days and aggregated all prior data into a single initial time step (using the last recorded value for each feature), resulting in a sequence of at most 8 time steps.
\end{itemize}

For conventional ML/DL models, any remaining missing values in structured EHR data were imputed using the Last Observation Carried Forward (LOCF) method~\cite{wells2013LOCF_Imputation}. For LLM-based approaches, we retained missing values as ``NaN'' to provide raw, unaltered data. Structured data were standardized using z-score normalization, and outliers (absolute z-score > 10,000) were removed~\cite{gao2024comprehensive,zhu2023pyehr}. For all clinical notes, we followed the preprocessing steps from Clinical-Longformer~\cite{li2023ClinicalLongformer}, including removing de-identification tags, normalizing text, and converting to lowercase. To prevent label leakage, we additionally removed explicit mentions of outcomes (e.g., ``expired'', ``deceased'', ``readmitted''). To address the critical issue of potential data contamination, where models (particularly proprietary LLMs with opaque training corpora) might have been inadvertently trained on benchmark data, we provide a detailed timeline analysis in Appendix~\ref{sec:appendix_contamination}. Dataset statistics are detailed in Table~\ref{tab:dataset_stats}.

\subsection{Benchmarking Task Formulation}

Following previous clinical prediction benchmarking studies~\cite{harutyunyan2019multitask,gao2024comprehensive}, we chose three widely recognized clinical tasks to assess model performance. To ensure reproducibility and clarity, we provide precise specifications for each task, including the observation period, prediction timepoint, and whether the task is prospective (predicting a future event) or retrospective (classifying an event that has already occurred based on a complete record). A summary is provided in Table~\ref{tab:data_task_summary}, with detailed definitions below.

\begin{table}[!ht]
\centering
\caption{\textit{Summary of task specifications, datasets, and timepoints.} This table provides a clear and rigorous definition for each prediction task, specifying the data used, the observation window, the precise timepoint at which a prediction is made, and the prediction target.}
\label{tab:data_task_summary}
\resizebox{\textwidth}{!}{%
\begin{tabular}{@{}llllll@{}}
\toprule
\textbf{Task} & \textbf{Dataset} & \textbf{Data Modality} & \textbf{Observation Window} & \textbf{Prediction Timepoint} & \textbf{Prediction Target} \\ \midrule
\multirow{3}{*}{In-hospital Mortality} & MIMIC-IV / TJH & Structured EHR & In-hospital since admission & At discharge (Retrospective) & Mortality during the hospital stay \\
 & MIMIC-IV & Discharge Notes & At discharge & At discharge (Retrospective) & Mortality during the hospital stay \\
 & MIMIC-III & Admission Notes & First 24 hours of admission & Within 24 hours of admission (Prospective) & Mortality during the hospital stay \\ \midrule
\multirow{2}{*}{30-day Readmission} & MIMIC-IV & Structured EHR & In-hospital since admission & At discharge (Prospective) & Readmission within 30 days post-discharge \\
 & MIMIC-IV & Discharge Notes & At discharge & At discharge (Prospective) & Readmission within 30 days post-discharge \\ \midrule
Length-of-Stay & TJH & Structured EHR & In-hospital since admission & At each patient visit (Prospective) & Remaining duration of hospital stay (in days) \\ \bottomrule
\end{tabular}%
}
\end{table}

\begin{itemize}
\item \textbf{In-hospital mortality prediction:} A binary classification task to predict whether a patient will die during their hospital stay. The label is 1 if a patient's recorded time of death is between their admission and discharge times; otherwise, it is 0. This task is evaluated under three distinct settings:
\begin{itemize}
\item \textit{Retrospective (Structured EHR):} Using structured EHR data from the entire hospital stay on both MIMIC-IV and TJH datasets, with the prediction made at the time of discharge.
\item \textit{Retrospective (Discharge Notes):} Using the complete discharge summary from MIMIC-IV, which summarizes the entire hospitalization, with the prediction made at the time of discharge.
\item \textit{Prospective (Admission Notes):} Using clinical notes recorded within the first 24 hours of admission from the MIMIC-III dataset to predict future mortality, making this a true prospective task.
\end{itemize}
\item \textbf{30-day readmission prediction:} A prospective binary classification task to predict whether a patient will be readmitted within 30 days of discharge. The label is 1 if a subsequent admission occurs within this window. Following established benchmarks, if a patient dies within 30 days post-discharge, this is also counted as a positive readmission event. The prediction is made at the time of discharge using either structured EHR data or discharge notes from MIMIC-IV. This task was evaluated exclusively on the MIMIC-IV dataset, as the TJH dataset was specifically curated for an in-hospital mortality prediction study and does not contain the necessary post-discharge follow-up data required for a readmission analysis~\cite{tjh}.
\item \textbf{Length-of-stay (LOS) prediction:} A prospective regression task to predict the remaining duration of a patient's hospital stay in days. The prediction is made dynamically at each patient visit using cumulative structured EHR data from the TJH dataset, allowing for early intervention.
\end{itemize}

\subsection{Benchmarking Model Selection}

\begin{table}[!ht]
    \footnotesize
    \centering
    \caption{\textit{Baseline models compared in this benchmark, including conventional ML/DL models, BERT-style models, Base LLMs, and Reasoning LLMs.} Conventional ML/DL models focus on clinical predictive tasks utilizing structured EHR data. BERT-style models primarily process unstructured clinical notes. LLMs, being general-purpose, are suitable for evaluation across both structured EHR data and unstructured clinical notes. Non-generative tasks are highlighted in \textcolor{blue}{\textbf{Blue}}, while generative tasks are marked in \textcolor{red}{\textbf{Red}}. In the case of GPT-style models, tasks like document classification, though inherently classification tasks, are conducted in a generative manner through the use of prompts.}
    \label{tab:baseline_models_with_tasks}
\resizebox{1.0\textwidth}{!}{
    \begin{tabular}{cccccccc}
    \toprule
         \multicolumn{2}{c}{\textbf{Methods}} & \textbf{Release Time} & \textbf{Parameters} & \textbf{Embedding Size} & \textbf{Context Window} & \textbf{Pretraining Data} & \textbf{Evaluation Tasks} \\
        \midrule
        \multirow{14}{*}{Conventional ML/DL} 
         & Decision Tree & 1960s & - & - & - & - & - \\ \cmidrule{2-8}
         & Random Forest~\cite{breiman2001randomforest} & 2001-01 & - & - & - & - & - \\ \cmidrule{2-8}
         & XGBoost~\cite{chen2016xgboost} & 2016-03 & - & - & - & - & - \\
         \cmidrule{2-8}
         & CatBoost~\cite{prokhorenkova2018catboost} & 2018-12 & - & - & - & - & - \\ \cmidrule{2-8}
         & RNN~\cite{medsker2001recurrent} & 1980s & 22.3K & 128 & - & - & - \\ \cmidrule{2-8}
         & LSTM~\cite{hochreiter1997lstm} & 1997-11 & 89.1K & 128 & - & - & - \\
         \cmidrule{2-8}
         & GRU~\cite{chung2014empirical} & 2014-12 & 88.4K & 128 & - & - & - \\
         \cmidrule{2-8}
         & AdaCare~\cite{ma2020adacare} & 2019-11 & 202K & 128 & - & - & \makecell{\textcolor{blue}{Mortality prediction} \\ \textcolor{blue}{Decompensation Prediction}} \\
         \cmidrule{2-8}
         & ConCare~\cite{ma2020concare} & 2019-11 & 3.3M & 128 & - & - & \makecell{\textcolor{blue}{Mortality prediction} \\ \textcolor{blue}{Length-of-stay prediction}} \\
         \cmidrule{2-8}
         & GRASP~\cite{zhang2021grasp} & 2021-02 & 3.3M & 128 & - & - & \makecell{\textcolor{blue}{Mortality prediction} \\ \textcolor{blue}{Sepsis prediction}} \\
         \cmidrule{2-8}
         & AICare~\cite{ma2023aicare} & 2023-12 & 2.2M & 128 & - & - & \textcolor{blue}{Mortality prediction} \\
        \midrule
        \multirow{17}{*}{BERT-style} 
         & BERT~\cite{devlin2018bert} & 2018-10 & 110M & 768 & 512 & \makecell{BooksCorpus \\ English Wikipedia} & \makecell{\textcolor{blue}{Sentence pair classification} \\ \textcolor{blue}{Single sentence classification} \\ \textcolor{blue}{Named entity recognition} \\ \textcolor{red}{Question answering}} \\
         \cmidrule{2-8}
         & ClinicalBERT~\cite{huang2019clinicalbert} & 2019-04 & 110M & 768 & 512 & MIMIC-III clinical notes & \makecell{\textcolor{blue}{Clinical NLP inference} \\ \textcolor{blue}{Named entity recognition}}\\
         \cmidrule{2-8}
         & BioBERT~\cite{lee2020biobert} & 2019-09 & 110M & 768 & 512 & PubMed & \makecell{\textcolor{blue}{Named entity recognition} \\ \textcolor{blue}{Relation extraction} \\ \textcolor{red}{Question answering}} \\
         \cmidrule{2-8}
         & GatorTron~\cite{yang2022gatortron} & 2022-02 & 345M & 1024 & 512 & \makecell{UF Health IDR \\ MIMIC-III clinical notes \\ PubMed \\ Wikipedia} & \makecell{\textcolor{blue}{Clinical NLP inference} \\ \textcolor{blue}{Clinical concept extraction} \\ \textcolor{blue}{Relation extraction} \\ \textcolor{blue}{Semantic similarity} \\ \textcolor{red}{Question answering}}\\
         \cmidrule{2-8}
         & Clinical-Longformer~\cite{li2023ClinicalLongformer} & 2023-01 & 102M & 768 & 4096 & MIMIC-III clinical notes & \makecell{\textcolor{blue}{Named entity recognition} \\ \textcolor{blue}{Document classification} \\ \textcolor{red}{Question answering}} \\
         \midrule
         \multirow{23}{*}{Base LLM}
         & GPT-2~\cite{radford2019gpt2} & 2019-02 & 117M & 768 & 1024 & Web text & \makecell{\textcolor{red}{Question answering} \\ \textcolor{red}{Translation} \\ \textcolor{red}{Summarization}} \\
         \cmidrule{2-8}
         & BioGPT~\cite{luo2022biogpt} & 2022-10 & 347M & 1024 & 1024 & PubMed & \makecell{\textcolor{blue}{Document classification} \\ \textcolor{blue}{Relation extraction} \\ \textcolor{red}{Question answering}}\\
         \cmidrule{2-8}
         & Meditron~\cite{chen2023meditron} & 2023-09 & 7B & 4096 & 2048 & \makecell{Clinical guidelines \\ PubMed abstracts \\ PubMed papers \\ Experience replies} & \makecell{\textcolor{red}{Medical multiple-choice questions} \\ \textcolor{red}{Question answering}} \\
         \cmidrule{2-8}
         & BioMistral~\cite{chen2023meditron} & 2024-02 & 7B & 4096 & 2048 & \makecell{BMC Papers} & \makecell{\textcolor{red}{Medical multiple-choice questions} \\ \textcolor{red}{Question answering}} \\
         \cmidrule{2-8}
         & OpenBioLLM~\cite{OpenBioLLMs} & 2024-05 & 8B & 4096 & 8192 & \makecell{9 biomedical datasets \\ Medical instruct dataset} & \makecell{\textcolor{blue}{Clinical entity recognition} \\ \textcolor{red}{Question answering} \\ \textcolor{red}{Summarization}} \\
         \cmidrule{2-8}
         & Qwen2.5~\cite{yang2024qwen25} & 2024-12 & 7B & 3584 & 128K & \makecell{Web text \\ Human feedbacks} & \makecell{\textcolor{red}{Question answering} \\ \textcolor{red}{Text generation}} \\
         \cmidrule{2-8}
         & Gemma-3~\cite{team2025gemma} & 2025-03 & 4B & 2560 & 128K & \makecell{Web text and images \\ Human feedbacks} & \makecell{\textcolor{red}{Question answering} \\ \textcolor{red}{Text generation}} \\
         \cmidrule{2-8}
         \cmidrule{2-8}
         & DeepSeek-V3.1~\cite{liu2024deepseekv3,deepseek2025v31} & 2025-08 & 671B & 7168 & 128K & \makecell{Web text \\ Human feedbacks} & \makecell{\textcolor{red}{Question answering} \\ \textcolor{red}{Text generation}} \\
         \cmidrule{2-8}
         & GPT-4o~\cite{hurst2024gpt} & 2024-05 & Unknown & Unknown & 128K & \makecell{-} & \makecell{\textcolor{red}{Question answering} \\
         \textcolor{red}{Text generation}} \\
         \midrule
         \multirow{10}{*}{Reasoning LLM} & HuatuoGPT-o1~\cite{chen2024huatuogpt} & 2024-12 & 7B & 3584 & 128K & \makecell{Medical exam questions} & \makecell{\textcolor{red}{Medical multiple-choice questions} \\ \textcolor{red}{Question answering}} \\
         \cmidrule{2-8}
         & DeepSeek-R1-Distill-Qwen~\cite{guo2025deepseekr1} & 2025-01 & 7B & 3584 & 128K & \makecell{Web text \\ Human feedbacks} & \makecell{\textcolor{red}{Question answering} \\ \textcolor{red}{Text generation}} \\
         \cmidrule{2-8}
         & DeepSeek-R1~\cite{guo2025deepseekr1} & 2025-01 & 671B & 7168 & 128K & \makecell{Web text \\ Human feedbacks} & \makecell{\textcolor{red}{Question answering} \\ \textcolor{red}{Text generation}} \\
         \cmidrule{2-8}
         & DeepSeek-V3.1-Think~\cite{liu2024deepseekv3,deepseek2025v31} & 2025-08 & 671B & 7168 & 128K & \makecell{Web text \\ Human feedbacks} & \makecell{\textcolor{red}{Question answering} \\ \textcolor{red}{Text generation}} \\
         \cmidrule{2-8}
         & GPT o3-mini-high~\cite{openai2025o3mini} & 2025-01 & Unknown & Unknown & 200K & \makecell{-} & \makecell{\textcolor{red}{Question answering} \\ \textcolor{red}{Text generation}} \\
         \cmidrule{2-8}
         & GPT-5~\cite{openai2025gpt5} & 2025-08 & Unknown & Unknown & 400K & \makecell{-} & \makecell{\textcolor{red}{Question answering} \\ \textcolor{red}{Text generation}} \\
    \bottomrule
    \end{tabular}
}
\end{table}

We examined a diverse array of state-of-the-art models to ensure generalizable and comprehensive conclusions, spanning machine learning and deep learning-based clinical prediction models, BERT-style language models and GPT-style LLMs, including base LLMs and advanced reasoning LLMs:

\begin{itemize}
\item \textbf{Conventional ML/DL models for clinical predictions} included four conventional machine learning models (CatBoost~\cite{prokhorenkova2018catboost}, Decision tree, Random Forest~\cite{breiman2001randomforest}, XGBoost~\cite{chen2016xgboost}) and three deep learning models (GRU~\cite{chung2014empirical}, LSTM~\cite{hochreiter1997lstm}, RNN~\cite{medsker2001recurrent}), as well as advanced predictive models designed for longitudinal EHR data (AdaCare~\cite{ma2020adacare}, ConCare~\cite{ma2020concare}, GRASP~\cite{zhang2021grasp}, AICare~\cite{ma2023aicare}).
\item \textbf{BERT-style models} included BERT models pretrained on general texts (BERT~\cite{devlin2018bert}) and the biomedical corpus (ClinicalBERT~\cite{huang2019clinicalbert}, BioBERT~\cite{lee2020biobert}, GatorTron~\cite{yang2022gatortron}, Clinical-Longformer~\cite{li2023ClinicalLongformer}).
\item \textbf{Large language models} included several general purpose LLMs (GPT-2~\cite{radford2019gpt2}, GPT-4o~\cite{hurst2024gpt}, Gemma-3\cite{team2025gemma}, Qwen2.5~\cite{yang2024qwen25}, and DeepSeek-V3.1~\cite{liu2024deepseekv3,deepseek2025v31}), along with medically finetuned variants (BioGPT~\cite{luo2022biogpt}, Meditron~\cite{chen2023meditron}, OpenBioLLM~\cite{OpenBioLLMs}, and BioMistral~\cite{labrak2024biomistral}). We also include advanced reasoning LLMs, such as HuatuoGPT-o1-7B~\cite{chen2024huatuogpt}, DeepSeek-V3.1-Think~\cite{liu2024deepseekv3,deepseek2025v31}, DeepSeek-R1 (version 250528)\cite{guo2025deepseekr1}, GPT-5 (High)\cite{openai2025gpt5}, DeepSeek-R1-Distill-Qwen~\cite{guo2025deepseekr1}, and GPT o3-mini-high\cite{openai2025o3mini}).
\end{itemize}

We summarized the details of all baseline models in Table~\ref{tab:baseline_models_with_tasks}. Throughout the experiments, we strictly adhered to the data use agreement. The performance of OpenAI models on all datasets was processed using the secure Azure OpenAI API, with human review of the data waived. Additionally, all other models, including ML, DL, and other LLMs, were deployed locally. The code for this benchmarking work, including all preprocessing scripts and prompt templates, can be accessed online (\url{https://github.com/yhzhu99/ehr-llm-benchmark}). 

\noindent\textbf{Optimized prompt strategy for LLMs with structured EHR data:} The decoder-only architecture of LLMs, trained primarily on unstructured natural language texts, encounters challenges with the structured, sparse, and longitudinal nature of EHR data~\cite{sui2024table,jiang2023structgpt}. We utilized a prompting strategy to better deliver structured EHR data to LLMs. The prompting strategy employed a feature-wise list-style format for inputting EHR data and provided LLMs with feature units and reference ranges. We manually looked up the unit and reference ranges for each clinical feature in the medical guidelines. We also explored the in-context learning strategy in the prompt by providing example cases. The prompt templates for the prediction tasks with EHR data are shown in Table~\ref{tab:task_description_and_response}, \ref{tab:tjh_optimized_icl_full_ehr_prompt}, \ref{tab:mimic4_optimized_icl_full_ehr_prompt}.

\subsection{Multimodal Integration Experiments}
Recognizing that a holistic patient view often requires synthesizing information from multiple sources, we designed a set of experiments to evaluate model performance on multimodal data. We combined structured EHR time-series data with unstructured discharge summaries from the MIMIC-IV dataset for both the in-hospital mortality and 30-day readmission prediction tasks. We assessed two primary integration strategies: a prompt-based approach for zero-shot LLMs and a feature-fusion approach for finetuned models.

\paragraph{Prompt-based multimodal integration.}
For LLMs, we developed a unified prompt that presents both structured EHR data and the clinical note in a single context, testing the models' inherent ability to synthesize information from disparate data types. The structured data was formatted as a time-series list, followed by the full text of the clinical note. The specific templates are provided in the Appendix (Tables \ref{tab:mimic4_multimodal_mortality_prompt} and \ref{tab:mimic4_multimodal_readmission_prompt}).

\paragraph{Feature-fusion for finetuned models.}
For finetuned models, we adopted a feature-fusion strategy. This involved first generating high-quality vector embeddings for each modality using the best-performing unimodal encoders identified in our main experiments. The resulting embeddings were then combined using four standard fusion techniques:
\begin{itemize}
\item \textbf{Addition}: Element-wise summation of the embedding vectors.
\item \textbf{Concatenation}: Appending the vectors to form a single, larger representation.
\item \textbf{Self-Attention}: Applying a multi-head self-attention mechanism to each embedding independently before concatenation.
\item \textbf{Cross-Attention}: Employing a bidirectional multi-head attention mechanism to allow the embeddings from each modality to mutually inform one another before concatenation.
\end{itemize}
A final classification layer was then trained on these fused representations.

\subsection{Evaluation Settings}
\subsubsection{Evaluation of Prediction Performance}
When evaluating the performance on unstructured clinical notes data, three settings were considered for evaluation: freeze, finetune, and prompt. In the freeze setting, we use the pretrained model to generate text embeddings and train a classifier on these embeddings to assess the out-of-the-box performance of the language models. For feature extraction, we utilized the embedding of the first token ([CLS]) from the final layer of BERT-style models and the embedding of the last token from the final layer for GPT-style models. The finetune setting involves further finetuning of the language model parameters together with the classifier training process. 

The generation setting prompts the large language models to directly generate prediction results from clinical notes. We use the instruction-tuned version of LLMs to make sure the task can be properly understood. The LLMs were also encouraged to output their thinking process before the final prediction. This was a deliberate methodological choice to simulate a realistic clinical decision-support scenario, where a standalone probability score is insufficient to build trust or guide practice~\cite{topol2019high,zhang2024rethinking}. While prompting for only a numerical output can simplify automated parsing~\cite{hegselmann2023tabllm}, it disconnects the model's output from the essential clinical workflow of evidence-based reasoning. The templates for the prediction tasks with clinical notes data are shown in Table~\ref{tab:mimic3_note_mortality}, Table~\ref{tab:mimic4_note_mortality} and Table~\ref{tab:mimic4_note_readmission}. 

For proprietary models (e.g., GPT-5) or large-scale models (e.g., DeepSeek-R1-671B), we only evaluated them under the prompt setting. For structured EHR data, the length of the prompt can reach several thousand tokens. Due to cost issues, we only evaluate them under the prompt setting to show their out-of-the-box clinical prediction ability. For a fair comparison against the LLMs without further domain-specific training (zero-shot performance), the ML/DL models were evaluated after minimal training (few-shot setting). Following the common medical few-shot learning settings~\cite{ge2022few}, the models were trained using only 10 samples (5 positive and 5 negative cases). We also provided the performance trained on the full dataset for reference. Additionally, we conducted multimodal integration analyses using both MIMIC-IV's structured EHR and clinical notes data, exploring fusion strategies to assess combined predictive power.

The classification tasks (mortality and readmission prediction) were evaluated using the Area Under the ROC Curve (AUROC) and the Area Under the Precision-Recall Curve (AUPRC). The regression task (LOS prediction) was evaluated using mean absolute error (MAE), mean squared error (MSE), and root mean squared error (RMSE). 

To ensure a robust interpretation of our results, we report all performance metrics with 95\% confidence intervals (CIs) computed via a bootstrapping methodology. For each experiment, we drew 100 bootstrap samples with replacement from the test set. The performance metric was calculated for each sample, and from the resulting distribution of 100 values, we report the mean and standard deviation. The 95\% CI is defined by the 2.5th and 97.5th percentiles of this distribution. Results in all tables are presented in the format: \textbf{mean}$^{\text{std}}_{[2.5\%, 97.5\%]}$.

In addition to these performance metrics, we conducted a comprehensive fairness analysis to assess potential model biases across key demographic subgroups. The detailed methodology and results of this ethical evaluation are presented in Appendix~\ref{sec:appendix_fairness}. The detailed experiment settings can be found in Appendix~\ref{sec:detail_experimental_setups}.

\subsubsection{Human Evaluation of LLM Reasoning}\label{sec:human_eval}
To validate our prompting strategy and assess the clinical utility of the LLM-generated explanations, we conducted a formal human evaluation study. This qualitative analysis was designed to move beyond predictive accuracy and measure the quality, safety, and usefulness of the models' reasoning, which is critical for their adoption in high-stakes clinical environments~\cite{sivaraman2023ignore,burgess2023healthcare}.

\paragraph{Expert panel and case selection.} We recruited a panel of five clinicians with expertise in internal and critical care medicine. We created a diverse case portfolio by randomly sampling 25 cases from the test set for each of the two primary prediction tasks (in-hospital mortality and 30-day readmission) across both data modalities (MIMIC-IV structured EHR and MIMIC-IV discharge notes). This resulted in a total pool of 100 unique patient cases. For each case, we selected the output from the best-performing LLM identified in our main experiments for that specific task setting.

\begin{figure}[!ht]
\centering
\includegraphics[width=0.49\textwidth]{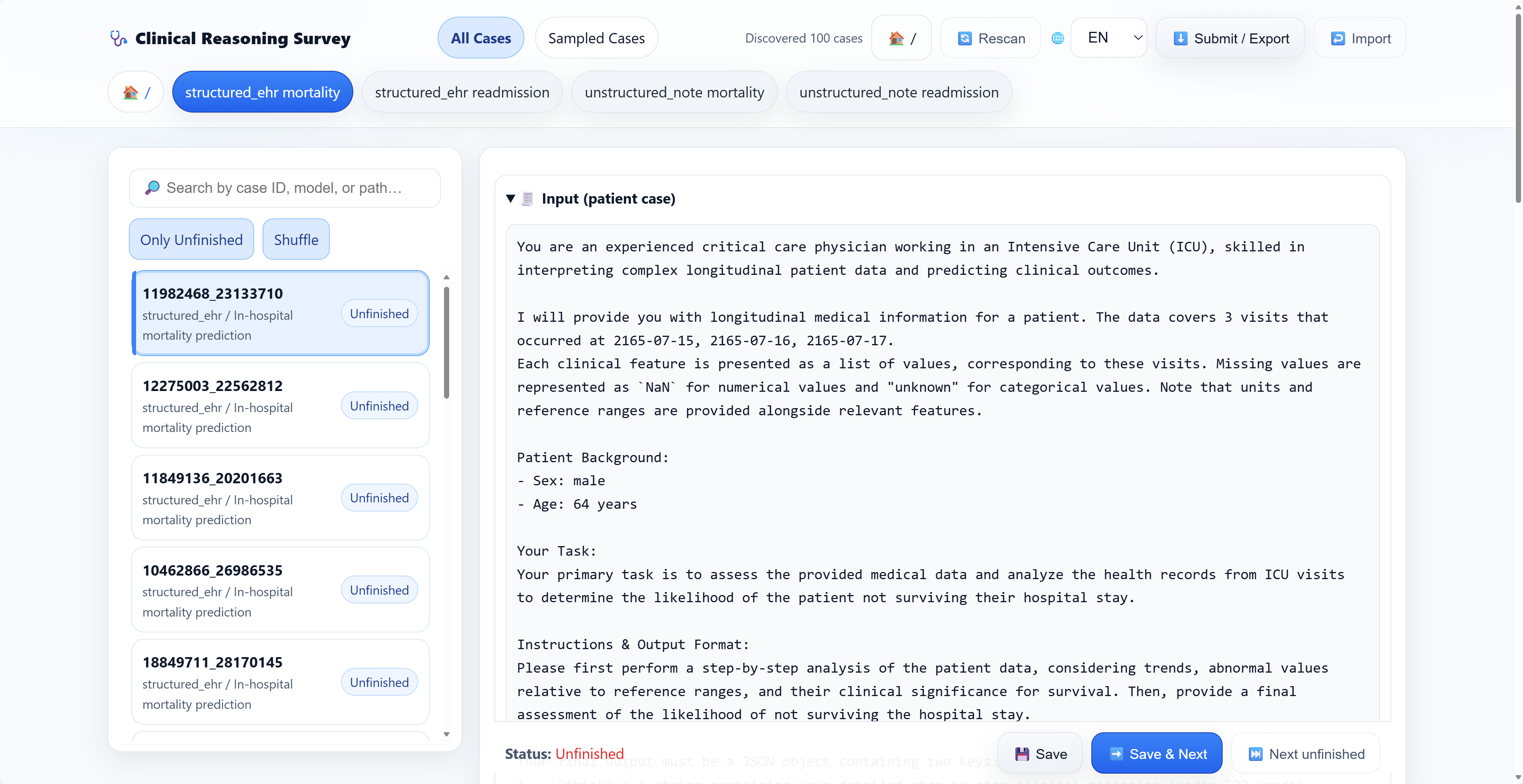}\hfill
\includegraphics[width=0.49\textwidth]{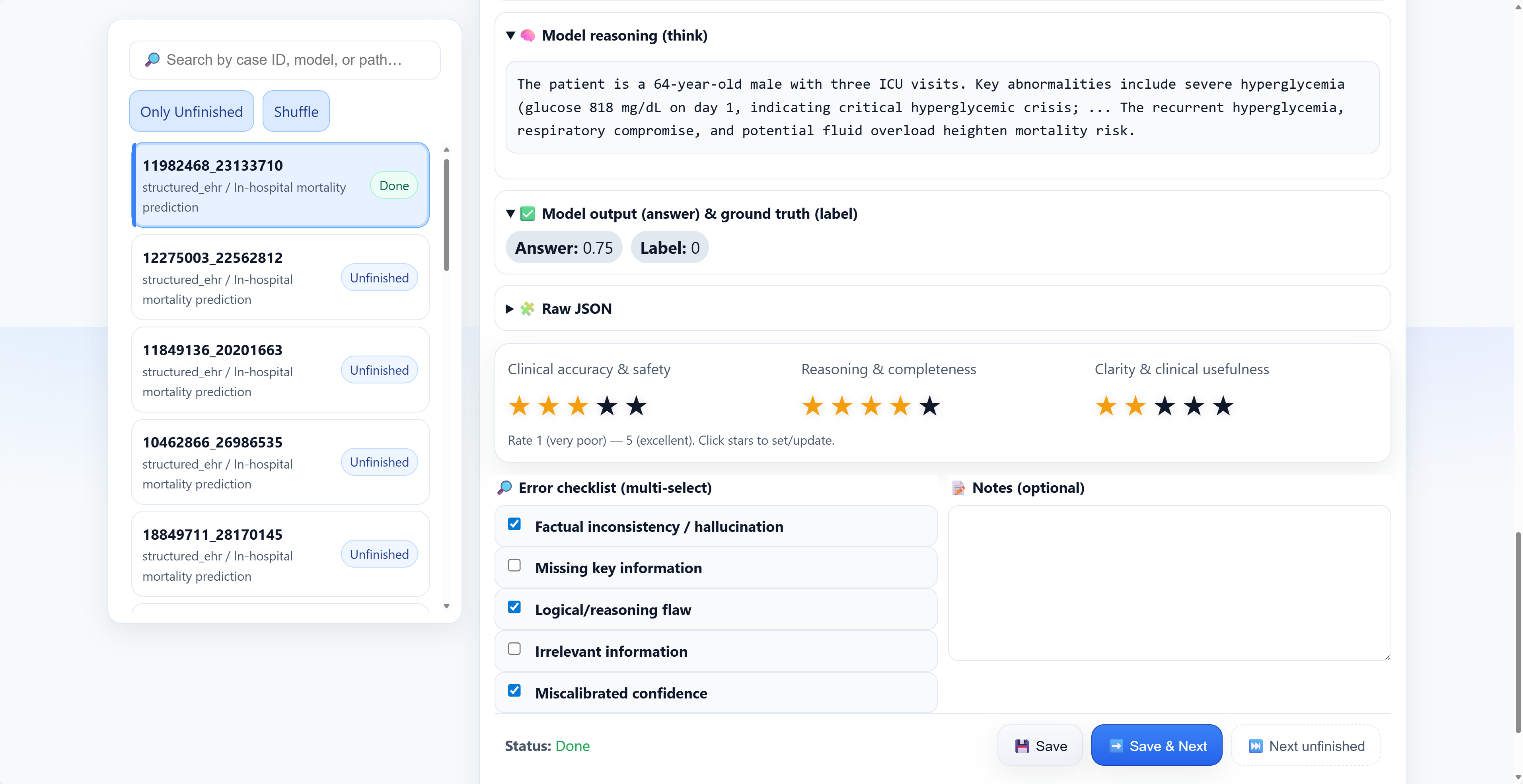}
\caption{\textit{Web-based interface for the human evaluation study.} Expert evaluators were presented with the patient's clinical data (input), the LLM's generated reasoning and prediction, and the ground truth label. They used the interface to provide scores on a 1--5 Likert scale for three quality dimensions and to select predefined error types from a checklist.}
\label{fig:human_eval_interface}
\end{figure}

\begin{figure}[!ht]
    \centering
    \includegraphics[width=1.0\linewidth]{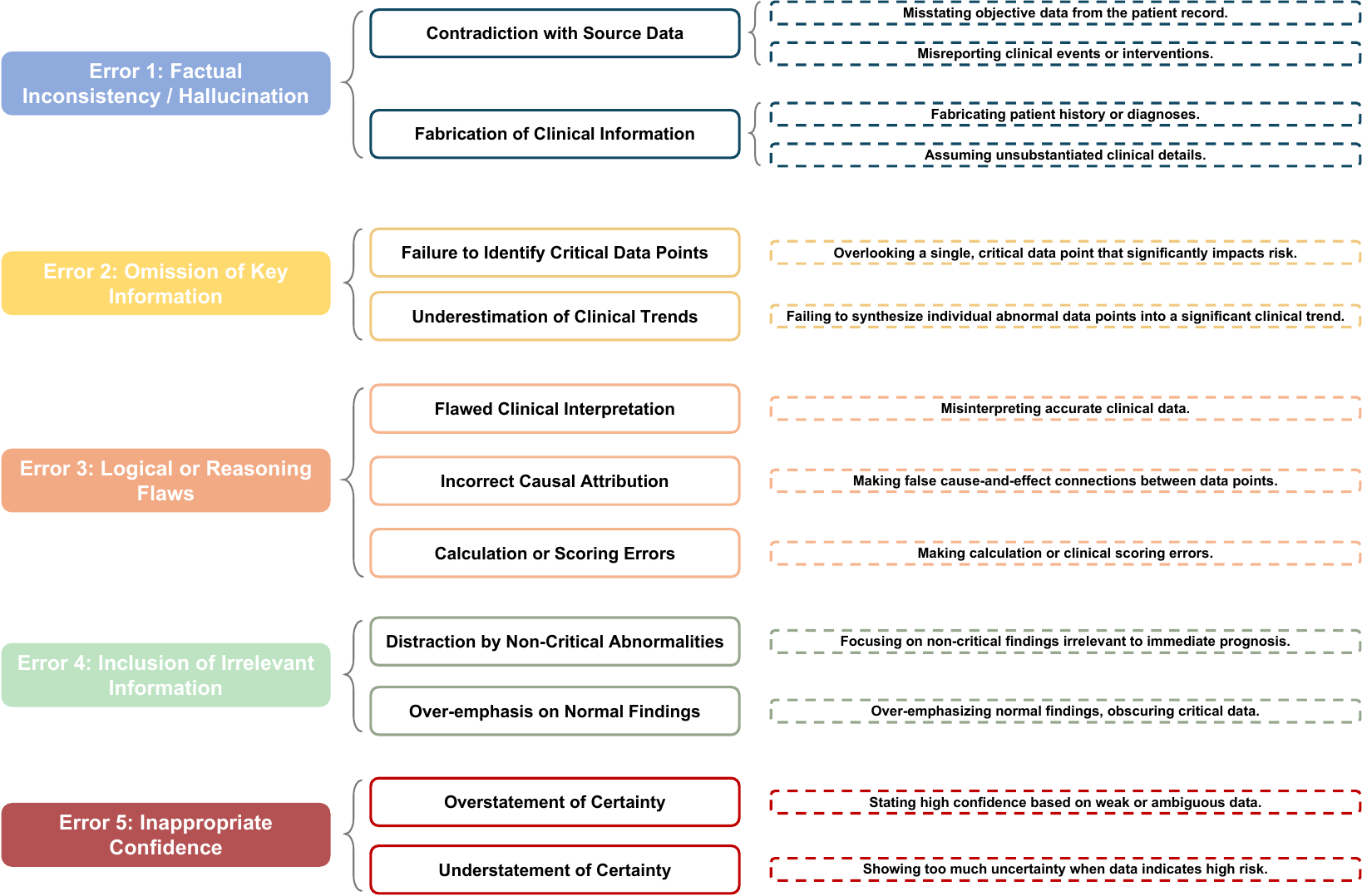}
    \caption{\textit{Hierarchically structured error taxonomy for LLM-generated clinical reasoning.} This taxonomy was developed by two expert clinicians using an open-coding thematic analysis of a pilot set of model outputs. It provided a standardized framework for the detailed error analysis.}
    \label{fig:error_taxonomy_tree}
\end{figure}

\paragraph{Evaluation framework and platform.} Each expert evaluated a random subset of 20 cases (5 from each setting). For each case, the experts were presented with the complete input provided to the LLM (either the structured EHR data or the clinical note), the model's full, unaltered output (both the reasoning and the final prediction score), and the ground truth label. The evaluation was conducted using a custom-built, secure web-based annotation system (Figure~\ref{fig:human_eval_interface}), where experts were presented with the source data, the LLM's full output, and the ground truth label. A public-facing demo of the system is available online (\url{https://yhzhu99.github.io/ehr-llm-benchmark/human-eval.html}). The framework included two components:

\begin{itemize}
\item \textbf{Quantitative Scoring:} Experts rated each explanation on a 1-5 Likert scale across three dimensions: 

\begin{enumerate}
    \item \textbf{Clinical Accuracy and Safety.} How clinically sound and factually consistent is the reasoning with the provided source data? Does it contain factually incorrect statements or clinically unsafe assertions? (1 = Very inaccurate/Unsafe, 5 = Highly accurate/Safe).
    \item \textbf{Reasoning and Completeness.} Does the model identify and logically connect the most important risk factors to its prediction? Does it omit critical information that a clinician would consider essential? (1 = Illogical/Incomplete, 5 = Logical/Comprehensive).
    \item \textbf{Clarity and Clinical Utility.} Is the explanation clear, concise, and presented in a way that would genuinely aid a clinician's decision-making process? (1 = Unclear/Useless, 5 = Very clear/Useful).
\end{enumerate}

\item \textbf{Qualitative Error Checklist:} Experts identified the presence of specific, predefined error types from a comprehensive taxonomy (Figure~\ref{fig:error_taxonomy_tree}) developed through an iterative open-coding process by two clinicians on a pilot set of cases. The primary error categories included: 

\begin{enumerate}
    \item \textbf{Factual Inconsistency / Hallucination.} The reasoning includes information that is absent from or directly contradicts the provided patient record (e.g., citing a non-existent comorbidity or an incorrect lab value). This is a form of hallucination.
    \item \textbf{Omission of Key Information.} The reasoning fails to mention critical patient data that is highly relevant to the prediction (e.g., ignoring a recent lab result indicating acute kidney injury when assessing mortality risk).
    \item \textbf{Flawed Logic or Reasoning.} The facts cited from the patient record are correct, but the clinical conclusion drawn from them is unsound, illogical, or represents a misinterpretation of their significance (e.g., incorrectly linking a stable chronic condition to an acute risk).
    \item \textbf{Inclusion of Irrelevant Information.} The reasoning focuses on non-contributory or distracting details from the patient record, which can obscure the key risk factors and weaken the clinical utility of the explanation.
    \item \textbf{Inappropriate Confidence.} The model expresses a level of certainty that is not clinically warranted. This can manifest as overconfidence in a speculative conclusion or, conversely, as undue hesitancy regarding a clear and significant risk factor.
\end{enumerate}

\end{itemize}

\subsubsection{Exception Handling and Failure-to-Predict Rates} 
A critical aspect of our evaluation was the systematic handling of instances where an LLM failed to follow prompt instructions. We established a protocol to ensure fair and reproducible comparisons. Minor formatting errors (e.g., a missing comma in a JSON output) were manually corrected to extract the valid prediction. For more significant failures where a model did not provide a prediction (a ``failure to predict''), we did not discard the sample. Instead, we imputed a default, non-informative value to penalize the model appropriately in the performance metrics. For binary classification tasks, we imputed a probability of 0.5 (equivalent to random chance), and for the regression task, we imputed the median value from the training set. This protocol ensures that all models are evaluated on the identical test set, with less reliable models being impacted accordingly.

\section{Results}

Our subsequent analysis provides a comprehensive benchmark of LLMs in clinical settings. We structure our investigation to first evaluate predictive performance across fundamental data modalities: unstructured clinical notes, where LLMs' linguistic capabilities are tested, and structured EHR data, a critical but less-explored domain for these models. We then assess their ability to integrate these sources in multimodal prediction tasks. Moving beyond quantitative accuracy, our evaluation delves into the qualitative aspects of model behavior through a human-led expert review of their clinical reasoning and a systematic analysis of common error patterns.

\subsection{Benchmarking Results of Prediction Tasks Using Clinical Notes Data}
\begin{table}[!ht]
\footnotesize
\centering
\caption{\textit{Performance comparison of BERT-style and GPT-style models on MIMIC-III mortality, MIMIC-IV mortality and readmission prediction tasks using unstructured clinical notes.} \textbf{Bold} denotes the best performance among all baselines of all settings. We use a bootstrapping strategy on all test set samples 100 times to report the mean±std results. The values in brackets represent the 95\% confidence interval, with the lower and upper bounds corresponding to the 2.5 and 97.5 percentiles, respectively. All metrics are multiplied by 100 for readability purposes.}
\label{tab:clinical_notes_supervised_results}
\resizebox{0.98\textwidth}{!}{
\begin{tabular}{c|c|c|cc|cc|cc}
\toprule
\multicolumn{2}{c|}{\multirow{2}{*}{\textbf{Method}}} & {\multirow{2}{*}{\textbf{Setting}}} & \multicolumn{2}{c|}{\textbf{MIMIC-III Outcome}} & \multicolumn{2}{c|}{\textbf{MIMIC-IV Outcome}} & \multicolumn{2}{c}{\textbf{MIMIC-IV Readmission}} \\
\multicolumn{2}{c|}{} &  & AUROC ($\uparrow$) & AUPRC ($\uparrow$) & AUROC ($\uparrow$) & AUPRC ($\uparrow$) & AUROC ($\uparrow$) & AUPRC ($\uparrow$) \\ 
\midrule
\multirow{10}{*}{BERT-style LM} & \multirow{2}{*}{BERT} & freeze & \textbothscript{\coloroutroc{79.44}}{4.43}{72.52}{87.41} & \textbothscript{\coloroutprc{35.03}}{9.37}{18.42}{54.94} & \textbothscript{\coloroutroc{69.96}}{6.78}{56.82}{80.33} & \textbothscript{\coloroutprc{24.99}}{7.33}{13.36}{38.43} & \textbothscript{\colorreadroc{62.93}}{4.15}{55.41}{70.84} & \textbothscript{\colorreadprc{38.77}}{6.13}{29.22}{52.16} \\
 &  & finetune & \textbothscript{\coloroutroc{79.21}}{4.69}{70.33}{88.38} & \textbothscript{\coloroutprc{36.11}}{9.46}{19.30}{53.28} & \textbothscript{\coloroutroc{81.04}}{5.14}{71.02}{90.25} & \textbothscript{\coloroutprc{40.90}}{9.21}{24.17}{57.33} & \textbothscript{\colorreadroc{67.99}}{4.61}{60.22}{76.00} & \textbothscript{\colorreadprc{47.20}}{7.06}{34.60}{58.39} \\ \cmidrule{2-9}

 & \multirow{2}{*}{Clinical-Longformer} & freeze & \textbothscript{\coloroutroc{84.13}}{4.14}{76.58}{91.08} & \textbothscript{\coloroutprc{40.07}}{10.21}{21.94}{57.28} & \textbothscript{\coloroutroc{75.68}}{6.15}{62.75}{85.54} & \textbothscript{\coloroutprc{36.81}}{10.80}{17.18}{55.36} & \textbothscript{\colorreadroc{65.04}}{4.98}{54.38}{74.26} & \textbothscript{\colorreadprc{48.70}}{6.04}{36.39}{60.57} \\
 &  & finetune & \textbothscript{\coloroutroc{82.96}}{3.80}{76.04}{89.17} & \textbothscript{\coloroutprc{41.97}}{9.90}{23.37}{61.03} & \textbothscript{\coloroutroc{88.06}}{4.45}{77.84}{95.62} & \textbothscript{\coloroutprc{62.92}}{9.40}{45.90}{81.03} & \textbothscript{\colorreadroc{75.56}}{4.42}{66.65}{83.52} & \textbothscript{\colorreadprc{59.19}}{6.65}{45.26}{71.11} \\ \cmidrule{2-9}

 & \multirow{2}{*}{BioBERT} & freeze & \textbothscript{\coloroutroc{76.19}}{5.23}{65.61}{85.41} & \textbothscript{\coloroutprc{25.91}}{7.36}{13.08}{38.38} & \textbothscript{\coloroutroc{75.30}}{4.95}{65.96}{85.68} & \textbothscript{\coloroutprc{29.49}}{8.17}{14.78}{43.90} & \textbothscript{\colorreadroc{62.86}}{4.43}{55.24}{73.80} & \textbothscript{\colorreadprc{37.10}}{5.40}{28.13}{49.01} \\
 &  & finetune & \textbothscript{\coloroutroc{83.75}}{4.24}{75.07}{91.87} & \textbothscript{\coloroutprc{41.88}}{9.61}{22.87}{60.82} & \textbothscript{\coloroutroc{75.85}}{6.63}{62.72}{88.99} & \textbothscript{\coloroutprc{36.93}}{10.89}{17.23}{59.53} & \textbothscript{\colorreadroc{70.75}}{4.68}{61.87}{80.76} & \textbothscript{\colorreadprc{54.06}}{7.12}{41.31}{67.34} \\ \cmidrule{2-9}

 & \multirow{2}{*}{GatorTron} & freeze & \textbothscript{\coloroutroc{87.70}}{3.73}{80.64}{94.43} & \textbothscript{\coloroutprc{48.56}}{10.87}{24.50}{67.34} & \textbothscript{\coloroutroc{89.47}}{4.45}{79.29}{96.78} & \textbothscript{\coloroutprc{54.53}}{9.90}{33.65}{73.84} & \textbothscript{\colorreadroc{70.75}}{4.24}{62.90}{79.97} & \textbothscript{\colorreadprc{47.74}}{7.09}{35.04}{60.80} \\
 &  & finetune & \textbothscript{\coloroutroc{87.97}}{3.55}{79.58}{93.86} & \textbothscript{\coloroutprc{50.69}}{10.95}{28.45}{68.03} & \textbothscript{\coloroutroc{91.47}}{3.44}{85.18}{97.53} & \textbothscript{\coloroutprc{71.43}}{8.79}{51.61}{86.52} & \textbothscript{\colorreadroc{75.96}}{4.34}{68.37}{84.92} & \textbothscript{\colorreadprc{59.01}}{6.85}{45.54}{72.08} \\ \cmidrule{2-9}

 & {\multirow{2}{*}{ClinicalBERT}} & freeze & \textbothscript{\coloroutroc{83.99}}{4.47}{74.31}{92.15} & \textbothscript{\coloroutprc{35.79}}{8.51}{20.05}{54.10} & \textbothscript{\coloroutroc{73.55}}{5.69}{61.56}{84.64} & \textbothscript{\coloroutprc{26.98}}{8.32}{14.16}{43.85} & \textbothscript{\colorreadroc{64.94}}{4.40}{57.15}{72.63} & \textbothscript{\colorreadprc{40.90}}{6.10}{31.64}{50.67} \\
 &  & finetune & \textbothscript{\coloroutroc{82.84}}{4.87}{72.08}{91.09} & \textbothscript{\coloroutprc{31.43}}{7.80}{14.47}{47.34} & \textbothscript{\coloroutroc{84.95}}{4.72}{75.66}{92.65} & \textbothscript{\coloroutprc{53.01}}{10.67}{31.62}{71.69} & \textbothscript{\colorreadroc{71.93}}{4.35}{64.13}{81.01} & \textbothscript{\colorreadprc{53.82}}{6.85}{42.34}{67.02} \\ \midrule

\multirow{18}{*}{GPT-style Base LLM} & \multirow{2}{*}{GPT-2-117M} & freeze & \textbothscript{\coloroutroc{70.63}}{7.19}{52.82}{82.87} & \textbothscript{\coloroutprc{25.00}}{8.04}{10.96}{39.97} & \textbothscript{\coloroutroc{65.60}}{6.36}{53.22}{79.07} & \textbothscript{\coloroutprc{17.08}}{5.85}{8.31}{31.60} & \textbothscript{\colorreadroc{60.43}}{5.00}{50.42}{71.43} & \textbothscript{\colorreadprc{37.05}}{6.14}{26.26}{49.81} \\
 &  & finetune & \textbothscript{\coloroutroc{84.68}}{4.72}{76.31}{92.73} & \textbothscript{\coloroutprc{44.00}}{10.87}{22.51}{65.03} & \textbothscript{\coloroutroc{82.30}}{5.50}{70.86}{90.65} & \textbothscript{\coloroutprc{46.68}}{10.38}{25.64}{65.17} & \textbothscript{\colorreadroc{71.81}}{4.58}{63.35}{79.64} & \textbothscript{\colorreadprc{50.14}}{6.67}{36.96}{62.47} \\ \cmidrule{2-9}

 & {\multirow{2}{*}{BioGPT-347M}} & freeze & \textbothscript{\coloroutroc{72.35}}{6.49}{59.32}{84.53} & \textbothscript{\coloroutprc{29.36}}{9.25}{11.18}{46.13} & \textbothscript{\coloroutroc{77.41}}{3.78}{70.24}{84.06} & \textbothscript{\coloroutprc{24.94}}{7.17}{12.46}{39.82} & \textbothscript{\colorreadroc{66.54}}{4.04}{57.39}{73.47} & \textbothscript{\colorreadprc{37.62}}{5.37}{27.45}{50.21} \\
 &  & finetune & \textbothscript{\coloroutroc{86.43}}{3.73}{79.06}{93.56} & \textbothscript{\coloroutprc{41.74}}{11.16}{18.73}{64.46} & \textbothscript{\coloroutroc{88.39}}{4.51}{78.21}{95.35} & \textbothscript{\coloroutprc{62.62}}{10.42}{44.43}{79.48} & \textbothscript{\colorreadroc{71.66}}{4.10}{63.66}{79.27} & \textbothscript{\colorreadprc{52.23}}{6.63}{41.02}{65.14} \\ \cmidrule{2-9}

 & {\multirow{2}{*}{Meditron-7B}} & freeze & \textbothscript{\coloroutroc{70.23}}{5.92}{59.58}{81.00} & \textbothscript{\coloroutprc{24.92}}{8.33}{11.61}{42.10} & \textbothscript{\coloroutroc{71.98}}{6.15}{59.60}{82.84} & \textbothscript{\coloroutprc{26.77}}{8.52}{11.78}{42.27} & \textbothscript{\colorreadroc{59.70}}{4.32}{51.02}{67.99} & \textbothscript{\colorreadprc{34.97}}{5.50}{26.30}{45.96} \\
 &  & finetune & \textbothscript{\coloroutroc{84.94}}{3.84}{77.77}{92.92} & \textbothscript{\coloroutprc{45.89}}{10.57}{26.76}{67.67} & \textbothscript{\coloroutroc{79.05}}{5.05}{68.87}{88.28} & \textbothscript{\coloroutprc{39.10}}{9.84}{21.88}{57.37} & \textbothscript{\colorreadroc{64.04}}{4.40}{55.71}{71.87} & \textbothscript{\colorreadprc{40.90}}{6.21}{30.65}{52.60} \\ \cmidrule{2-9}

 & {\multirow{2}{*}{BioMistral-7B}} & freeze & \textbothscript{\coloroutroc{62.57}}{6.56}{49.73}{73.03} & \textbothscript{\coloroutprc{20.31}}{7.41}{8.18}{36.28} & \textbothscript{\coloroutroc{70.78}}{7.30}{57.30}{84.19} & \textbothscript{\coloroutprc{33.64}}{9.81}{14.76}{50.29} & \textbothscript{\colorreadroc{67.41}}{4.09}{58.87}{74.18} & \textbothscript{\colorreadprc{42.95}}{5.55}{33.47}{54.67} \\
 &  & finetune & \textbothscript{\coloroutroc{82.81}}{4.43}{73.85}{90.69} & \textbothscript{\coloroutprc{52.05}}{9.23}{34.63}{69.28} & \textbothscript{\coloroutroc{74.42}}{5.88}{64.01}{85.39} & \textbothscript{\coloroutprc{40.31}}{9.32}{20.44}{56.86} & \textbothscript{\colorreadroc{75.66}}{4.37}{67.22}{83.81} & \textbothscript{\colorreadprc{61.60}}{6.28}{50.18}{73.19} \\ \cmidrule{2-9}

 & {\multirow{3}{*}{OpenBioLLM-8B}} & freeze & \textbothscript{\coloroutroc{74.79}}{5.57}{62.18}{84.25} & \textbothscript{\coloroutprc{25.65}}{7.08}{12.41}{40.22} & \textbothscript{\coloroutroc{61.24}}{6.51}{49.55}{72.64} & \textbothscript{\coloroutprc{12.85}}{3.55}{6.80}{21.46} & \textbothscript{\colorreadroc{57.09}}{3.50}{51.49}{64.09} & \textbothscript{\colorreadprc{32.06}}{5.08}{23.74}{42.99} \\
 &  & finetune & \textbothscript{\coloroutroc{83.43}}{4.84}{73.61}{92.35} & \textbothscript{\coloroutprc{43.77}}{9.89}{25.21}{59.20} & \textbothscript{\coloroutroc{80.85}}{5.71}{69.32}{90.78} & \textbothscript{\coloroutprc{55.56}}{11.92}{31.02}{75.66} & \textbothscript{\colorreadroc{77.22}}{3.73}{71.40}{85.29} & \textbothscript{\colorreadprc{63.91}}{5.82}{53.78}{74.04} \\
 &  & prompt & \textbothscript{\coloroutroc{51.13}}{2.85}{45.32}{55.47} & \textbothscript{\coloroutprc{9.95}}{2.31}{5.76}{14.20} & \textbothscript{\coloroutroc{65.70}}{5.43}{54.22}{75.99} & \textbothscript{\coloroutprc{13.81}}{3.35}{8.04}{19.91} & \textbothscript{\colorreadroc{58.47}}{4.21}{49.18}{66.08} & \textbothscript{\colorreadprc{32.99}}{4.66}{24.52}{42.44} \\ \cmidrule{2-9}

 & {\multirow{3}{*}{Qwen2.5-7B}} & freeze & \textbothscript{\coloroutroc{61.79}}{6.20}{47.86}{72.52} & \textbothscript{\coloroutprc{17.39}}{5.50}{9.37}{30.37} & \textbothscript{\coloroutroc{64.52}}{6.46}{50.88}{77.19} & \textbothscript{\coloroutprc{14.31}}{4.61}{7.90}{24.25} & \textbothscript{\colorreadroc{61.48}}{4.46}{53.46}{69.25} & \textbothscript{\colorreadprc{40.03}}{5.98}{28.79}{51.05} \\
 &  & finetune & \textbothscript{\coloroutroc{83.97}}{5.10}{73.50}{92.60} & \textbothscript{\coloroutprc{47.20}}{10.31}{25.95}{67.38} & \textbothscript{\coloroutroc{89.04}}{3.93}{81.63}{94.78} & \textbothscript{\coloroutprc{60.89}}{10.18}{37.67}{80.88} & \textbothscript{\colorreadroc{75.80}}{4.00}{68.51}{84.32} & \textbothscript{\colorreadprc{58.42}}{6.72}{46.63}{71.38} \\
 &  & prompt & \textbothscript{\coloroutroc{74.18}}{5.52}{63.44}{84.91} & \textbothscript{\coloroutprc{23.85}}{7.06}{11.62}{37.00} & \textbothscript{\coloroutroc{88.39}}{4.11}{79.74}{94.75} & \textbothscript{\coloroutprc{44.33}}{8.33}{26.89}{61.62} & \textbothscript{\colorreadroc{77.32}}{4.75}{67.46}{84.95} & \textbothscript{\colorreadprc{66.64}}{6.05}{54.62}{76.82} \\ \cmidrule{2-9}

 & {\multirow{2}{*}{Gemma-3-4B}} & freeze & \textbothscript{\coloroutroc{77.96}}{4.61}{69.43}{85.73} & \textbothscript{\coloroutprc{32.77}}{9.36}{15.03}{47.99} & \textbothscript{\coloroutroc{66.24}}{6.26}{53.07}{77.24} & \textbothscript{\coloroutprc{28.78}}{9.54}{13.17}{46.71} & \textbothscript{\colorreadroc{63.83}}{4.27}{55.88}{72.40} & \textbothscript{\colorreadprc{35.15}}{4.70}{27.57}{43.28} \\
 &  & prompt & \textbothscript{\coloroutroc{74.97}}{4.88}{65.46}{85.19} & \textbothscript{\coloroutprc{27.65}}{8.39}{12.27}{43.36} & \textbothscript{\coloroutroc{96.57}}{1.32}{93.76}{98.66} & \textbothscript{\coloroutprc{68.97}}{9.29}{49.29}{84.65} & \textbothscript{\colorreadroc{85.21}}{3.92}{77.63}{91.80} & \textbothscript{\colorreadprc{71.51}}{5.77}{60.13}{83.17} \\ \cmidrule{2-9}

 & {DeepSeek-V3.1} & prompt &\textbothscript{\coloroutroc{88.22}}{3.20}{81.46}{93.73} & \textbothscript{\coloroutprc{47.44}}{11.73}{26.66}{67.26} & \textbothscript{\textbf{\coloroutroc{97.89}}}{0.74}{96.30}{99.07} & \textbothscript{\coloroutprc{69.15}}{8.95}{54.07}{85.94} & \textbothscript{\colorreadroc{80.11}}{3.79}{74.45}{86.99} & \textbothscript{\colorreadprc{72.60}}{5.32}{62.92}{82.07} \\ \cmidrule{2-9}
 
 & {GPT-4o} & prompt & \textbothscript{\coloroutroc{87.83}}{3.79}{79.09}{93.96} & \textbothscript{\coloroutprc{46.98}}{11.07}{23.28}{65.01} & \textbothscript{\coloroutroc{93.99}}{2.00}{89.94}{97.74} & \textbothscript{\coloroutprc{58.04}}{10.01}{38.26}{74.10} & \textbothscript{\colorreadroc{75.89}}{5.51}{64.63}{84.85} & \textbothscript{\colorreadprc{66.38}}{6.65}{54.89}{78.05} \\ \midrule 

 \multirow{10}{*}{Reasoning LLM} & \multirow{3}{*}{HuatuoGPT-o1-7B}& freeze & \textbothscript{\coloroutroc{67.03}}{6.25}{53.56}{78.47} & \textbothscript{\coloroutprc{22.63}}{6.61}{11.86}{38.76} & \textbothscript{\coloroutroc{60.69}}{5.52}{49.14}{70.24} & \textbothscript{\coloroutprc{11.67}}{2.88}{6.92}{17.61} & \textbothscript{\colorreadroc{59.14}}{4.26}{51.61}{66.31} & \textbothscript{\colorreadprc{37.67}}{6.05}{27.97}{48.48} \\
 &  & finetune & \textbothscript{\coloroutroc{85.99}}{4.15}{77.19}{93.64} & \textbothscript{\coloroutprc{46.46}}{10.63}{22.99}{63.85} & \textbothscript{\coloroutroc{88.16}}{3.95}{80.44}{95.26} & \textbothscript{\coloroutprc{62.39}}{9.52}{38.11}{76.29} & \textbothscript{\colorreadroc{77.82}}{3.84}{71.37}{85.64} & \textbothscript{\colorreadprc{62.21}}{6.25}{50.00}{73.01} \\
 &  & prompt & \textbothscript{\coloroutroc{70.13}}{5.57}{57.72}{78.71} & \textbothscript{\coloroutprc{16.41}}{4.58}{8.07}{24.45} & \textbothscript{\coloroutroc{83.84}}{3.55}{76.47}{89.76} & \textbothscript{\coloroutprc{29.72}}{6.79}{16.30}{42.02} & \textbothscript{\colorreadroc{67.33}}{5.35}{58.60}{78.24} & \textbothscript{\colorreadprc{49.50}}{7.09}{35.77}{62.65} \\ \cmidrule{2-9}

 & \multirow{3}{*}{DeepSeek-R1-7B} & freeze & \textbothscript{\coloroutroc{77.77}}{4.72}{68.63}{85.90} & \textbothscript{\coloroutprc{26.79}}{7.13}{15.23}{43.79} & \textbothscript{\coloroutroc{56.76}}{6.73}{43.21}{69.55} & \textbothscript{\coloroutprc{16.72}}{7.24}{6.36}{32.22} & \textbothscript{\colorreadroc{53.21}}{3.81}{47.27}{61.39} & \textbothscript{\colorreadprc{33.12}}{5.35}{23.18}{45.66} \\
 &  & finetune & \textbothscript{\coloroutroc{82.98}}{4.92}{72.89}{91.98} & \textbothscript{\coloroutprc{43.23}}{10.21}{25.64}{65.69} & \textbothscript{\coloroutroc{87.32}}{5.98}{74.68}{94.95} & \textbothscript{\coloroutprc{55.75}}{12.11}{30.84}{73.12} & \textbothscript{\colorreadroc{74.54}}{3.94}{67.12}{82.69} & \textbothscript{\colorreadprc{51.13}}{6.80}{37.24}{63.84} \\
 &  & prompt & \textbothscript{\coloroutroc{70.27}}{6.33}{56.87}{82.57} & \textbothscript{\coloroutprc{21.89}}{7.61}{10.08}{37.12} & \textbothscript{\coloroutroc{83.17}}{4.63}{73.15}{91.58} & \textbothscript{\coloroutprc{36.25}}{8.75}{20.30}{53.90} & \textbothscript{\colorreadroc{61.69}}{4.65}{52.74}{70.65} & \textbothscript{\colorreadprc{34.50}}{5.75}{23.89}{45.31} \\ \cmidrule{2-9}
 
 & {DeepSeek-R1} & prompt & \textbothscript{\textbf{\coloroutroc{90.75}}}{2.59}{85.85}{94.84} & \textbothscript{\coloroutprc{48.72}}{11.32}{26.91}{67.63} & \textbothscript{\coloroutroc{96.25}}{1.28}{93.18}{98.35} & \textbothscript{\coloroutprc{57.27}}{10.24}{35.88}{76.49} & \textbothscript{\colorreadroc{41.95}}{6.35}{29.26}{52.78} & \textbothscript{\colorreadprc{37.99}}{6.14}{25.53}{47.47} \\ \cmidrule{2-9}

 & {DeepSeek-V3.1-Think} & prompt & \textbothscript{\coloroutroc{88.42}}{2.65}{83.19}{93.11} & \textbothscript{\coloroutprc{47.89}}{10.78}{25.70}{66.96} & \textbothscript{\coloroutroc{91.53}}{4.53}{82.45}{97.75} & \textbothscript{\coloroutprc{53.73}}{9.19}{37.64}{70.71} & \textbothscript{\colorreadroc{67.41}}{5.40}{57.19}{77.66} & \textbothscript{\colorreadprc{58.02}}{6.70}{42.64}{68.67} \\ \cmidrule{2-9}

 & {o3-mini-high} & prompt & \textbothscript{\coloroutroc{88.68}}{3.53}{81.73}{95.36} & \textbothscript{\coloroutprc{53.54}}{10.92}{32.10}{71.28} & \textbothscript{\coloroutroc{97.45}}{1.01}{95.11}{99.57} & \textbothscript{\textbf{\coloroutprc{71.72}}}{10.31}{51.45}{91.96} & \textbothscript{\textbf{\colorreadroc{87.59}}}{3.34}{80.53}{94.74} & \textbothscript{\colorreadprc{75.48}}{5.74}{65.54}{85.32} \\ \cmidrule{2-9}
 
 & GPT-5 & prompt & \textbothscript{\coloroutroc{89.75}}{3.67}{82.01}{95.89} & \textbothscript{\textbf{\coloroutprc{58.07}}}{11.56}{35.39}{75.67} & \textbothscript{\coloroutroc{97.60}}{0.75}{95.96}{98.79} & \textbothscript{\coloroutprc{66.36}}{8.91}{49.15}{84.36} & \textbothscript{\colorreadroc{86.34}}{3.54}{79.78}{92.62} & \textbothscript{\textbf{\colorreadprc{76.90}}}{5.54}{65.17}{86.33} \\
\bottomrule
\end{tabular}
}
\end{table}
Our analysis of prediction tasks using unstructured clinical notes distinguishes between clinically relevant prospective prediction and retrospective document classification, revealing the robust capabilities of modern LLMs in both settings (Table \ref{tab:clinical_notes_supervised_results}).

For the methodologically rigorous prospective in-hospital mortality task, which uses only admission notes from the first 24 hours (MIMIC-III), state-of-the-art LLMs demonstrated powerful zero-shot predictive capabilities. As anticipated, this task proved more challenging than retrospective analysis, resulting in generally lower performance. We find that leading LLMs outperformed specialized, finetuned BERT-style models. The DeepSeek-R1 achieved the highest AUROC (90.75\%), representing a significant improvement over the 87.97\% AUROC from the finetuned GatorTron model, a strong clinical NLP baseline. Other advanced models, including the newly released GPT-5 (89.75\% AUROC), and DeepSeek-V3.1-Think (88.42\% AUROC), also delivered highly competitive performance, surpassing the finetuned specialist model. These results provide compelling evidence that the predictive power of modern LLMs extends to challenging, clinically realistic prospective scenarios.

In the retrospective document classification tasks using MIMIC-IV discharge summaries, which assess a model's ability to synthesize information from a complete hospital record, the latest generation of LLMs achieved exceptionally high performance in a zero-shot setting. For mortality classification, DeepSeek-V3.1 attained a near-perfect AUROC of 97.89\%, followed closely by GPT-5 (97.6\%) and o3-mini-high (97.45\%). For 30-day readmission prediction, a prospective task predicted at discharge, o3-mini-high (87.59\% AUROC) and GPT-5 (86.34\% AUROC) demonstrated top performance. In all note-based tasks, these zero-shot LLM results substantially surpassed the best finetuned BERT-style model, GatorTron.

Across both prospective and retrospective evaluations, a key observation is the consistent high performance of open-source models. While proprietary models like GPT-4o and GPT-5 performed strongly, open-source competitors such as DeepSeek-R1 and DeepSeek-V3.1 achieved comparable or superior results. This is particularly relevant for healthcare, as it enables organizations to leverage cutting-edge AI for clinical prediction while maintaining full control over sensitive patient data through secure on-premise deployments.

Regarding model settings, for the most advanced LLMs, the zero-shot prompt setting consistently yielded the best results, underscoring their powerful out-of-the-box clinical text understanding. For less advanced LLMs (e.g., Qwen2.5-7B, HuatuoGPT-o1-7B), finetuning generally improved performance over using pretrained embeddings (freeze setting) or direct prompting. As noted previously, several older or smaller finetuned LLMs (e.g., BioGPT, Meditron, GPT-2) failed to follow prompts correctly and were excluded from the ``prompt'' evaluation.

\subsection{Benchmarking Results of Prediction Tasks Using Structured EHR Data}

\begin{table*}[!ht]
\centering
\caption{\textit{Performance of mortality and readmission prediction on TJH and MIMIC-IV datasets using structured EHR.} \textbf{Bold} indicates the best performance excluding results using the full dataset. \textit{Italic} indicates the proposed prompting framework outperforms basic prompts. We use bootstrapping on all test set samples 100 times to report the mean±std results. The values in brackets represent the 95\% confidence interval, with the lower and upper bounds corresponding to the 2.5 and 97.5 percentiles, respectively. All metrics are multiplied by 100 for readability purposes.}
\label{tab:ehr_results}
\centering
\footnotesize
\resizebox{\linewidth}{!}
{
\begin{tabular}{c|c|c|cc|ccc|cc|cc}
\toprule
\multicolumn{2}{c|}{\multirow{2}{*}{\textbf{Methods}}} & {\multirow{2}{*}{\textbf{Setting}}} & \multicolumn{2}{c|}{\textbf{TJH Outcome}} & \multicolumn{3}{c|}{\textbf{TJH LOS}} & \multicolumn{2}{c|}{\textbf{MIMIC-IV Outcome}} & \multicolumn{2}{c}{\textbf{MIMIC-IV Readmission}} \\

\multicolumn{2}{c|}{} &  & AUROC ($\uparrow$) & {AUPRC ($\uparrow$)} & MAE ($\downarrow$) & MSE ($\downarrow$) & {RMSE ($\downarrow$)} & AUROC ($\uparrow$) & {AUPRC ($\uparrow$)} & AUROC ($\uparrow$) & AUPRC ($\uparrow$) \\ \midrule

{\multirow{8}{*}{ML}} & \multirow{2}{*}{CatBoost} & 10 shot & \textbothscript{\textbf{\colortjhroc{99.43}}}{0.31}{98.63}{99.86} & \textbothscript{\textbf{\colortjhprc{99.31}}}{0.39}{98.27}{99.84} & \textbothscript{\colortjhmae{4.14}}{0.18}{3.85}{4.51} & \textbothscript{\colortjhmse{24.04}}{3.63}{17.73}{31.82} & \textbothscript{\colortjhrmse{4.89}}{0.37}{4.21}{5.64} & \textbothscript{\colormimicoutroc{62.18}}{7.41}{48.23}{77.33} & \textbothscript{\colormimicoutprc{19.48}}{5.81}{9.47}{33.78} & \textbothscript{\colormimicreadroc{52.29}}{5.39}{41.26}{62.44} & \textbothscript{\colormimicreadprc{26.96}}{5.10}{18.07}{36.47} \\ 
 &  & full shot & \textbothscript{\colortjhroc{99.16}}{0.47}{97.92}{99.82} & \textbothscript{\colortjhprc{98.99}}{0.59}{97.41}{99.81} & \textbothscript{\colortjhmae{3.09}}{0.21}{2.78}{3.54} & \textbothscript{\colortjhmse{18.61}}{4.24}{11.48}{27.57} & \textbothscript{\colortjhrmse{4.29}}{0.49}{3.39}{5.25} & \textbothscript{\colormimicoutroc{71.18}}{6.72}{58.63}{83.07} & \textbothscript{\colormimicoutprc{28.27}}{8.53}{12.32}{43.51} & \textbothscript{\colormimicreadroc{61.78}}{4.82}{53.16}{71.69} & \textbothscript{\colormimicreadprc{31.76}}{5.71}{20.78}{43.65} \\ \cmidrule{2-12}
 & \multirow{2}{*}{DT} & 10 shot & \textbothscript{\colortjhroc{79.64}}{2.78}{73.62}{84.02} & \textbothscript{\colortjhprc{69.20}}{4.42}{59.08}{76.00} & \textbothscript{\colortjhmae{6.25}}{0.72}{5.15}{7.70} & \textbothscript{\colortjhmse{118.50}}{17.13}{93.28}{158.93} & \textbothscript{\colortjhrmse{10.86}}{0.78}{9.66}{12.61} & \textbothscript{\colormimicoutroc{59.48}}{5.84}{48.15}{69.76} & \textbothscript{\colormimicoutprc{11.98}}{3.23}{6.91}{17.83} & \textbothscript{\colormimicreadroc{55.68}}{3.38}{49.91}{61.78} & \textbothscript{\colormimicreadprc{24.91}}{3.75}{18.12}{31.94} \\ 
 &  & full shot & \textbothscript{\colortjhroc{92.20}}{1.83}{88.70}{95.54} & \textbothscript{\colortjhprc{87.79}}{3.04}{82.26}{93.06} & \textbothscript{\colortjhmae{3.53}}{0.42}{2.66}{4.20} & \textbothscript{\colortjhmse{50.40}}{9.04}{32.44}{66.14} & \textbothscript{\colortjhrmse{7.07}}{0.65}{5.70}{8.13} & \textbothscript{\colormimicoutroc{51.81}}{3.69}{45.99}{59.06} & \textbothscript{\colormimicoutprc{10.48}}{2.64}{6.42}{16.60} & \textbothscript{\colormimicreadroc{51.55}}{2.72}{46.39}{55.84} & \textbothscript{\colormimicreadprc{23.65}}{3.72}{17.50}{32.97} \\ \cmidrule{2-12}
 & \multirow{2}{*}{RF} & 10 shot & \textbothscript{\colortjhroc{99.16}}{0.44}{98.25}{99.79} & \textbothscript{\colortjhprc{98.99}}{0.54}{97.56}{99.73} & \textbothscript{\colortjhmae{4.60}}{0.22}{4.17}{4.96} & \textbothscript{\colortjhmse{31.21}}{3.70}{24.44}{38.28} & \textbothscript{\colortjhrmse{5.58}}{0.33}{4.94}{6.19} & \textbothscript{\colormimicoutroc{59.92}}{8.79}{41.40}{76.81} & \textbothscript{\colormimicoutprc{22.51}}{7.55}{10.00}{36.97} & \textbothscript{\colormimicreadroc{58.33}}{5.27}{49.27}{67.85} & \textbothscript{\colormimicreadprc{31.21}}{5.46}{21.29}{41.72} \\ 
 &  & full shot & \textbothscript{\colortjhroc{99.18}}{0.46}{97.95}{99.84} & \textbothscript{\colortjhprc{99.05}}{0.56}{97.46}{99.83} & \textbothscript{\colortjhmae{3.09}}{0.24}{2.65}{3.62} & \textbothscript{\colortjhmse{23.04}}{4.74}{15.23}{32.82} & \textbothscript{\colortjhrmse{4.77}}{0.49}{3.90}{5.73} & \textbothscript{\colormimicoutroc{67.06}}{5.54}{54.44}{76.96} & \textbothscript{\colormimicoutprc{15.89}}{3.84}{8.77}{23.22} & \textbothscript{\colormimicreadroc{57.43}}{4.32}{49.20}{65.44} & \textbothscript{\colormimicreadprc{29.67}}{5.34}{20.69}{40.52} \\ \cmidrule{2-12}
 & \multirow{2}{*}{XGBoost} & 10 shot & \textbothscript{\colortjhroc{62.66}}{3.38}{57.07}{69.99} & \textbothscript{\colortjhprc{53.15}}{4.37}{45.58}{61.53} & \textbothscript{\colortjhmae{4.22}}{0.18}{3.90}{4.57} & \textbothscript{\colortjhmse{24.73}}{3.59}{18.53}{32.41} & \textbothscript{\colortjhrmse{4.96}}{0.36}{4.30}{5.69} & \textbothscript{\colormimicoutroc{55.77}}{5.43}{43.88}{66.86} & \textbothscript{\colormimicoutprc{10.76}}{2.55}{6.56}{15.20} & \textbothscript{\colormimicreadroc{56.40}}{4.53}{48.66}{66.23} & \textbothscript{\colormimicreadprc{29.10}}{4.63}{20.31}{38.08} \\ 
 &  & full shot & \textbothscript{\colortjhroc{98.05}}{0.94}{96.09}{99.70} & \textbothscript{\colortjhprc{95.58}}{2.18}{90.74}{99.47} & \textbothscript{\colortjhmae{3.06}}{0.21}{2.72}{3.50} & \textbothscript{\colortjhmse{18.70}}{4.29}{11.62}{27.68} & \textbothscript{\colortjhrmse{4.30}}{0.49}{3.41}{5.26} & \textbothscript{\colormimicoutroc{64.62}}{4.97}{56.22}{76.20} & \textbothscript{\colormimicoutprc{17.66}}{5.12}{9.02}{30.21} & \textbothscript{\colormimicreadroc{64.23}}{4.34}{56.21}{72.04} & \textbothscript{\colormimicreadprc{34.31}}{6.65}{23.59}{48.39} \\ \midrule

{\multirow{14}{*}{DL}} & \multirow{2}{*}{GRU} & 10 shot & \textbothscript{\colortjhroc{87.79}}{2.26}{83.90}{92.53} & \textbothscript{\colortjhprc{83.42}}{3.72}{76.19}{90.66} & \textbothscript{\colortjhmae{4.43}}{0.23}{3.96}{4.86} & \textbothscript{\colortjhmse{31.35}}{3.69}{23.62}{37.72} & \textbothscript{\colortjhrmse{5.59}}{0.33}{4.86}{6.14} & \textbothscript{\colormimicoutroc{74.96}}{6.67}{59.76}{86.97} & \textbothscript{\colormimicoutprc{25.74}}{7.04}{13.82}{42.66} & \textbothscript{\colormimicreadroc{58.19}}{5.42}{46.47}{69.32} & \textbothscript{\colormimicreadprc{37.76}}{7.17}{22.78}{49.47} \\ 
 &  & full shot & \textbothscript{\colortjhroc{93.57}}{1.71}{90.45}{97.09} & \textbothscript{\colortjhprc{90.40}}{3.19}{84.61}{96.04} & \textbothscript{\colortjhmae{2.50}}{0.21}{2.14}{2.94} & \textbothscript{\colortjhmse{17.35}}{4.27}{9.51}{26.35} & \textbothscript{\colortjhrmse{4.13}}{0.51}{3.08}{5.13} & \textbothscript{\colormimicoutroc{92.49}}{3.03}{84.54}{97.17} & \textbothscript{\colormimicoutprc{72.05}}{7.58}{56.46}{86.06} & \textbothscript{\colormimicreadroc{81.30}}{3.81}{72.76}{88.39} & \textbothscript{\colormimicreadprc{63.70}}{6.56}{48.64}{75.53} \\ \cmidrule{2-12}
 & \multirow{2}{*}{LSTM} & 10 shot & \textbothscript{\colortjhroc{92.46}}{1.97}{88.64}{95.93} & \textbothscript{\colortjhprc{85.71}}{4.29}{77.08}{92.66} & \textbothscript{\colortjhmae{3.83}}{0.23}{3.38}{4.25} & \textbothscript{\colortjhmse{25.94}}{3.91}{18.91}{33.54} & \textbothscript{\colortjhrmse{5.08}}{0.38}{4.35}{5.79} & \textbothscript{\colormimicoutroc{56.42}}{7.93}{39.11}{72.11} & \textbothscript{\colormimicoutprc{16.52}}{5.56}{8.78}{29.08} & \textbothscript{\colormimicreadroc{50.98}}{5.70}{38.90}{61.46} & \textbothscript{\colormimicreadprc{35.39}}{5.89}{24.42}{46.71} \\ 
 &  & full shot & \textbothscript{\colortjhroc{92.98}}{1.91}{89.45}{96.17} & \textbothscript{\colortjhprc{86.97}}{4.12}{79.21}{93.33} & \textbothscript{\colortjhmae{2.64}}{0.21}{2.29}{3.07} & \textbothscript{\colortjhmse{17.76}}{4.21}{10.32}{26.55} & \textbothscript{\colortjhrmse{4.18}}{0.50}{3.21}{5.15} & \textbothscript{\colormimicoutroc{93.12}}{3.24}{87.39}{97.66} & \textbothscript{\colormimicoutprc{76.18}}{7.90}{58.08}{89.89} & \textbothscript{\textbf{\colormimicreadroc{82.52}}}{3.78}{75.04}{89.06} & \textbothscript{\colormimicreadprc{66.32}}{6.58}{51.28}{77.90} \\ \cmidrule{2-12}

 & \multirow{2}{*}{RNN} & 10 shot & \textbothscript{\colortjhroc{95.53}}{1.28}{92.74}{97.53} & \textbothscript{\colortjhprc{94.27}}{1.78}{90.50}{97.43} & \textbothscript{\colortjhmae{4.32}}{0.19}{4.01}{4.74} & \textbothscript{\colortjhmse{25.97}}{3.58}{20.02}{33.96} & \textbothscript{\colortjhrmse{5.08}}{0.35}{4.47}{5.83} & \textbothscript{\colormimicoutroc{62.03}}{7.97}{46.79}{75.84} & \textbothscript{\colormimicoutprc{26.07}}{8.62}{11.21}{42.05} & \textbothscript{\colormimicreadroc{60.80}}{5.22}{50.68}{69.42} & \textbothscript{\colormimicreadprc{44.51}}{6.86}{30.82}{56.46} \\ 
 &  & full shot & \textbothscript{\colortjhroc{96.42}}{1.10}{94.47}{98.38} & \textbothscript{\colortjhprc{95.18}}{1.62}{91.86}{98.23} & \textbothscript{\textbf{\colortjhmae{2.07}}}{0.24}{1.65}{2.52} & \textbothscript{\colortjhmse{16.64}}{5.00}{7.80}{26.85} & \textbothscript{\colortjhrmse{4.03}}{0.61}{2.79}{5.18} & \textbothscript{\colormimicoutroc{91.76}}{3.12}{86.10}{97.75} & \textbothscript{\colormimicoutprc{72.03}}{8.15}{57.09}{85.47} & \textbothscript{\colormimicreadroc{80.90}}{3.64}{72.24}{88.74} & \textbothscript{\colormimicreadprc{64.07}}{6.56}{51.71}{76.88} \\ \cmidrule{2-12}

 & \multirow{2}{*}{AdaCare} & 10 shot & \textbothscript{\colortjhroc{77.11}}{3.97}{69.50}{84.43} & \textbothscript{\colortjhprc{76.24}}{4.70}{68.16}{84.47} & \textbothscript{\colortjhmae{3.79}}{0.20}{3.59}{4.30} & \textbothscript{\colortjhmse{23.38}}{3.70}{17.20}{30.39} & \textbothscript{\colortjhrmse{4.82}}{0.38}{4.27}{5.60} & \textbothscript{\colormimicoutroc{80.02}}{6.32}{68.03}{92.62} & \textbothscript{\colormimicoutprc{54.93}}{9.85}{35.25}{75.52} & \textbothscript{\colormimicreadroc{64.39}}{4.76}{55.03}{72.69} & \textbothscript{\colormimicreadprc{42.47}}{6.43}{30.96}{55.65} \\ 
 &  & full shot & \textbothscript{\colortjhroc{99.02}}{0.46}{98.18}{99.90} & \textbothscript{\colortjhprc{98.86}}{0.53}{97.81}{99.86} & \textbothscript{\colortjhmae{2.49}}{0.24}{2.02}{2.99} & \textbothscript{\colortjhmse{18.12}}{4.63}{10.07}{27.10} & \textbothscript{\colortjhrmse{4.22}}{0.54}{3.17}{5.21} & \textbothscript{\textbf{\colormimicoutroc{94.28}}}{3.52}{87.36}{98.90} & \textbothscript{\textbf{\colormimicoutprc{81.93}}}{6.97}{68.90}{92.91} & \textbothscript{\colormimicreadroc{82.26}}{3.80}{74.93}{89.61} & \textbothscript{\textbf{\colormimicreadprc{68.82}}}{6.76}{53.32}{80.39} \\ \cmidrule{2-12}
 & \multirow{2}{*}{AICare} & 10 shot & \textbothscript{\colortjhroc{86.79}}{2.41}{82.23}{91.49} & \textbothscript{\colortjhprc{82.98}}{3.96}{74.01}{90.48} & \textbothscript{\colortjhmae{3.24}}{0.22}{2.84}{3.72} & \textbothscript{\colortjhmse{20.35}}{4.43}{13.04}{29.71} & \textbothscript{\colortjhrmse{4.48}}{0.49}{3.61}{5.45} & \textbothscript{\colormimicoutroc{60.87}}{6.25}{50.66}{72.76} & \textbothscript{\colormimicoutprc{23.69}}{8.14}{8.38}{38.67} & \textbothscript{\colormimicreadroc{66.32}}{4.55}{56.63}{74.18} & \textbothscript{\colormimicreadprc{43.76}}{7.02}{30.23}{56.03} \\ 
 &  & full shot & \textbothscript{\colortjhroc{95.97}}{1.31}{94.12}{98.00} & \textbothscript{\colortjhprc{94.56}}{2.02}{92.59}{98.71} & \textbothscript{\colortjhmae{2.17}}{0.23}{1.90}{2.78} & \textbothscript{\textbf{\colortjhmse{16.28}}}{4.84}{8.41}{26.58} & \textbothscript{\textbf{\colortjhrmse{3.99}}}{0.60}{2.90}{5.16} & \textbothscript{\colormimicoutroc{92.89}}{3.66}{84.49}{97.89} & \textbothscript{\colormimicoutprc{77.84}}{7.10}{62.84}{90.80} & \textbothscript{\colormimicreadroc{80.20}}{4.23}{72.62}{88.94} & \textbothscript{\colormimicreadprc{65.50}}{6.72}{52.16}{77.46} \\ \cmidrule{2-12}
 & \multirow{2}{*}{ConCare} & 10 shot & \textbothscript{\colortjhroc{90.98}}{1.97}{87.14}{94.20} & \textbothscript{\colortjhprc{91.43}}{2.12}{87.26}{94.66} & \textbothscript{\colortjhmae{4.01}}{0.29}{3.49}{4.71} & \textbothscript{\colortjhmse{30.03}}{6.41}{19.20}{45.39} & \textbothscript{\colortjhrmse{5.45}}{0.58}{4.38}{6.74} & \textbothscript{\colormimicoutroc{72.58}}{6.42}{60.50}{84.35} & \textbothscript{\colormimicoutprc{30.73}}{8.44}{16.93}{47.61} & \textbothscript{\colormimicreadroc{70.30}}{4.62}{60.91}{79.64} & \textbothscript{\colormimicreadprc{48.68}}{7.41}{32.68}{61.04} \\ 
 &  & full shot & \textbothscript{\colortjhroc{91.00}}{2.14}{86.93}{94.86} & \textbothscript{\colortjhprc{91.72}}{2.30}{87.22}{95.30} & \textbothscript{\colortjhmae{2.32}}{0.24}{1.96}{2.77} & \textbothscript{\colortjhmse{17.94}}{4.99}{9.62}{27.49} & \textbothscript{\colortjhrmse{4.19}}{0.59}{3.10}{5.24} & \textbothscript{\colormimicoutroc{94.08}}{3.70}{87.49}{99.27} & \textbothscript{\colormimicoutprc{80.65}}{6.98}{64.35}{92.09} & \textbothscript{\colormimicreadroc{79.17}}{4.42}{72.19}{88.72} & \textbothscript{\colormimicreadprc{64.27}}{6.97}{52.25}{77.17} \\ \cmidrule{2-12}
 & \multirow{2}{*}{GRASP} & 10 shot & \textbothscript{\colortjhroc{87.25}}{2.53}{83.27}{92.32} & \textbothscript{\colortjhprc{84.32}}{3.56}{78.48}{91.24} & \textbothscript{\colortjhmae{5.81}}{0.18}{5.07}{5.96} & \textbothscript{\colortjhmse{41.25}}{2.96}{34.85}{47.42} & \textbothscript{\colortjhrmse{6.42}}{0.23}{5.90}{6.89} & \textbothscript{\colormimicoutroc{69.89}}{8.80}{49.99}{82.96} & \textbothscript{\colormimicoutprc{45.96}}{10.33}{26.79}{65.79} & \textbothscript{\colormimicreadroc{53.76}}{5.54}{41.85}{62.36} & \textbothscript{\colormimicreadprc{36.35}}{6.24}{24.60}{47.37} \\ 
 &  & full shot & \textbothscript{\colortjhroc{94.25}}{1.58}{91.05}{97.13} & \textbothscript{\colortjhprc{92.03}}{2.54}{86.62}{96.31} & \textbothscript{\colortjhmae{3.84}}{0.20}{3.44}{4.14} & \textbothscript{\colortjhmse{23.89}}{3.07}{15.64}{28.21} & \textbothscript{\colortjhrmse{4.88}}{0.31}{3.96}{5.31} & \textbothscript{\colormimicoutroc{93.14}}{3.03}{87.25}{97.86} & \textbothscript{\colormimicoutprc{72.55}}{8.36}{53.53}{86.06} & \textbothscript{\colormimicreadroc{77.76}}{4.17}{68.79}{84.87} & \textbothscript{\colormimicreadprc{62.42}}{7.08}{47.68}{73.98} \\ \midrule

 \multirow{15}{*}{Base LLM} & \multirow{3}{*}{OpenBioLLM-8B} & base prompt & \textbothscript{\colortjhroc{49.37}}{2.70}{45.31}{54.65} & \textbothscript{\colortjhprc{46.17}}{4.40}{37.39}{54.21} & \textbothscript{\colortjhmae{6.50}}{0.32}{5.86}{7.03} & \textbothscript{\colortjhmse{61.41}}{6.84}{48.81}{74.11} & \textbothscript{\colortjhrmse{7.82}}{0.44}{6.99}{8.61} & \textbothscript{\colormimicoutroc{52.35}}{4.65}{41.60}{59.99} & \textbothscript{\colormimicoutprc{10.31}}{2.28}{6.24}{14.36} & \textbothscript{\colormimicreadroc{55.58}}{4.25}{47.97}{63.83} & \textbothscript{\colormimicreadprc{35.83}}{4.45}{28.13}{42.89} \\ 
 &  & optimized prompt & \textbothscript{\colortjhroc{48.80}}{3.00}{42.80}{54.17} & \textbothscript{\colortjhprc{45.15}}{4.15}{36.66}{54.33} & \textbothscript{\colortjhmae{5.86}}{0.34}{5.10}{6.48} & \textbothscript{\colortjhmse{54.83}}{6.19}{43.00}{66.38} & \textbothscript{\colortjhrmse{7.39}}{0.42}{6.56}{8.15} & \textbothscript{\colormimicoutroc{56.45}}{5.81}{45.33}{66.16} & \textbothscript{\colormimicoutprc{16.23}}{5.61}{7.37}{28.32} & \textbothscript{\colormimicreadroc{46.86}}{3.55}{40.66}{53.63} & \textbothscript{\colormimicreadprc{21.34}}{3.15}{14.96}{27.42} \\ 
 &  & opt.+ICL & \textbothscript{\colortjhroc{56.75}}{3.92}{48.51}{62.78} & \textbothscript{\colortjhprc{49.76}}{4.67}{40.22}{59.44} & \textbothscript{\colortjhmae{8.97}}{0.36}{8.19}{9.61} & \textbothscript{\colortjhmse{105.23}}{6.52}{93.34}{119.17} & \textbothscript{\colortjhrmse{10.25}}{0.32}{9.66}{10.92} & \textbothscript{\colormimicoutroc{58.69}}{6.06}{46.38}{68.91} & \textbothscript{\colormimicoutprc{12.85}}{3.77}{6.97}{21.81} & \textbothscript{\colormimicreadroc{50.21}}{4.97}{40.27}{60.94} & \textbothscript{\colormimicreadprc{24.23}}{4.02}{17.33}{31.07} \\ \cmidrule{2-12}
 & \multirow{3}{*}{Qwen2.5-7B} & base prompt & \textbothscript{\colortjhroc{72.96}}{4.00}{65.81}{79.92} & \textbothscript{\colortjhprc{64.49}}{5.06}{52.82}{73.50} & \textbothscript{\colortjhmae{12.44}}{0.43}{11.75}{13.39} & \textbothscript{\colortjhmse{184.73}}{12.89}{164.20}{211.25} & \textbothscript{\colortjhrmse{13.58}}{0.47}{12.85}{14.57} & \textbothscript{\colormimicoutroc{70.68}}{5.15}{60.61}{79.22} & \textbothscript{\colormimicoutprc{17.55}}{4.54}{10.00}{26.13} & \textbothscript{\colormimicreadroc{67.39}}{8.74}{52.28}{83.12} & \textbothscript{\colormimicreadprc{41.26}}{11.07}{20.46}{60.93} \\ 
 &  & optimized prompt & \textbothscript{\colortjhroc{78.28}}{3.25}{73.16}{85.05} & \textbothscript{\colortjhprc{67.10}}{4.70}{58.57}{77.44} & \textbothscript{\colortjhmae{13.59}}{0.53}{12.59}{14.54} & \textbothscript{\colortjhmse{237.29}}{21.79}{198.27}{280.56} & \textbothscript{\colortjhrmse{15.39}}{0.70}{14.08}{16.75} & \textbothscript{\colormimicoutroc{68.20}}{6.84}{56.55}{80.94} & \textbothscript{\colormimicoutprc{21.38}}{6.14}{10.45}{34.33} & \textbothscript{\colormimicreadroc{51.65}}{4.90}{42.38}{61.40} & \textbothscript{\colormimicreadprc{25.49}}{4.30}{17.13}{34.03} \\ 
 &  & opt.+ICL & \textbothscript{\colortjhroc{79.83}}{2.68}{75.43}{84.92} & \textbothscript{\colortjhprc{70.87}}{4.61}{62.77}{80.11} & \textbothscript{\colortjhmae{15.34}}{0.71}{14.03}{16.56} & \textbothscript{\colortjhmse{313.70}}{27.84}{269.25}{373.57} & \textbothscript{\colortjhrmse{17.69}}{0.79}{16.41}{19.33} & \textbothscript{\colormimicoutroc{61.57}}{7.12}{47.34}{74.87} & \textbothscript{\colormimicoutprc{13.58}}{3.17}{7.83}{19.49} & \textbothscript{\colormimicreadroc{55.86}}{3.98}{48.11}{63.76} & \textbothscript{\colormimicreadprc{25.32}}{4.02}{17.12}{32.58} \\ \cmidrule{2-12}
 & \multirow{3}{*}{Gemma-3-4B} & base prompt & \textbothscript{\colortjhroc{65.64}}{3.53}{58.80}{73.49} & \textbothscript{\colortjhprc{64.83}}{4.68}{55.23}{74.46} & \textbothscript{\colortjhmae{13.25}}{0.35}{12.48}{13.91} & \textbothscript{\colortjhmse{199.67}}{9.20}{179.56}{216.22} & \textbothscript{\colortjhrmse{14.13}}{0.33}{13.40}{14.70} & \textbothscript{\colormimicoutroc{63.46}}{7.68}{50.84}{78.52} & \textbothscript{\colormimicoutprc{17.76}}{5.71}{7.55}{28.51} & \textbothscript{\colormimicreadroc{63.60}}{7.81}{47.37}{77.21} & \textbothscript{\colormimicreadprc{32.38}}{8.19}{18.38}{47.65} \\ 
 &  & optimized prompt & \textbothscript{\colortjhroc{70.97}}{3.21}{63.29}{76.59} & \textbothscript{\colortjhprc{66.54}}{4.33}{58.64}{74.89} & \textbothscript{\colortjhmae{14.24}}{0.37}{13.51}{14.90} & \textbothscript{\colortjhmse{228.07}}{10.44}{208.00}{245.64} & \textbothscript{\colortjhrmse{15.10}}{0.35}{14.42}{15.67} & \textbothscript{\colormimicoutroc{61.16}}{6.62}{46.99}{73.50} & \textbothscript{\colormimicoutprc{13.18}}{3.60}{6.44}{20.53} & \textbothscript{\colormimicreadroc{63.57}}{3.90}{55.32}{72.12} & \textbothscript{\colormimicreadprc{30.00}}{4.64}{20.55}{39.42} \\ 
 &  & opt.+ICL & \textbothscript{\colortjhroc{76.01}}{3.46}{69.22}{82.49} & \textbothscript{\colortjhprc{71.62}}{4.97}{61.03}{80.13} & \textbothscript{\colortjhmae{14.87}}{0.39}{14.23}{15.49} & \textbothscript{\colortjhmse{250.80}}{10.36}{234.69}{267.30} & \textbothscript{\colortjhrmse{15.83}}{0.33}{15.32}{16.35} & \textbothscript{\colormimicoutroc{57.78}}{7.40}{43.85}{74.09} & \textbothscript{\colormimicoutprc{15.16}}{5.11}{7.63}{28.74} & \textbothscript{\colormimicreadroc{60.02}}{4.23}{52.96}{67.34} & \textbothscript{\colormimicreadprc{29.05}}{5.37}{18.73}{39.33} \\ \cmidrule{2-12}
 & \multirow{3}{*}{DeepSeek-V3.1} & base prompt & \textbothscript{\colortjhroc{88.63}}{2.41}{83.56}{93.38} & \textbothscript{\colortjhprc{80.65}}{4.11}{73.07}{88.64} & \textbothscript{\colortjhmae{16.53}}{0.51}{15.53}{17.67} & \textbothscript{\colortjhmse{320.66}}{16.80}{292.01}{360.04} & \textbothscript{\colortjhrmse{17.90}}{0.47}{17.09}{18.97} & \textbothscript{\colormimicoutroc{79.52}}{5.87}{66.94}{89.69} & \textbothscript{\colormimicoutprc{35.35}}{9.46}{15.86}{50.63} & \textbothscript{\colormimicreadroc{59.63}}{4.60}{51.09}{68.36} & \textbothscript{\colormimicreadprc{30.09}}{4.73}{21.92}{39.17} \\ 
 &  & optimized prompt & \textbothscript{\colortjhroc{85.84}}{2.39}{81.15}{90.53} & \textbothscript{\colortjhprc{75.66}}{4.22}{67.76}{84.22} & \textbothscript{\colortjhmae{17.36}}{0.46}{16.53}{18.25} & \textbothscript{\colortjhmse{342.88}}{15.72}{312.02}{374.44} & \textbothscript{\colortjhrmse{18.51}}{0.42}{17.66}{19.35} & \textbothscript{\colormimicoutroc{76.05}}{6.90}{61.38}{87.61} & \textbothscript{\colormimicoutprc{36.07}}{9.50}{17.72}{54.69} & \textbothscript{\colormimicreadroc{63.90}}{4.04}{56.59}{71.11} & \textbothscript{\colormimicreadprc{33.84}}{5.71}{24.27}{43.69} \\ 
 &  & opt.+ICL & \textbothscript{\colortjhroc{85.87}}{2.70}{80.83}{91.10} & \textbothscript{\colortjhprc{75.96}}{4.32}{68.15}{83.45} & \textbothscript{\colortjhmae{18.74}}{0.45}{17.79}{19.52} & \textbothscript{\colortjhmse{387.27}}{16.23}{355.98}{416.66} & \textbothscript{\colortjhrmse{19.67}}{0.41}{18.87}{20.41} & \textbothscript{\colormimicoutroc{73.66}}{7.55}{58.62}{86.70} & \textbothscript{\colormimicoutprc{33.86}}{8.95}{15.28}{48.10} & \textbothscript{\colormimicreadroc{63.61}}{4.33}{55.10}{72.41} & \textbothscript{\colormimicreadprc{32.64}}{5.58}{22.40}{41.47} \\ \cmidrule{2-12}
 & \multirow{3}{*}{GPT-4o} & base prompt & \textbothscript{\colortjhroc{96.76}}{1.30}{94.05}{98.80} & \textbothscript{\colortjhprc{93.90}}{2.55}{88.70}{98.00} & \textbothscript{\colortjhmae{14.56}}{0.47}{13.60}{15.53} & \textbothscript{\colortjhmse{258.23}}{14.89}{231.15}{294.59} & \textbothscript{\colortjhrmse{16.06}}{0.46}{15.20}{17.16} & \textbothscript{\colormimicoutroc{76.96}}{6.59}{65.04}{89.73} & \textbothscript{\colormimicoutprc{33.90}}{9.17}{17.97}{50.42} & \textbothscript{\colormimicreadroc{59.16}}{5.45}{47.26}{68.76} & \textbothscript{\colormimicreadprc{28.70}}{5.08}{18.74}{38.12} \\ 
 &  & optimized prompt & \textbothscript{\colortjhroc{92.73}}{1.72}{89.46}{96.13} & \textbothscript{\colortjhprc{88.44}}{2.70}{83.49}{93.27} & \textbothscript{\colortjhmae{15.62}}{0.45}{14.84}{16.51} & \textbothscript{\colortjhmse{290.26}}{14.97}{263.23}{320.58} & \textbothscript{\colortjhrmse{17.03}}{0.44}{16.22}{17.90} & \textbothscript{\colormimicoutroc{76.18}}{6.41}{63.56}{86.96} & \textbothscript{\colormimicoutprc{29.91}}{9.01}{15.25}{47.44} & \textbothscript{\colormimicreadroc{61.30}}{5.20}{50.31}{70.09} & \textbothscript{\colormimicreadprc{32.22}}{5.55}{21.06}{41.89} \\ 
 &  & opt.+ICL & \textbothscript{\colortjhroc{95.72}}{1.21}{93.27}{97.75} & \textbothscript{\colortjhprc{93.04}}{2.08}{88.82}{96.55} & \textbothscript{\colortjhmae{17.93}}{0.58}{17.08}{18.96} & \textbothscript{\colortjhmse{378.94}}{21.17}{338.34}{417.32} & \textbothscript{\colortjhrmse{19.46}}{0.55}{18.45}{20.45} & \textbothscript{\colormimicoutroc{85.99}}{3.85}{78.13}{92.07} & \textbothscript{\colormimicoutprc{42.20}}{9.92}{24.74}{60.81} & \textbothscript{\colormimicreadroc{62.72}}{4.87}{52.04}{71.98} & \textbothscript{\colormimicreadprc{34.43}}{5.73}{23.71}{44.58} \\ \midrule
 
{\multirow{18}{*}{Reasoning LLM}} & \multirow{3}{*}{HuatuoGPT-o1-7B} & base prompt & \textbothscript{\colortjhroc{77.74}}{3.34}{70.82}{83.68} & \textbothscript{\colortjhprc{71.89}}{4.83}{62.64}{81.69} & \textbothscript{\colortjhmae{8.87}}{0.26}{8.41}{9.47} & \textbothscript{\colortjhmse{98.61}}{5.83}{89.18}{110.83} & \textbothscript{\colortjhrmse{9.93}}{0.29}{9.55}{10.62} & \textbothscript{\colormimicoutroc{73.20}}{6.12}{60.35}{84.94} & \textbothscript{\colormimicoutprc{22.32}}{6.87}{9.89}{37.89} & \textbothscript{\colormimicreadroc{54.31}}{19.32}{21.41}{89.17} & \textbothscript{\colormimicreadprc{32.78}}{15.90}{8.36}{60.57} \\ 
 &  & optimized prompt & \textbothscript{\colortjhroc{75.64}}{3.43}{68.75}{81.73} & \textbothscript{\colortjhprc{66.20}}{4.79}{56.99}{74.46} & \textbothscript{\colortjhmae{9.57}}{0.34}{8.86}{10.21} & \textbothscript{\colortjhmse{114.03}}{7.10}{98.29}{127.61} & \textbothscript{\colortjhrmse{10.67}}{0.33}{9.91}{11.30} & \textbothscript{\colormimicoutroc{66.07}}{6.29}{53.21}{75.78} & \textbothscript{\colormimicoutprc{15.28}}{4.18}{8.79}{23.39} & \textbothscript{\colormimicreadroc{63.17}}{5.06}{53.20}{72.16} & \textbothscript{\colormimicreadprc{34.60}}{6.32}{22.73}{46.52} \\ 
 &  & opt.+ICL & \textbothscript{\colortjhroc{85.34}}{2.61}{79.68}{91.52} & \textbothscript{\colortjhprc{77.31}}{4.26}{69.50}{85.81} & \textbothscript{\colortjhmae{11.12}}{0.42}{10.47}{12.12} & \textbothscript{\colortjhmse{157.38}}{11.66}{138.05}{183.69} & \textbothscript{\colortjhrmse{12.54}}{0.47}{11.75}{13.55} & \textbothscript{\colormimicoutroc{70.39}}{7.60}{51.48}{80.40} & \textbothscript{\colormimicoutprc{20.33}}{5.51}{10.52}{31.29} & \textbothscript{\colormimicreadroc{50.54}}{4.88}{41.45}{58.92} & \textbothscript{\colormimicreadprc{24.30}}{4.22}{15.75}{31.96} \\ \cmidrule{2-12}
 & \multirow{3}{*}{DeepSeek-R1-7B} & base prompt & \textbothscript{\colortjhroc{53.59}}{3.88}{46.61}{61.13} & \textbothscript{\colortjhprc{49.06}}{4.05}{41.77}{57.93} & \textbothscript{\colortjhmae{5.24}}{0.22}{4.71}{5.86} & \textbothscript{\colortjhmse{37.66}}{4.89}{29.20}{45.64} & \textbothscript{\colortjhrmse{6.12}}{0.39}{5.26}{6.97} & \textbothscript{\colormimicoutroc{53.59}}{5.26}{44.31}{65.10} & \textbothscript{\colormimicoutprc{10.53}}{2.75}{5.90}{16.59} & \textbothscript{\colormimicreadroc{42.25}}{21.66}{6.01}{85.76} & \textbothscript{\colormimicreadprc{25.83}}{11.10}{8.95}{50.28} \\
 &  & optimized prompt & \textbothscript{\colortjhroc{58.84}}{3.16}{53.00}{65.88} & \textbothscript{\colortjhprc{51.71}}{4.49}{43.77}{61.21} & \textbothscript{\colortjhmae{6.09}}{0.30}{5.52}{6.63} & \textbothscript{\colortjhmse{53.01}}{6.20}{42.35}{64.09} & \textbothscript{\colortjhrmse{7.27}}{0.42}{6.51}{8.01} & \textbothscript{\colormimicoutroc{51.97}}{4.81}{42.55}{61.41} & \textbothscript{\colormimicoutprc{10.37}}{2.38}{6.30}{15.41} & \textbothscript{\colormimicreadroc{55.74}}{4.39}{49.19}{65.11} & \textbothscript{\colormimicreadprc{26.13}}{4.32}{18.63}{35.49} \\ 
 &  & opt.+ICL & \textbothscript{\colortjhroc{52.70}}{1.95}{48.73}{56.67} & \textbothscript{\colortjhprc{47.89}}{4.09}{38.25}{57.33} & \textbothscript{\colortjhmae{5.59}}{0.24}{5.18}{6.29} & \textbothscript{\colortjhmse{44.07}}{4.66}{35.87}{54.04} & \textbothscript{\colortjhrmse{6.63}}{0.35}{5.53}{7.81} & \textbothscript{\colormimicoutroc{40.94}}{3.97}{33.75}{49.82} & \textbothscript{\colormimicoutprc{9.43}}{2.27}{5.90}{13.68} & \textbothscript{\colormimicreadroc{53.19}}{4.13}{45.98}{61.23} & \textbothscript{\colormimicreadprc{24.53}}{3.66}{19.05}{31.79} \\ \cmidrule{2-12}
 & \multirow{3}{*}{DeepSeek-R1} & base prompt & \textbothscript{\colortjhroc{94.05}}{1.58}{91.22}{96.87} & \textbothscript{\colortjhprc{89.40}}{3.66}{81.61}{94.92} & \textbothscript{\colortjhmae{15.23}}{0.49}{14.18}{16.15} & \textbothscript{\colortjhmse{280.00}}{14.16}{250.71}{301.63} & \textbothscript{\colortjhrmse{16.73}}{0.43}{15.83}{17.37} & \textbothscript{\colormimicoutroc{76.01}}{6.97}{60.48}{89.51} & \textbothscript{\colormimicoutprc{38.84}}{9.97}{17.73}{56.21} & \textbothscript{\colormimicreadroc{62.40}}{5.24}{52.94}{72.00} & \textbothscript{\colormimicreadprc{31.92}}{5.89}{20.67}{43.13} \\ 
 &  & optimized prompt & \textbothscript{\colortjhroc{88.31}}{2.20}{84.34}{92.22} & \textbothscript{\colortjhprc{82.02}}{3.78}{75.02}{89.16} & \textbothscript{\colortjhmae{15.97}}{0.46}{15.03}{16.73} & \textbothscript{\colortjhmse{298.13}}{13.81}{271.61}{319.96} & \textbothscript{\colortjhrmse{17.26}}{0.40}{16.48}{17.89} & \textbothscript{\colormimicoutroc{76.34}}{5.82}{62.88}{86.79} & \textbothscript{\colormimicoutprc{36.37}}{9.72}{17.15}{55.76} & \textbothscript{\colormimicreadroc{58.26}}{4.58}{51.11}{67.77} & \textbothscript{\colormimicreadprc{28.11}}{4.57}{19.76}{36.74} \\ 
 &  & opt.+ICL & \textbothscript{\colortjhroc{90.61}}{1.81}{87.73}{94.67} & \textbothscript{\colortjhprc{85.20}}{3.33}{78.79}{91.35} & \textbothscript{\colortjhmae{16.03}}{0.52}{15.02}{16.85} & \textbothscript{\colortjhmse{300.42}}{15.03}{273.38}{325.36} & \textbothscript{\colortjhrmse{17.33}}{0.43}{16.53}{18.04} & \textbothscript{\colormimicoutroc{75.09}}{5.99}{64.88}{85.51} & \textbothscript{\colormimicoutprc{27.12}}{8.39}{12.36}{44.87} & \textbothscript{\colormimicreadroc{61.15}}{4.61}{52.39}{68.79} & \textbothscript{\colormimicreadprc{30.60}}{6.01}{20.78}{44.17} \\ \cmidrule{2-12}
 & \multirow{3}{*}{DeepSeek-V3.1-Think} & base prompt & \textbothscript{\colortjhroc{82.47}}{2.80}{76.73}{87.34} & \textbothscript{\colortjhprc{70.69}}{4.41}{61.66}{78.85} & \textbothscript{\colortjhmae{14.66}}{0.40}{13.91}{15.37} & \textbothscript{\colortjhmse{246.74}}{11.08}{225.40}{267.14} & \textbothscript{\colortjhrmse{15.70}}{0.35}{15.01}{16.34} & \textbothscript{\colormimicoutroc{84.93}}{5.74}{72.11}{93.74} & \textbothscript{\colormimicoutprc{56.44}}{9.89}{34.64}{73.20} & \textbothscript{\colormimicreadroc{67.34}}{4.64}{58.93}{75.89} & \textbothscript{\colormimicreadprc{36.79}}{6.44}{25.27}{48.91} \\ 
 &  & optimized prompt & \textbothscript{\colortjhroc{82.25}}{3.00}{75.20}{87.59} & \textbothscript{\colortjhprc{72.70}}{4.19}{64.88}{80.47} & \textbothscript{\colortjhmae{15.30}}{0.50}{14.27}{16.26} & \textbothscript{\colortjhmse{273.78}}{14.18}{247.14}{301.38} & \textbothscript{\colortjhrmse{16.54}}{0.43}{15.72}{17.36} & \textbothscript{\colormimicoutroc{79.41}}{5.80}{67.91}{88.61} & \textbothscript{\colormimicoutprc{39.64}}{11.01}{15.96}{59.54} & \textbothscript{\colormimicreadroc{60.98}}{5.39}{52.07}{72.24} & \textbothscript{\colormimicreadprc{33.35}}{6.32}{22.49}{46.01} \\ 
 &  & opt.+ICL & \textbothscript{\colortjhroc{84.08}}{2.69}{78.42}{88.95} & \textbothscript{\colortjhprc{75.66}}{4.16}{67.73}{82.74} & \textbothscript{\colortjhmae{16.44}}{0.42}{15.61}{17.13} & \textbothscript{\colortjhmse{301.33}}{13.82}{275.59}{325.16} & \textbothscript{\colortjhrmse{17.35}}{0.40}{16.60}{18.03} & \textbothscript{\colormimicoutroc{83.33}}{6.38}{70.32}{94.57} & \textbothscript{\colormimicoutprc{52.71}}{11.36}{32.05}{73.37} & \textbothscript{\colormimicreadroc{69.35}}{4.58}{58.76}{77.40} & \textbothscript{\colormimicreadprc{38.97}}{6.80}{27.90}{53.30} \\ \cmidrule{2-12}
 & \multirow{3}{*}{o3-mini-high} & base prompt & \textbothscript{\colortjhroc{85.43}}{2.15}{81.86}{89.29} & \textbothscript{\colortjhprc{76.52}}{3.91}{70.30}{84.15} & \textbothscript{\colortjhmae{13.44}}{0.46}{12.83}{14.61} & \textbothscript{\colortjhmse{217.77}}{15.54}{201.41}{257.06} & \textbothscript{\colortjhrmse{14.75}}{0.52}{14.19}{16.03} & \textbothscript{\colormimicoutroc{71.13}}{5.41}{61.36}{81.25} & \textbothscript{\colormimicoutprc{18.35}}{5.02}{10.28}{27.71} & \textbothscript{\colormimicreadroc{62.89}}{4.72}{55.15}{72.86} & \textbothscript{\colormimicreadprc{33.69}}{5.82}{21.90}{43.54} \\ 
 &  & optimized prompt & \textbothscript{\colortjhroc{82.49}}{2.64}{76.95}{86.75} & \textbothscript{\colortjhprc{69.95}}{4.41}{61.26}{77.10} & \textbothscript{\colortjhmae{16.35}}{0.47}{15.38}{17.15} & \textbothscript{\colortjhmse{313.98}}{16.47}{282.50}{344.78} & \textbothscript{\colortjhrmse{17.71}}{0.46}{16.81}{18.57} & \textbothscript{\colormimicoutroc{76.11}}{6.34}{62.90}{87.50} & \textbothscript{\colormimicoutprc{41.72}}{10.59}{22.37}{59.52} & \textbothscript{\colormimicreadroc{62.21}}{4.93}{53.81}{72.70} & \textbothscript{\colormimicreadprc{38.86}}{6.34}{26.98}{52.32} \\ 
 &  & opt.+ICL & \textbothscript{\colortjhroc{84.42}}{2.52}{78.78}{89.67} & \textbothscript{\colortjhprc{75.65}}{4.48}{67.60}{82.35} & \textbothscript{\colortjhmae{15.63}}{0.40}{14.95}{16.44} & \textbothscript{\colortjhmse{276.67}}{13.52}{253.14}{302.90} & \textbothscript{\colortjhrmse{16.63}}{0.41}{15.91}{17.40} & \textbothscript{\colormimicoutroc{71.23}}{7.19}{55.46}{83.34} & \textbothscript{\colormimicoutprc{28.99}}{7.88}{12.20}{43.10} & \textbothscript{\colormimicreadroc{63.30}}{4.85}{53.88}{72.38} & \textbothscript{\colormimicreadprc{36.13}}{6.18}{24.27}{46.56} \\ \cmidrule{2-12}
 & \multirow{3}{*}{GPT-5} & base prompt & \textbothscript{\colortjhroc{93.98}}{2.06}{88.97}{97.52} & \textbothscript{\colortjhprc{86.73}}{4.39}{77.97}{94.71} & \textbothscript{\colortjhmae{15.87}}{0.39}{15.14}{16.62} & \textbothscript{\colortjhmse{285.72}}{12.69}{262.20}{311.22} & \textbothscript{\colortjhrmse{16.90}}{0.38}{16.19}{17.64} & \textbothscript{\colormimicoutroc{80.82}}{5.20}{71.33}{90.97} & \textbothscript{\colormimicoutprc{34.34}}{9.04}{17.15}{51.90} & \textbothscript{\colormimicreadroc{65.74}}{5.01}{55.34}{74.92} & \textbothscript{\colormimicreadprc{34.05}}{6.11}{22.78}{45.40} \\ 
 &  & optimized prompt & \textbothscript{\colortjhroc{83.65}}{2.82}{77.37}{89.07} & \textbothscript{\colortjhprc{71.79}}{4.49}{64.18}{79.55} & \textbothscript{\colortjhmae{16.17}}{0.43}{15.35}{16.91} & \textbothscript{\colortjhmse{297.86}}{13.92}{268.69}{321.90} & \textbothscript{\colortjhrmse{17.25}}{0.40}{16.39}{17.94} & \textbothscript{\colormimicoutroc{80.50}}{5.98}{68.64}{91.07} & \textbothscript{\colormimicoutprc{38.66}}{10.51}{18.46}{60.30} & \textbothscript{\colormimicreadroc{65.64}}{4.39}{57.94}{73.66} & \textbothscript{\colormimicreadprc{32.59}}{5.54}{22.92}{42.66} \\ 
 &  & opt.+ICL & \textbothscript{\colortjhroc{85.82}}{2.57}{80.35}{89.51} & \textbothscript{\colortjhprc{75.53}}{4.33}{66.53}{82.52} & \textbothscript{\colortjhmae{17.93}}{0.45}{17.06}{18.77} & \textbothscript{\colortjhmse{360.77}}{16.33}{331.24}{397.59} & \textbothscript{\colortjhrmse{18.99}}{0.43}{18.20}{19.94} & \textbothscript{\colormimicoutroc{81.25}}{5.52}{70.98}{90.55} & \textbothscript{\colormimicoutprc{34.33}}{9.21}{18.44}{53.93} & \textbothscript{\colormimicreadroc{65.65}}{5.13}{55.89}{74.54} & \textbothscript{\colormimicreadprc{34.48}}{6.44}{23.71}{48.57} \\ 
\bottomrule
\end{tabular}
}
\end{table*}

Our benchmarking reveals that state-of-the-art LLMs, notably GPT-5 and DeepSeek-V3.1-Think, are capable of performing zero-shot clinical predictions directly from structured EHR data (Table~\ref{tab:ehr_results}). However, their performance in this zero-shot setting generally remains below that of conventional machine learning (ML) and deep learning (DL) models trained on the complete dataset (full shot). Specialized DL models like AdaCare and AICare, when fully trained, consistently achieved the highest performance across most tasks on both datasets.

A key finding emerges when comparing zero-shot LLMs to traditional models trained with limited data (10 shot). On the more complex MIMIC-IV dataset, top-performing LLMs demonstrated remarkable strength. Specifically, GPT-4o using an optimized prompt with in-context learning (opt.+ICL) achieved a zero-shot AUROC of 85.99\% for mortality prediction, surpassing all evaluated 10-shot conventional models, including the best performing 10-shot model, AdaCare (80.02\% AUROC). This highlights the potent data-efficiency of leading LLMs for certain clinical prediction tasks. Conversely, on the TJH dataset tasks, even 10-shot conventional models (e.g., CatBoost achieving 99.43\% AUROC for outcome prediction) reached near-perfect performance, there was less room for zero-shot LLMs to demonstrate an advantage. This performance disparity is likely attributable to inherent differences in task complexity and data heterogeneity, a point analyzed further in Appendix~\ref{sec:appendix_disparities}.

We evaluated LLMs using three prompt strategies: a basic sequential feature input (base prompt), an optimized prompt with better formatting and instructions, and an optimized prompt with in-context learning (opt.+ICL). The impact of these strategies was highly variable, depending on the model, task, and dataset complexity. For instance, on the challenging MIMIC-IV mortality task, adding in-context learning significantly boosted GPT-4o's performance from 76.18\% AUROC (optimized) to 85.99\% (opt.+ICL). In contrast, for some large-scale models, increased prompt complexity occasionally led to diminished performance; GPT-5's AUROC on the TJH outcome task, for example, decreased when moving from a base to an optimized prompt. While not always improving accuracy metrics, prompt engineering was critical for model reliability. As detailed in our failure analysis (Appendix Section~\ref{sec:failure}), using an optimized prompt often dramatically reduced the ``failure-to-predict'' rate for models like HuatuoGPT-o1-7B and Gemma-3-4B, ensuring they produced a valid output.

Among the LLMs tested, GPT-5 and the large-scale DeepSeek-V3.1-Think consistently demonstrated superior zero-shot classification capabilities compared to smaller models like Qwen2.5-7B, Gemma-3-4B, and specialized models like HuatuoGPT-o1-7B. On the MIMIC-IV dataset, the open-source DeepSeek-V3.1-Think achieved an AUROC equivalent to or even higher than that of GPT-5. Similar to our note-based experiments, we encountered challenges with prompt adherence for several models, particularly those finetuned on specialized corpora, which often deviated from the requested output format and were subsequently removed from the table.

While a performance gap persists between zero-shot LLMs and fully trained conventional models, the strong few-shot and zero-shot performance of models like DeepSeek underscores their potential value in data-scarce clinical scenarios, such as predicting outcomes for emerging diseases or in settings with limited historical data. 

\subsection{Benchmarking Results of Multimodal Prediction Tasks}
Our multimodal experiments, which integrate both structured EHR data and unstructured clinical notes, yield a complex and nuanced picture of performance, as detailed in Table~\ref{tab:mm_results}. The results reveal that combining data modalities does not uniformly lead to improvements, with outcomes dependent on the integration method, the clinical task, and the inherent predictive power of each data source. For finetuned models, we use the best-performing unimodal encoders identified in our main experiments:
\begin{itemize}
    \item For the \textbf{in-hospital mortality task}, based on their strong unimodal performance, we selected AdaCare as the encoder for structured EHR data and a finetuned GatorTron for clinical notes.
    \item For the \textbf{30-day readmission task}, we selected LSTM for structured EHR data and a finetuned HuatuoGPT-o1-7B for clinical notes.
\end{itemize}

\begin{table*}[!ht]
\centering
\caption{\textit{Performance of mortality prediction and MIMIC-IV datasets using multimodal EHR data.} \textbf{Bold} indicates the best performance excluding results using the full dataset. \textit{Italic} indicates the proposed prompting framework outperforms basic prompts. We use bootstrapping on all test set samples 100 times to report the mean±std results. The values in brackets represent the 95\% confidence interval, with the lower and upper bounds corresponding to the 2.5 and 97.5 percentiles, respectively. All metrics are multiplied by 100 for readability purposes.}
\label{tab:mm_results}
\centering
\footnotesize
\begin{tabular}{c|c|cc|cc}
\toprule
\multicolumn{2}{c|}{\multirow{2}{*}{\textbf{Methods}}} & \multicolumn{2}{c|}{\textbf{MIMIC-IV Outcome}} & \multicolumn{2}{c}{\textbf{MIMIC-IV Readmission}} \\
\multicolumn{2}{c|}{} & AUROC ($\uparrow$) & AUPRC ($\uparrow$) & AUROC ($\uparrow$) & AUPRC ($\uparrow$) \\ \midrule

\multirow{16}{*}{Prompt} 
 & OpenBioLLM-8B & \textbothscript{\colormmoutroc{47.78}}{4.58}{38.45}{56.56} & \textbothscript{\colormmoutprc{8.34}}{1.99}{4.91}{11.81} & \textbothscript{\colormmreadroc{52.82}}{4.20}{45.11}{60.82} & \textbothscript{\colormmreadprc{28.84}}{4.17}{22.21}{37.80} \\ \cmidrule{2-6}
 & Qwen2.5-7B & \textbothscript{\colormmoutroc{63.89}}{7.71}{49.89}{79.60} & \textbothscript{\colormmoutprc{19.07}}{6.01}{7.54}{30.87} & \textbothscript{\colormmreadroc{61.82}}{4.26}{54.44}{70.15} & \textbothscript{\colormmreadprc{35.65}}{5.18}{26.27}{45.63} \\ \cmidrule{2-6}
 & Gemma-3-4B & \textbothscript{\colormmoutroc{65.36}}{7.89}{53.44}{81.38} & \textbothscript{\colormmoutprc{28.98}}{9.63}{12.23}{46.43} & \textbothscript{\colormmreadroc{64.78}}{3.67}{57.30}{72.12} & \textbothscript{\colormmreadprc{38.05}}{4.56}{30.17}{46.79} \\ \cmidrule{2-6}
 & DeepSeek-V3.1 & \textbothscript{\colormmoutroc{91.58}}{2.02}{87.11}{95.02} & \textbothscript{\colormmoutprc{49.29}}{8.39}{29.43}{63.68} & \textbothscript{\colormmreadroc{57.34}}{4.56}{48.29}{65.12} & \textbothscript{\colormmreadprc{31.91}}{5.16}{23.59}{42.37} \\ \cmidrule{2-6}
 & GPT-4o & \textbothscript{\colormmoutroc{84.52}}{4.68}{75.48}{93.43} & \textbothscript{\colormmoutprc{43.82}}{9.96}{26.39}{63.26} & \textbothscript{\colormmreadroc{65.65}}{3.85}{59.06}{74.12} & \textbothscript{\colormmreadprc{38.64}}{5.05}{29.60}{47.68} \\ \cmidrule{2-6}
 & HuatuoGPT-o1-7B & \textbothscript{\colormmoutroc{71.12}}{6.56}{57.49}{81.15} & \textbothscript{\colormmoutprc{17.59}}{5.22}{7.98}{27.29} & \textbothscript{\colormmreadroc{62.11}}{4.49}{53.60}{69.15} & \textbothscript{\colormmreadprc{36.75}}{5.16}{27.91}{46.61} \\ \cmidrule{2-6}
 & DeepSeek-R1-7B & \textbothscript{\colormmoutroc{58.01}}{6.85}{45.46}{71.94} & \textbothscript{\colormmoutprc{14.00}}{5.15}{6.86}{28.03} & \textbothscript{\colormmreadroc{50.57}}{4.09}{43.63}{58.99} & \textbothscript{\colormmreadprc{26.67}}{3.76}{20.37}{34.87} \\ \cmidrule{2-6}
 & DeepSeek-R1 & \textbothscript{\colormmoutroc{87.99}}{4.14}{79.16}{95.15} & \textbothscript{\colormmoutprc{49.22}}{9.99}{32.81}{70.94} & \textbothscript{\colormmreadroc{55.32}}{4.83}{45.70}{64.08} & \textbothscript{\colormmreadprc{33.15}}{5.27}{24.50}{44.40} \\ \cmidrule{2-6}
 & DeepSeek-V3.1-Think & \textbothscript{\colormmoutroc{79.87}}{4.17}{72.61}{86.63} & \textbothscript{\colormmoutprc{39.33}}{9.28}{22.76}{58.75} & \textbothscript{\colormmreadroc{66.97}}{5.06}{56.53}{75.98} & \textbothscript{\colormmreadprc{46.44}}{6.72}{34.72}{60.45} \\ \cmidrule{2-6}
 & o3-mini-high & \textbothscript{\colormmoutroc{86.17}}{5.08}{73.51}{93.29} & \textbothscript{\colormmoutprc{48.84}}{9.24}{29.05}{63.95} & \textbothscript{\colormmreadroc{70.38}}{4.60}{59.60}{78.93} & \textbothscript{\colormmreadprc{49.47}}{5.72}{38.76}{60.41} \\ \cmidrule{2-6}
 & GPT-5 & \textbothscript{\colormmoutroc{92.03}}{2.03}{87.05}{95.02} & \textbothscript{\colormmoutprc{47.29}}{9.89}{24.94}{66.65} & \textbothscript{\colormmreadroc{73.75}}{3.77}{65.96}{81.22} & \textbothscript{\colormmreadprc{50.53}}{6.05}{40.36}{62.77} \\ \midrule
 
\multirow{5}{*}{Tuning}
 & Add & \textbothscript{\colormmoutroc{94.08}}{2.66}{88.55}{98.51} & \textbothscript{\colormmoutprc{79.30}}{8.22}{61.26}{92.49} & \textbothscript{\colormmreadroc{80.99}}{3.96}{72.38}{88.43} & \textbothscript{\colormmreadprc{66.80}}{5.86}{54.45}{76.33} \\ \cmidrule{2-6}
 & Concat & \textbothscript{\colormmoutroc{94.91}}{2.32}{90.81}{98.83} & \textbothscript{\textbf{\colormmoutprc{80.99}}}{7.48}{63.98}{93.52} & \textbothscript{\colormmreadroc{81.55}}{4.20}{72.69}{89.15} & \textbothscript{\colormmreadprc{68.12}}{5.80}{56.94}{77.89} \\ \cmidrule{2-6}
 & Self-Attention & \textbothscript{\textbf{\colormmoutroc{95.62}}}{1.92}{91.92}{98.94} & \textbothscript{\colormmoutprc{78.42}}{8.15}{62.04}{92.26} & \textbothscript{\colormmreadroc{80.38}}{4.23}{71.74}{88.12} & \textbothscript{\colormmreadprc{68.13}}{5.80}{55.27}{78.98} \\ \cmidrule{2-6}
 & Cross-Attention & \textbothscript{\colormmoutroc{95.35}}{2.08}{91.60}{98.98} & \textbothscript{\colormmoutprc{79.95}}{7.63}{63.43}{93.42} & \textbothscript{\textbf{\colormmreadroc{82.18}}}{4.06}{73.66}{89.27} & \textbothscript{\textbf{\colormmreadprc{71.34}}}{5.53}{59.76}{79.49} \\ \bottomrule
\end{tabular}
\end{table*}

Feature-fusion models show competent but not superior performance. Our feature-fusion approaches demonstrate that structured integration can yield strong predictive models. For in-hospital mortality, the Self-Attention fusion model achieved an AUROC of 95.62\%, a slight improvement over the best-performing unimodal structured EHR model (AdaCare, 94.28\%). However, this result does not surpass the performance of the top-performing LLM on clinical notes alone (DeepSeek-V3.1, 97.89\%). A similar trend was observed for 30-day readmission, where the best fusion model (Cross-Attention, 82.18\%) performed comparably to the best unimodal structured EHR model (LSTM, 82.52\%) but failed to match the best note-based model (o3-mini-high, 87.59\%). This suggests that for these tasks, unstructured clinical notes are an exceptionally rich source of information, and while feature fusion effectively leverages both data types, it struggles to add significant value when one modality is already dominant.

Prompt-based multimodal integration yields complex outcomes for LLMs. A key finding is that providing LLMs with combined structured and unstructured data in a single prompt does not straightforwardly improve upon their best unimodal performance. The multimodal performance is generally positioned between the LLMs' performance on structured EHR data alone and their performance on clinical notes alone. For instance, in mortality prediction, GPT-5's multimodal AUROC of 92.03\% improves upon its EHR-only performance (81.25\%) but remains substantially below the level achieved using clinical notes alone (97.60\%). Similarly, for readmission, its multimodal AUROC of 73.75\% is better than its EHR-only performance (65.65\%) but represents a considerable drop from its note-only performance (86.34\%). This indicates that while the models can process the combined input, they fail to effectively synthesize the information to reach the high predictive accuracy offered by the clinical narrative.

The observed performance gap suggests that current LLMs face challenges in zero-shot multimodal synthesis. One potential explanation is an attentional bias, where the model may disproportionately focus on one data type (e.g., the structured data presented first), thereby underutilizing the highly predictive signals in the narrative text. Furthermore, the artificial format of concatenated structured lists and free text can be an out-of-distribution challenge for models pretrained on natural language. Without explicit finetuning, these models may struggle to reconcile or properly weigh potentially conflicting signals from the two data types, leading to a suboptimal fusion of information and tempering expectations for out-of-the-box multimodal applications.

\subsection{Human Evaluation of LLM Reasoning Quality}
Our human evaluation by clinical experts revealed that the reasoning provided by top-performing LLMs was generally of high quality and clinically relevant. The quantitative scores, summarized in Table~\ref{tab:human_eval_quantitative}, show that the reasoning quality of the top-performing LLMs is generally decent, particularly when analyzing unstructured clinical notes for mortality prediction, where the reasoning was rated as highly accurate, complete, and useful (mean scores of 4.30, 4.60, and 4.65, respectively).

\begin{table}[!ht]
\centering
\caption{\textit{Human evaluation of LLM reasoning quality across clinical prediction tasks.} Mean scores ± standard deviation are reported on a 1--5 scale, where 5 represents the highest quality. The best-performing LLM from our main experiments was evaluated for each task.}
\label{tab:human_eval_quantitative}
\resizebox{\textwidth}{!}{
\begin{tabular}{@{}lccc@{}}
\toprule
\textbf{Task (MIMIC-IV)} & \textbf{Clinical Accuracy \& Safety} & \textbf{Reasoning \& Completeness} & \textbf{Clarity \& Clinical Utility} \\ \midrule
\multicolumn{4}{l}{\textit{\textbf{Structured EHR Data}}} \\
In-hospital Mortality & 3.41\std{1.00} & 4.14\std{0.67} & 4.03\std{0.76} \\
30-day Readmission    & 3.60\std{1.14}  & 4.30\std{0.73} & 4.05\std{1.19} \\ \midrule
\multicolumn{4}{l}{\textit{\textbf{Clinical Notes Data (Discharge Summary)}}} \\
In-hospital Mortality & 4.30\std{1.13} & 4.60\std{0.68} & 4.65\std{0.59} \\
30-day Readmission    & 2.80\std{1.57} & 3.93\std{1.22} & 3.13\std{1.46} \\ \bottomrule
\end{tabular}
}
\end{table}

For structured EHR data, the reasoning was also rated favorably, with ``Reasoning \& Completeness'' scoring above 4.0 for both tasks. This indicates the models are effective at identifying key risk factors from tabular data and logically connecting them to the prediction. However, the ``Clinical Accuracy \& Safety'' scores were moderately lower (3.41 and 3.60), reflecting instances where models occasionally misinterpreted or hallucinated minor details from the dense numerical input.

Notably, the most challenging setting for the LLM was predicting 30-day readmission from clinical notes. This task received the lowest scores across all dimensions, particularly for accuracy (2.80) and utility (3.13). This suggests that predicting readmission from a discharge summary requires a more nuanced synthesis of clinical and social factors, a task where the models' reasoning is currently less reliable and more prone to errors.

To move beyond aggregate performance metrics and understand the specific failure modes of LLMs in clinical prediction, we conducted a systematic error analysis of their generated reasoning. Leveraging our human evaluation framework and a clinician-developed error taxonomy (Figure~\ref{fig:error_taxonomy_tree}), we analyzed the common patterns of the confusion matrix. Our findings, summarized in Table~\ref{tab:error_analysis_frequency}, reveal distinct and clinically significant cognitive errors. Our key findings are as follows:

\begin{table}[!ht]
\centering
\caption{\textit{Frequency of error types in LLM reasoning across different prediction outcomes.} Values represent the percentage of cases within each outcome category where the specified error was identified by at least one expert evaluator. The denominator for each column is the total number of cases sampled for that category (TP: [n=15], TN: [n=24], FP: [n=60], FN: [n=1]).}
\label{tab:error_analysis_frequency}
\resizebox{\textwidth}{!}{%
\begin{tabular}{@{}lcccc@{}}
\toprule
\textbf{Error Type} & \textbf{True Positives (TP)} & \textbf{True Negatives (TN)} & \textbf{False Positives (FP)} & \textbf{False Negatives (FN)} \\ \midrule
Factual Inconsistency           & 0.0\%     & 0.0\% & \textbf{28.3\%} & 0.0\% \\
Omission of Key Information     & 0.0\%     & 0.0\% & 3.3\%  & 0.0\% \\
Flawed Logic or Reasoning       & 0.0\%     & 4.2\% & 0.0\%  & \textbf{100.0\%} \\
Inclusion of Irrelevant Info.   & 0.0\%     & 0.0\% & 0.0\%  & 0.0\% \\
Inappropriate Confidence        & 6.7\%     & 4.2\% & 13.3\% & \textbf{100.0\%} \\ \bottomrule
\end{tabular}%
}
\end{table}

\begin{itemize}
    \item \textbf{False Positives (FPs) are primarily driven by factual inconsistency.} The most frequent error in FP cases was ``Factual Inconsistency / Hallucination''. In these instances, the model often justified a high-risk prediction by citing incorrect data or hallucinating a non-existent comorbidity, effectively creating a high-risk narrative based on flawed evidence.

    \item \textbf{False Negatives (FNs) are characterized by flawed clinical reasoning.} As confirmed by our deep-dive, the defining error in FN cases was ``Flawed Logic or Reasoning''. The models consistently identified the correct risk factors but failed to appreciate their collective weight, leading to a dangerous underestimation of the patient's true risk. This suggests a weakness in higher-order clinical judgment rather than data extraction.

    \item \textbf{Errors persist in correct predictions} Critically, errors were also present, albeit at lower frequencies, in correctly classified cases (TPs and TNs). For example, a model might correctly predict a high-risk outcome (TP) but base its reasoning on a minor factual inconsistency. This underscores the importance of evaluating the reasoning process, as a model that is correct for the wrong reasons is unreliable and poses risks in a clinical setting.
\end{itemize}

In summary, this detailed error analysis provides a deeper understanding of LLM predictive behavior, showing that FPs and FNs often stem from fundamentally different types of cognitive failures: flawed synthesis versus incomplete information gathering. These insights are crucial for guiding future research toward improving model reliability and are a vital step toward the safe and effective deployment of LLMs in clinical decision support.

\section{Discussion}
Our comprehensive benchmarking reveals a significant shift in the capabilities of Large Language Models (LLMs) for non-generative clinical prediction tasks, challenging the prevailing assumption of their general inferiority to specialized models. 

\subsection{Ethical Considerations and Model Fairness}
The responsible deployment of predictive models in healthcare necessitates a rigorous evaluation of their ethical implications, particularly regarding fairness and potential biases across patient subgroups. To address this, we conducted a comprehensive fairness analysis across demographic attributes (Age, Gender, Race), with the full methodology and results detailed in Appendix~\ref{sec:appendix_fairness}. Our analysis revealed several critical trends.

First, state-of-the-art LLMs in zero-shot settings generally demonstrated greater fairness, with metrics for disparate impact and equal opportunity closer to ideal values compared to many conventional ML/DL models. Second, we observed that conventional models, while achieving high accuracy when trained on the full dataset, sometimes amplified underlying biases present in the data, particularly concerning age and gender. Third, the process of finetuning language models, though often beneficial for task-specific accuracy, was found to occasionally introduce or exacerbate fairness disparities, likely by overfitting to spurious demographic correlations in the training data. Finally, our results suggest that prompt engineering can serve as a tool for bias mitigation; the use of optimized prompts and in-context learning often not only improved predictive performance but also led to more equitable outcomes.

These findings collectively underscore that predictive accuracy is an incomplete measure of a model's clinical utility. A holistic assessment must include a thorough fairness audit. While no model was perfectly equitable, the promising fairness characteristics of thoughtfully prompted LLMs, coupled with the risks associated with conventional training and finetuning, highlight the need for ongoing research into developing and validating models that are not only accurate but also fair and trustworthy.

\subsection{Comparative Analysis with Existing Model Paradigms}
These findings necessitate a re-evaluation of the role of BERT-style models in clinical prediction. While previously considered strong contenders for clinical NLP, our results show they are now significantly outperformed by newer generation LLMs for prediction tasks based on clinical notes. Although finetuned BERT models might still offer a practical alternative where computational resources are extremely constrained or the deployment of very large LLMs is infeasible, their performance advantage has diminished in the face of rapidly advancing LLM capabilities. Conventional machine learning and EHR-specific deep learning models, however, maintain their top-tier status for structured EHR data predictions when sufficient training data is available, excelling at understanding numerical values and recognizing complex temporal patterns inherent in such datasets.

\subsection{Trustworthiness and Practical Deployment Challenges}
Beyond predictive performance, our study underscores the importance of evaluating the reasoning behind LLM predictions, a critical factor for clinical adoption. Our human evaluation demonstrated that the explanations from top-performing LLMs were rated highly by clinical experts for their accuracy, completeness, and utility. This validates our methodological choice to prompt for reasoning and suggests that these models can produce outputs that are not just numerically correct but also clinically coherent and useful for decision support. However, this evaluation also served to identify specific, recurring error patterns. While the overall quality was high, instances of factual inconsistency, omission of key risk factors, and flawed logic were still present. This highlights the reality that while modern LLMs are remarkably capable, they are not infallible. A detailed analysis of these failure modes is crucial for understanding their limitations and guiding future research toward improving the reliability and trustworthiness of LLM reasoning in high-stakes medical applications.

We observed important behavioral characteristics of LLMs. Our analysis of failure-to-predict rates revealed significant variability in model reliability. Consistent with findings in the broader LLM literature, prompt adherence remains a challenge, particularly for smaller or less advanced models. While state-of-the-art models like GPT-5 and DeepSeek-V3.1 demonstrated near-perfect reliability, others struggled, especially with complex prompts or unstructured data. We found that effective prompt engineering was critical for improving reliability, though overly complex prompts could sometimes be counterproductive. We also identified unique failure modes, such as excessive verbosity leading to truncated outputs. These findings underscore that practical deployment requires evaluating not just accuracy but also robustness and instruction-following capability. This reinforces the need for LLM research to expand beyond medical Q\&A to include finetuning and evaluation on diverse data types and emphasizes that extensive empirical testing and validation are crucial before deployment in critical clinical applications.

\subsection{Limitations and Future Directions}
Our study has limitations, the first of which is around the limited task and dataset diversity. Our structured data evaluations were confined to mortality and readmission predictions using text from the MIMIC-IV, and TJH datasets. Investigating the integration of additional data modalities (e.g., medical imaging and genomics) and diverse clinical tasks (e.g., diagnosis prediction, treatment recommendation, and adverse event forecasting) could provide a more comprehensive assessment of model efficacy. The second limitation is around model selection. Although our study included a comprehensive range of models, the rapid advancements in the field may lead to the development of new more effective models after our study period. We do believe however that these results capture a generalizable conclusion reflecting the underlying architecture of these approaches. However, continuous incorporation of emerging state-of-the-art models will ensure that benchmarks remain relevant and informative.

For the broader clinical large language model research, our benchmarking results and practical experiences have uncovered several critical research gaps and future directions:

\begin{itemize}
    \item \textbf{Broadening Training and Evaluation for Robust Non-Generative Capabilities:} While this study focuses on non-generative tasks, most clinical LLM development is still heavily skewed towards medical literature finetuning and generative task evaluation (e.g., Q\&A, summarization). There remains a pressing need to expand training paradigms and evaluation benchmarks to more systematically include diverse non-generative applications using real-world clinical data, including structured EHR, to better assess and enhance their utility in decision-driven clinical workflows. Our findings of LLM success on notes should encourage more exploration here.
    \item \textbf{Improving LLM Comprehension of Complex Longitudinal EHR Data:} Despite impressive zero-shot performance on structured EHR in data-scarce settings, LLMs still generally lag behind specialized conventional models when ample longitudinal data is available. Current LLM architectures may not optimally capture the complex temporal dependencies and nuanced feature interplay in rich EHR datasets. Future research could explore novel architectures, specialized pretraining on EHR data, or hybrid approaches, perhaps integrating LLMs with conventional time-series models or using agent-based frameworks to combine LLM reasoning with the precision of specialized models, thereby improving predictions and reducing errors.
    \item \textbf{Addressing Deployment Complexity, Efficiency, and Trustworthiness:} The practical deployment of the most powerful LLMs in secure healthcare environments faces hurdles due to high computational costs, potential privacy issues, and the need for robust prompt adherence. While the emergence of high-performing open-source models (e.g., DeepSeek) offers pathways to greater data control and on-premise deployment, challenges remain for institutions with limited GPU capacity or strict data governance policies. Continued research into model compression, efficient inference techniques, certified privacy-preserving methods, and improving LLM reliability and interpretability is essential for their responsible integration into clinical practice.
\end{itemize}

\subsection{Conclusion}
In conclusion, our ClinicRealm benchmark provides preliminary evidence that modern LLMs have emerged as competitive tools for non-generative clinical prediction. Contrary to previous assumptions, recent LLMs demonstrate state-of-the-art performance on prediction tasks using clinical notes and show considerable promise for structured EHR data, especially in data-limited situations or for rapid initial model deployment. 

Healthcare providers and institutions should recognize these advancements. While specialized conventional models remain the gold standard for structured EHR predictions given sufficient data, the narrowing performance gap and the unique strengths of LLMs (e.g., zero-shot learning from text, handling unstructured data) position them as powerful tools in the clinical analytics arsenal. The strong performance of open-source LLMs further democratizes access to these capabilities. Policymakers and researchers should foster an environment that supports the continued development, rigorous evaluation, and responsible integration of LLMs, focusing on enhancing their reliability, efficiency, and understanding of complex medical data. Rather than a blanket preference for one model type, optimal clinical decision support will likely involve a nuanced selection strategy, leveraging the distinct strengths of conventional models, BERT-style approaches (where appropriate for resource constraints), and the burgeoning capabilities of modern LLMs.

\section*{Data Privacy and Code Availability Statement}

To ensure the fairness and reproducibility of the comparison, this research did not involve the collection of new patient EHR data. The TJH EHR dataset utilized in this study is publicly available on GitHub (\url{https://github.com/HAIRLAB/Pre_Surv_COVID_19}). The MIMIC-IV datasets are open to researchers and can be accessed on request, including structured EHR data~\cite{mimiciv_v3_1} (\url{https://physionet.org/content/mimiciv/3.1/}) and clinical notes data~\cite{mimicivnote} (\url{https://physionet.org/content/mimic-iv-note/2.2/}). We used these datasets under their respective licenses. Throughout the experiments, we strictly adhered to the data use agreement, reaffirming our commitment to responsible data handling and usage. The performance of OpenAI models on all datasets was processed using the secure Azure OpenAI API, with human review of the data waived. Additionally, all other models, including ML, DL, and other LLMs, were deployed locally. The code for this benchmarking work can be accessed online (\url{https://github.com/yhzhu99/ehr-llm-benchmark}). We also provide the up-to-date benchmark results online (\url{https://yhzhu99.github.io/ehr-llm-benchmark/}).

\section*{Acknowledgements}

This work was supported by the National Natural Science Foundation of China (62402017), Beijing Natural Science Foundation (L244063), Xuzhou Scientific Technological Projects (KC23143), and Peking University Medicine plus X Pilot Program-Key Technologies R\&D Project (2024YXXLHGG007). Junyi Gao acknowledges the receipt of studentship awards from the Health Data Research UK-The Alan Turing Institute Wellcome PhD Programme in Health Data Science (grant 218529/Z/19/Z) and Baidu Scholarship. We extend our gratitude to Jingkun An, Yuning Tong, Enshen Zhou, Bowen Jiang, and Yifan He for their preliminary discussions and experiments. We thank Enshen Zhou for providing computing resources for part of our research. We also thank Ahmed Allam for his suggestions on experimental settings.

\section*{Author Contributions Statement}

Y.Z., J.G., Z.W. and L.M. conceived the experiments, Z.W., Y.Z. curated datasets, Z.W., Y.Z. conducted experiments on EHR data, Y.Z., J.G., Z.W., W.L., X.Z., and L.L. conducted experiments on clinical notes. Y.Z., J.G., Z.W., W.L. analysed the experimental results. All authors wrote and reviewed the manuscript.

\bibliography{ref}

\appendix
\cleardoublepage
\centerline{\Large\bfseries Appendix}
\vspace{1cm}

\startcontents[appendix]
\printcontents[appendix]{}{1}{\setcounter{tocdepth}{2}}

\section{Experimental Setups}\label{sec:detail_experimental_setups}

\paragraph{Datasets details.}

This study utilizes three primary datasets: TJH~\cite{tjh}, MIMIC-IV~\cite{mimic4}, and MIMIC-III~\cite{mimic3}.

\begin{itemize}
\item The \textbf{TJH dataset}~\cite{tjh} is derived from Tongji Hospital and is publicly available on GitHub. It comprises structured EHR data for 485 COVID-19 patients admitted during the initial COVID-19 outbreak, including 73 numerical lab tests and vital signs, plus two demographic features (age and gender). As this dataset was collected for the primary purpose of the \textit{Nature Machine Intelligence} study ``An interpretable mortality prediction model for COVID-19 patients'', its target variable is binary (survival vs. death). Crucially, it does not include post-discharge follow-up information, making it unsuitable for readmission prediction tasks.
\item The \textbf{MIMIC-IV dataset}\cite{mimic4} is sourced from the EHRs of the Beth Israel Deaconess Medical Center. We utilize version 3.1 of its structured EHR data~\cite{mimiciv_v3_1} and version 2.2 of its clinical notes (discharge summaries)\cite{mimicivnote}.
\item The \textbf{MIMIC-III dataset}\cite{mimic3} is an earlier version from the same medical center. We use version 1.4~\cite{mimic3} to extract admission notes for our prospective mortality prediction task.
\end{itemize}

Our preprocessing methodology adheres to established benchmark pipelines~\cite{gao2024comprehensive,zhu2023pyehr}. For MIMIC-III and MIMIC-IV, each hospital admission is treated as a separate sample.

For the \textbf{MIMIC-IV structured EHR data}, to mitigate issues arising from missing values, we first consolidate all data segments from the same admission on a daily basis, taking the last recorded value for each of the 17 physiologic variables within a 24-hour window. For patients whose hospital stays exceed seven days, we retain records exclusively from the final seven days, while earlier records are aggregated into a single initial time step (also using the last recorded value for each feature). For conventional ML/DL models, any remaining missing values are addressed using the Last Observation Carried Forward (LOCF) imputation strategy~\cite{wells2013LOCF_Imputation}. For LLMs, missing values are explicitly passed as ``NaN''. Structured data were standardized using z-score normalization, and extreme outliers (absolute z-score > 10,000) were removed.

Regarding the \textbf{clinical notes}, we used discharge summaries from MIMIC-IV and admission notes from MIMIC-III. To create the admission notes cohort from MIMIC-III, we extracted all ``Physician'', ``Nursing'', and ``Nursing/other'' notes recorded within the first 24 hours of admission and concatenated them chronologically. The preprocessing of all notes follows the Clinical-Longformer approach~\cite{li2023ClinicalLongformer}, which involves several key steps: removal of all de-identification placeholders, replacement of non-alphanumeric characters with spaces, conversion of all text to lowercase, and stripping of extraneous white spaces. Critically, to prevent label leakage, we programmatically removed any explicit mentions of outcomes (e.g., ``expired'', ``deceased'', ``readmitted'') from the notes.

Furthermore, consistent with benchmark practices, we employ a stratified shuffling strategy with random selection to partition the data into training, validation, and test sets. Crucially, the test set is held constant across all evaluated models to ensure a fair and robust comparison of performance. Detailed statistics for all data modalities are presented in Table~\ref{tab:dataset_stats}. The complete preprocessing code is publicly available at \url{https://github.com/PKU-AICare/mimic_preprocessor/}.

\paragraph{Computational infrastructure and software.}
The LLM generation experiments detailed herein are conducted from April 10, 2025, to April 24, 2025, and from September 6, 2025, to September 21, 2025.
For prediction tasks involving structured EHR data, both the training of machine learning/deep learning models and the LLM generation experiments are performed on a server equipped with 128GB of RAM and a single NVIDIA RTX 3090 GPU (CUDA 12.5).
For prediction tasks utilizing clinical notes:
\begin{itemize}
    \item Experiments with LLMs in the freeze setting (training only the MLP classifier head) and the finetuning of BioGPT are conducted on a Mac Studio M2 Ultra with 192GB of RAM for MIMIC-IV dataset, and on a Mac Studio M3 Ultra with 512GB of RAM for MIMIC-III dataset.
    \item Finetuning of the other LLMs is carried out on a system featuring an NVIDIA A100 GPU (80GB VRAM) and 64GB of RAM for MIMIC-IV dataset, and on the same Mac Studio M3 Ultra with 512GB of RAM for MIMIC-III dataset.
    \item LLM inference experiments using prompting methodologies are executed on the same NVIDIA RTX 3090 server used for the structured EHR data tasks for MIMIC-IV dataset, and on a MacBook M3 Air with 24GB of RAM for MIMIC-III dataset.
\end{itemize}
The primary software stack comprised Python 3.12, PyTorch 2.6.0, PyTorch Lightning 2.5.1, and Transformers 4.50.0.

\paragraph{Model training and hyperparameters.}
Across all training experiments, the AdamW optimizer~\cite{loshchilov2017decoupled} is employed. For conventional EHR prediction models, training proceeds for a maximum of 50 epochs on the designated training set. To mitigate overfitting, an early stopping strategy is implemented with a patience of 5 epochs, monitored by AUROC for classification tasks and MAE for the regression task. The learning rate is selected via grid search from the set $\{1 \times 10^{-2}, 1 \times 10^{-3}, 1 \times 10^{-4}\}$. These models utilize a hidden dimension of 128 and a batch size of 256.
For language models (LMs), the following training configurations are applied:
\begin{itemize}
    \item \textbf{Freeze Setting}: The MLP classifier head appended to each LM is trained for a maximum of 50 epochs with an early stopping patience of 10 epochs. The learning rate is set to $1 \times 10^{-4}$, and the batch size is 64.
    \item \textbf{Finetuning BERT-style LMs}: These models are finetuned for a maximum of 10 epochs with an early stopping patience of 3 epochs. The learning rate is $1 \times 10^{-5}$, and the batch size is 16.
    \item \textbf{Finetuning GPT-style LLMs}: These models are finetuned for a maximum of 5 epochs with an early stopping patience of 1 epoch. The learning rate is $2 \times 10^{-4}$, and the batch size is 8.
\end{itemize}

\paragraph{LM-specific configurations.}
For feature extraction from LMs, we utilize the embedding of the first token ([CLS]) from the final layer of BERT-style models. For GPT-style models, the embedding of the last token from their final layer is used.
In experiments involving clinical notes (both frozen backbone and finetuning settings), the maximum input sequence length is set to 512 tokens for all LMs. For LLM generation experiments, this is expanded to 8192 tokens to leverage the models' capacity for longer contexts.

\paragraph{Model access and deployment.}
We utilize OpenAI's APIs, including the GPT-5 \\ (\texttt{gpt-5-chat-latest} with reasoning effort of high), the GPT-4o (\texttt{chatgpt-4o-latest}) and GPT o3-mini-high (\texttt{o3-mini-high}) models. All other LMs evaluated in this study are deployed locally. LLM generation experiments using these local models for both structured EHR data and unstructured clinical notes are facilitated by LMStudio. For the freeze and finetuning experiments, LMs are loaded using the Hugging Face Transformers library without any additional quantization, thereby preserving their original precision and computational capabilities.

The detailed model settings and quantized versions fetched from HuggingFace are:
\begin{enumerate}
    \item BERT: \texttt{bert-base-uncased}
    \item ClinicalBERT: \texttt{medicalai/ClinicalBERT}
    \item BioBERT: \texttt{pritamdeka/BioBert-PubMed200kRCT}
    \item GatorTron: \texttt{UFNLP/gatortron-base} with 345 million parameters
    \item Clinical-Longformer: \texttt{yikuan8/Clinical-Longformer}

    \item GPT-2: 
        \begin{itemize}
            \item \textbf{prompt}: \texttt{QuantFactory/gpt2-large-GGUF} (8-bit quantized version)
            \item \textbf{embedding}: \texttt{openai-community/gpt2}
        \end{itemize}
    \item BioGPT: 
        \begin{itemize}
            \item \textbf{prompt}: \texttt{RichardErkhov/akhilanilkumar\_-\_biogpt-baseline-gguf} (8-bit quantized version)
            \item \textbf{embedding}: \texttt{microsoft/biogpt}
        \end{itemize}
    \item Meditron: 
        \begin{itemize}
            \item \textbf{prompt}: \texttt{mlx-community/meditron-7b} (4-bit quantized version)
            \item \textbf{embedding}: \texttt{epfl-llm/meditron-7b}
        \end{itemize}
    \item BioMistral: 
        \begin{itemize}
            \item \textbf{prompt}: \texttt{MaziyarPanahi/BioMistral-7B-GGUF} (8-bit quantized version)
            \item \textbf{embedding}: \texttt{BioMistral/BioMistral-7B}
        \end{itemize}
    \item OpenBioLLM: 
        \begin{itemize}
            \item \textbf{prompt}: \texttt{aaditya/OpenBioLLM-Llama3-8B-GGUF} (8-bit quantized version)
            \item \textbf{embedding}: \texttt{aaditya/Llama3-OpenBioLLM-8B}
        \end{itemize}
    \item Qwen2.5: 
        \begin{itemize}
            \item \textbf{prompt}: \texttt{lmstudio-community/Qwen2.5-7B-Instruct-1M-GGUF} (8-bit quantized version)
            \item \textbf{embedding}: \texttt{Qwen/Qwen2.5-7B}
        \end{itemize}
    \item Gemma-3: 
        \begin{itemize}
            \item \textbf{prompt}: \texttt{lmstudio-community/gemma-3-4b-it-GGUF} (8-bit quantized version)
            \item \textbf{embedding}: \texttt{google/gemma-3-4b-pt}
        \end{itemize}
    \item DeepSeek-V3.1: 
        \begin{itemize}
            \item \textbf{prompt}: \texttt{unsloth/DeepSeek-V3.1-GGUF} (8-bit quantized version)
        \end{itemize}

    \item HuatuoGPT-o1: 
        \begin{itemize}
            \item \textbf{prompt}: \texttt{QuantFactory/HuatuoGPT-o1-7B-GGUF} (8-bit quantized version)
            \item \textbf{embedding}: \texttt{FreedomIntelligence/HuatuoGPT-o1-7B}
        \end{itemize}
    \item DeepSeek-R1-Distill-Qwen: 
        \begin{itemize}
            \item \textbf{prompt}: \texttt{lmstudio-community/DeepSeek-R1-Distill-Qwen-7B-GGUF} (8-bit quantized version)
            \item \textbf{embedding}: \texttt{deepseek-ai/DeepSeek-R1-Distill-Qwen-7B}
        \end{itemize}
    \item DeepSeek-R1:
        \begin{itemize}
            \item \textbf{prompt}: \texttt{unsloth/DeepSeek-R1-0528-GGUF} (8-bit quantized version)
        \end{itemize}
    \item DeepSeek-V3.1-Think: 
        \begin{itemize}
            \item \textbf{prompt}: \texttt{unsloth/DeepSeek-V3.1-GGUF} (8-bit quantized version)
        \end{itemize}
\end{enumerate}

\section{Interpreting Performance Disparities Across Datasets}\label{sec:appendix_disparities}
Our results highlight a significant performance disparity for conventional models like CatBoost, which achieved near-perfect results on the TJH dataset but substantially lower scores on MIMIC-IV. This observation is not indicative of a model generalization failure but rather reflects the fundamental differences in task complexity and data characteristics between the two datasets.

The TJH dataset represents a highly focused and less complex clinical problem. It contains a relatively homogeneous cohort of 485 COVID-19 patients where mortality was found to be strongly correlated with a few key biomarkers (e.g., LDH, lymphocytes)~\cite{tjh}. The original study demonstrated that a simple decision-tree model could achieve very high accuracy, indicating that the predictive signals are strong and easily identifiable. Consequently, powerful models like CatBoost can readily learn these patterns, leading to the near-perfect performance we observed.

In stark contrast, MIMIC-IV presents a far more complex challenge. It is a large, heterogeneous dataset encompassing tens of thousands of patients with a wide spectrum of diseases, comorbidities, and backgrounds. This diversity introduces significant statistical noise and clinical complexity, making prediction tasks inherently more difficult. The predictive signals are more subtle, multifactorial, and subject to greater uncertainty.

\begin{figure}[!ht]
\centering
\subfigure[CatBoost on MIMIC-IV mortality]{
    \includegraphics[width=0.45\linewidth]{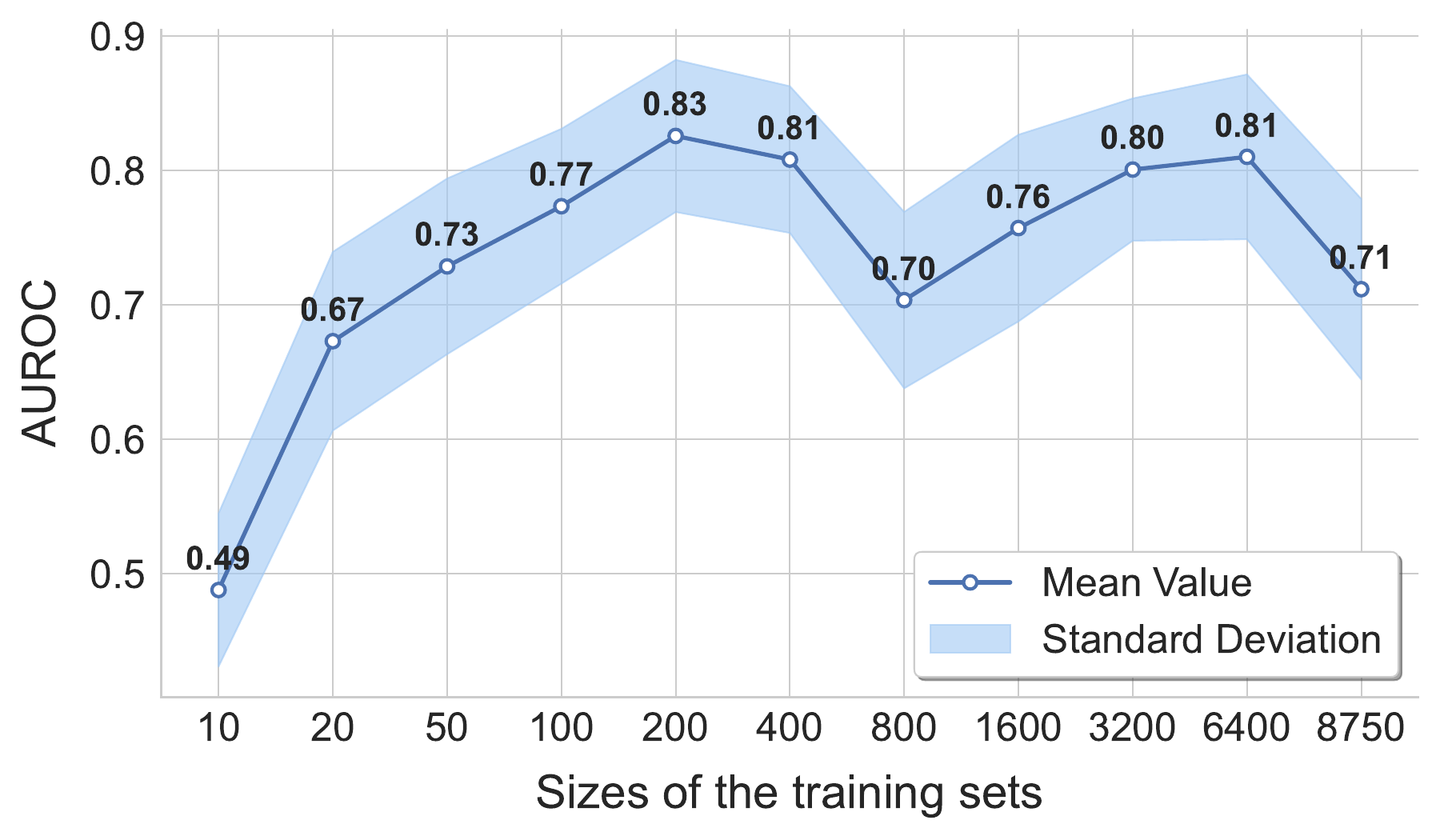}
    \label{fig:mimic4_mortality_CatBoost}
}
\subfigure[XGBoost on MIMIC-IV mortality]{
    \includegraphics[width=0.45\linewidth]{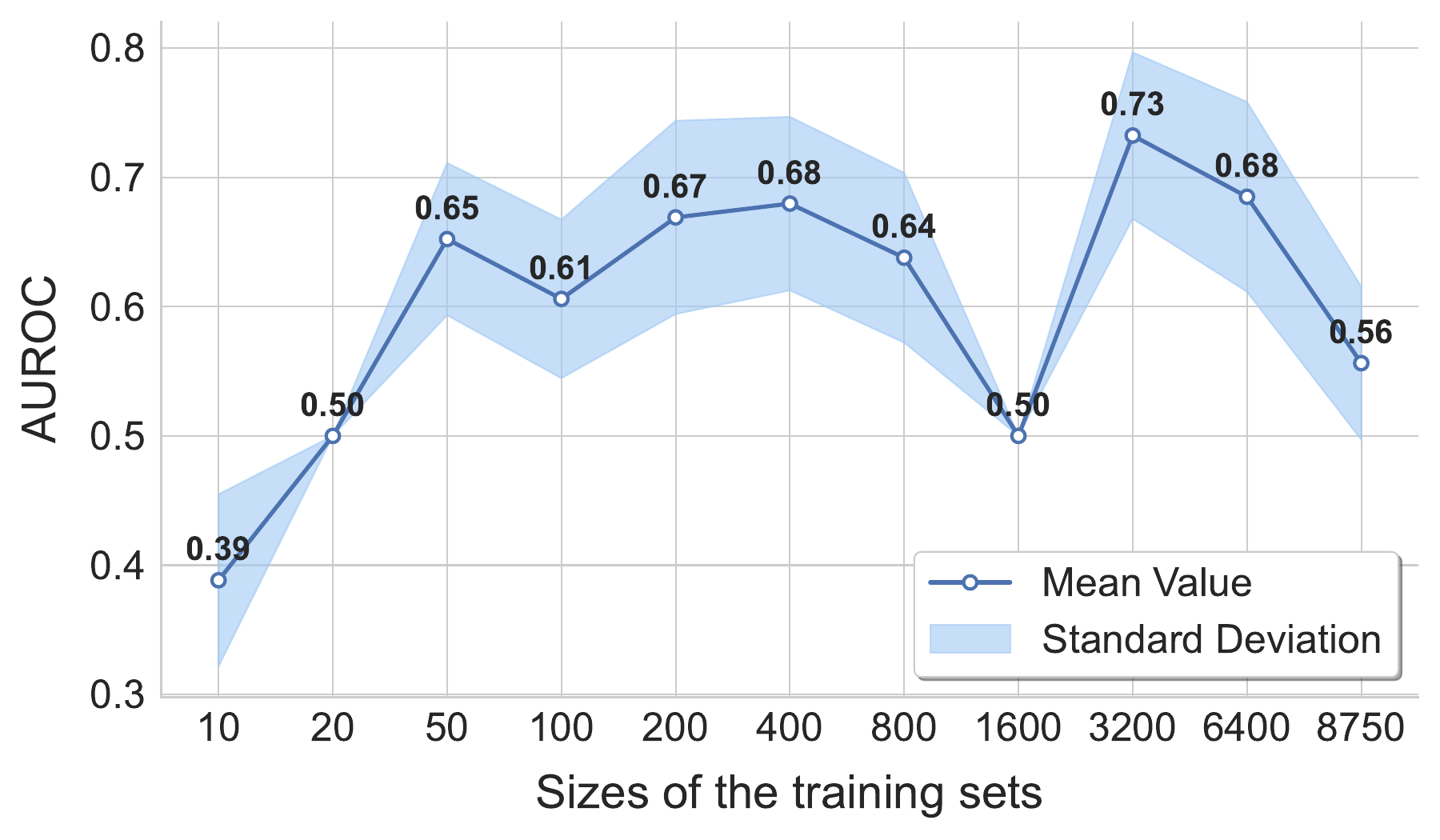}
    \label{fig:mimic4_mortality_XGBoost}
}
\subfigure[CatBoost on MIMIC-IV readmission]{
    \includegraphics[width=0.45\linewidth]{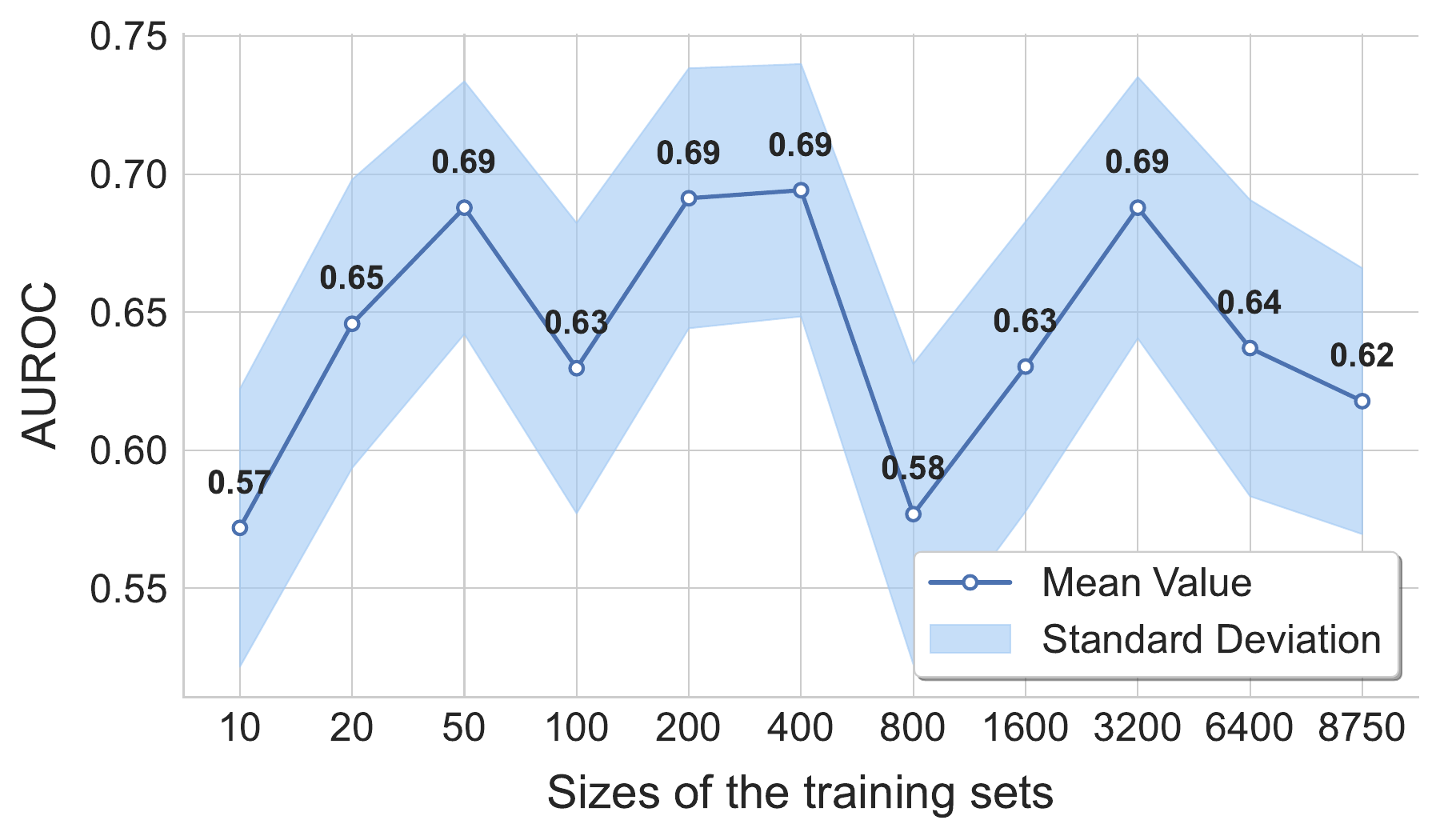}
    \label{fig:mimic4_readmission_CatBoost}
}
\subfigure[XGBoost on MIMIC-IV readmission]{
    \includegraphics[width=0.45\linewidth]{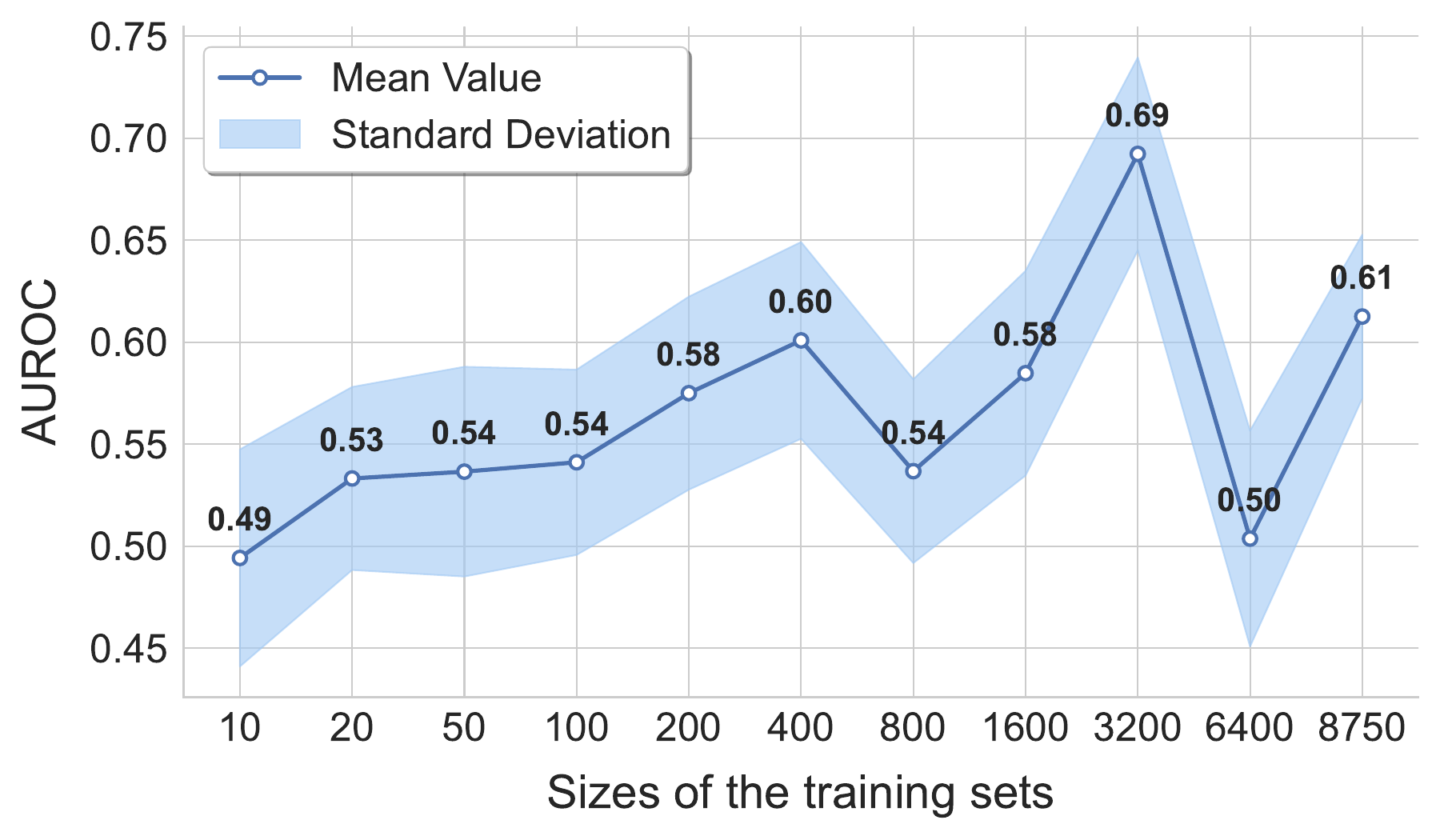}
    \label{fig:mimic4_readmission_XGBoost}
}
\caption{\textit{AUROC performance of CatBoost and XGBoost on MIMIC-IV with varying training set sizes.} The plots illustrate the AUROC for mortality (a, b) and readmission (c, d) prediction tasks. The shaded area represents the standard deviation over multiple runs. Performance for both models tends to saturate after the training set size reaches a few hundred samples, demonstrating that increasing data volume provides diminishing returns for these complex and heterogeneous prediction tasks.}
\label{fig:mimic4_performance_curve}
\end{figure}

While MIMIC-IV offers a much larger training set, performance on such complex data often exhibits diminishing returns. To investigate this, we analyzed model performance on progressively larger subsets of the MIMIC-IV training data (Figure~\ref{fig:mimic4_performance_curve}). The results clearly demonstrate a performance plateau. For both mortality and readmission tasks, the AUROC for CatBoost and XGBoost saturates after training on only a few hundred samples, with negligible gains from thousands of additional data points. This saturation indicates that the models have effectively captured the primary learnable patterns within the feature space, and the remaining performance gap is likely attributable to the inherent clinical uncertainty and irreducible variance within this diverse population.

In conclusion, the observed performance differences are explained by the nature of the clinical problems themselves. The TJH task is well-defined with strong signals and a high performance ceiling, whereas the MIMIC-IV tasks are broader and inherently more challenging, with a lower performance ceiling that cannot be overcome simply by adding more data. This context is crucial for interpreting the cross-dataset performance of all evaluated models.

\section{Ethical Considerations: Fairness and Subgroup Bias Analysis}\label{sec:appendix_fairness}

A thorough evaluation of the models benchmarked in our study requires a discussion of ethical considerations, particularly fairness and potential biases across patient subgroups. The deployment of predictive models in clinical settings carries a significant responsibility to ensure equitable performance for all patient populations. This section details our fairness analysis across all experimental settings, evaluating model performance with respect to sensitive demographic attributes, including Age, Gender, and Race.

\paragraph{Privileged and unprivileged groups.}
To quantify fairness, we first define privileged and unprivileged groups based on three sensitive attributes ($A$). Let $A=1$ denote the privileged group and $A=0$ denote the unprivileged group. Based on historical and societal biases often reflected in healthcare data, we define the groups as follows:
\begin{itemize}
    \item \textbf{Age:} The privileged group consists of patients aged 50 years or older ($A=1$), while the unprivileged group includes patients younger than 50 ($A=0$). This threshold is chosen to differentiate between younger and older adult populations who may experience different health outcomes and care patterns.
    \item \textbf{Gender:} The privileged group is defined as Male patients ($A=1$), and the unprivileged group as Female patients ($A=0$).
    \item \textbf{Race:} We categorize patients into two groups: White (privileged, $A=1$) and non-White (unprivileged, $A=0$), which includes all other racial and ethnic identities present in the datasets. This binary categorization is a common practice in fairness audits due to sample size limitations for individual minority groups.
\end{itemize}

\paragraph{Fairness metrics.}
We employ four standard group fairness metrics to assess potential biases. Let $\hat{Y}$ be the model's predicted outcome (where $\hat{Y}=1$ indicates a positive prediction, e.g., mortality) and $Y$ be the true outcome. The metrics are defined as follows:

\begin{itemize}
    \item \textbf{Disparate Impact (DI):} This metric is the ratio of the rate of positive predictions for the unprivileged group to that of the privileged group. It is calculated as $\text{DI} = \frac{P(\hat{Y}=1 | A=0)}{P(\hat{Y}=1 | A=1)}$. A value of 1.0 indicates perfect fairness, meaning both groups receive positive predictions at the same rate. Values substantially deviating from 1.0 (typically, outside the range [0.8, 1.25]) suggest potential bias.

    \item \textbf{Statistical Parity Difference (SPD):} This metric measures the absolute difference in the rate of positive predictions between the unprivileged and privileged groups: $\text{SPD} = P(\hat{Y}=1 | A=0) - P(\hat{Y}=1 | A=1)$. An ideal value is 0, indicating that the model's predictions are independent of the sensitive attribute. Negative values imply that the unprivileged group is less likely to receive a positive prediction.

    \item \textbf{Average Odds Difference (AOD):} This metric assesses equality of opportunity by averaging the difference in true positive rates (TPR) and false positive rates (FPR) between groups: $\text{AOD} = \frac{1}{2} [ (\text{TPR}_{A=0} - \text{TPR}_{A=1}) + (\text{FPR}_{A=0} - \text{FPR}_{A=1}) ]$. An ideal value is 0. This metric is less sensitive to underlying base rate differences between groups than SPD.

    \item \textbf{Equalized Opportunity Difference (EOD):} This metric focuses specifically on the difference in true positive rates between groups: $\text{EOD} = \text{TPR}_{A=0} - \text{TPR}_{A=1} = P(\hat{Y}=1 | A=0, Y=1) - P(\hat{Y}=1 | A=1, Y=1)$. It measures whether the model correctly identifies positive outcomes at an equal rate for both groups. An ideal value is 0.
\end{itemize}

\paragraph{Analysis of fairness results.}
We conducted a comprehensive fairness analysis for all models on the TJH mortality task, MIMIC-III and MIMIC-IV mortality tasks, the MIMIC-IV readmission task, and the multimodal MIMIC-IV tasks. The detailed results are presented in Tables~\ref{tab:fairness_tjh_mortality} through \ref{tab:fairness_multimodal_mimic4}. Our analysis reveals several important trends:

\begin{enumerate}
    \item \textbf{LLMs generally exhibit greater fairness in zero-shot settings.} Across multiple tasks and datasets, modern LLMs used in a zero-shot prompting setting (e.g., OpenBioLLM, Gemma-3, GPT-4o, DeepSeek variants) consistently demonstrate fairness metrics closer to the ideal values compared to conventional ML/DL models. For instance, in the TJH mortality prediction task (Table~\ref{tab:fairness_tjh_mortality}), the fully trained CatBoost model shows a substantial disparity based on Age (DI = 9.00, SPD = 0.5333). In contrast, LLMs like OpenBioLLM with an optimized prompt and in-context learning achieve near-parity on the same attribute (DI = 1.0255, SPD = 0.0119). This trend holds for tasks on MIMIC-IV with structured EHR as well (Table~\ref{tab:fairness_mimic4_mortality}).

    \item \textbf{Finetuning can introduce or exacerbate bias in language models.} For tasks on unstructured clinical notes (Table~\ref{tab:fairness_mimic4_mortality} and \ref{tab:fairness_mimic4_readmission}), we observe that finetuning BERT-style models or even some LLMs can sometimes lead to worse fairness outcomes compared to using them in a zero-shot (``prompt'') or feature-extraction (``freeze'') setting. For instance, on the MIMIC-IV mortality prediction task with clinical notes, the finetuned Clinical-Longformer shows notable disparity for Age (AOD = 0.2379) compared to freezed setting (AOD = 0.0312). This suggests that the finetuning process might overfit to spurious correlations associated with demographic groups in the training data.

    \item \textbf{Prompt engineering influences fairness.} The use of optimized prompts and in-context learning often leads to improved fairness. In many cases across the LLM results, the ``optimized prompt'' and ``opt.+ICL'' settings not only enhance predictive performance but also yield fairness metrics closer to the ideal. For example, on the TJH mortality task for GPT-4o, the ``optimized prompt'' setting reduced the SPD for Age from 0.5452 (``base prompt'') to 0.2667. This highlights the potential of prompt engineering as a tool for bias mitigation.
\end{enumerate}

In summary, our fairness analysis reveals that while no model is perfectly equitable, modern LLMs leveraged through thoughtful prompting strategies tend to demonstrate more fairness than many conventional models, especially those trained on large datasets where biases may be amplified. The analysis also suggests that finetuning, while potentially improving accuracy on a specific task, requires careful monitoring to avoid worsening fairness. These findings underscore the critical importance of not only evaluating predictive accuracy but also rigorously assessing the ethical implications and potential biases of any model before its consideration for clinical deployment.

\section{Analysis of Model Reliability and Failure-to-Predict Rates}\label{sec:failure}
Beyond predictive accuracy, a model's reliability in following instructions is crucial for practical deployment. We systematically quantified the ``failure-to-predict'' rate for each LLM, defined as instances where the model either explicitly refused to answer or produced an output that did not contain the required prediction. The results, detailed in Table~\ref{tab:failure_to_predict_rates}, reveal important patterns regarding model robustness and prompt sensitivity.

\begin{table}[!ht]
\centering
\caption{\textit{Failure-to-predict rates (\%) of LLMs across different tasks, datasets, and settings.} A failure is defined as an instance where the model's output did not contain the required prediction logits.}
\label{tab:failure_to_predict_rates}
\resizebox{1.0\textwidth}{!}{
\begin{tabular}{c|c|c|c|c|c|c|c}
\toprule
\textbf{Method} & \textbf{Modality} & \textbf{Setting} & \textbf{TJH Outcome} & \textbf{TJH LOS} & \textbf{MIMIC-IV Outcome} & \textbf{MIMIC-IV Readmission} & \textbf{MIMIC-III Outcome} \\
\midrule
\multirow{6}{*}{OpenBioLLM-8B} & Unstructured Note & prompt & - & - & 49.00 & 30.50 & 6.00 \\ \cmidrule{2-8}
 & \multirow{3}{*}{Structured EHR} & base prompt & 1.00 & 1.00 & 0.50 & 24.50 & - \\
 &  & optimized prompt & 17.00 & 17.50 & 10.50 & 8.50 & - \\
 &  & opt.+ICL & 3.00 & 4.50 & 3.00 & 0.50 & - \\ \cmidrule{2-8}
 & Multimodal & optimized prompt & - & - & 7.50 & 11.00 & - \\ \midrule
\multirow{6}{*}{Qwen2.5-7B} & Unstructured Note & prompt & - & - & 47.50 & 42.00 & 2.00 \\ \cmidrule{2-8}
 & \multirow{3}{*}{Structured EHR} & base prompt & 1.50 & 2.50 & 0.50 & 84.00 & - \\
 &  & optimized prompt & 1.00 & 5.50 & 1.00 & 4.00 & - \\
 &  & opt.+ICL & 2.00 & 2.50 & 0.50 & 0.50 & - \\ \cmidrule{2-8}
 & Multimodal & optimized prompt & - & - & 5.50 & 4.00 & - \\ \midrule
\multirow{6}{*}{Gemma-3-4B} & Unstructured Note & prompt & - & - & 11.50 & 9.00 & 0.00 \\ \cmidrule{2-8}
 & \multirow{3}{*}{Structured EHR} & base prompt & 0.50 & 0.50 & 0.00 & 72.00 & - \\
 &  & optimized prompt & 0.00 & 0.00 & 0.00 & 0.00 & - \\
 &  & opt.+ICL & 0.00 & 0.00 & 0.00 & 0.00 & - \\ \cmidrule{2-8}
 & Multimodal & optimized prompt & - & - & 0.00 & 0.00 & - \\ \midrule
\multirow{6}{*}{DeepSeek-V3.1} & Unstructured Note & prompt & - & - & 0.00 & 0.00 & 0.00 \\ \cmidrule{2-8}
 & \multirow{3}{*}{Structured EHR} & base prompt & 0.00 & 1.50 & 0.00 & 0.00 & - \\
 &  & optimized prompt & 0.00 & 0.00 & 0.00 & 0.00 & - \\
 &  & opt.+ICL & 0.00 & 0.00 & 0.00 & 0.00 & - \\ \cmidrule{2-8}
 & Multimodal & optimized prompt & - & - & 0.00 & 0.00 & - \\ \midrule
\multirow{6}{*}{GPT-4o} & Unstructured Note & prompt & - & - & 2.00 & 1.50 & 1.00 \\ \cmidrule{2-8}
 & \multirow{3}{*}{Structured EHR} & base prompt & 0.00 & 0.50 & 0.00 & 0.50 & - \\
 &  & optimized prompt & 3.00 & 1.50 & 1.00 & 0.50 & - \\
 &  & opt.+ICL & 0.00 & 1.00 & 0.00 & 0.00 & - \\ \cmidrule{2-8}
 & Multimodal & optimized prompt & - & - & 0.00 & 0.00 & - \\ \midrule
\multirow{6}{*}{HuatuoGPT-o1-7B} & Unstructured Note & prompt & - & - & 27.00 & 28.00 & 0.00 \\ \cmidrule{2-8}
 & \multirow{3}{*}{Structured EHR} & base prompt & 2.50 & 2.50 & 2.50 & 89.00 & - \\
 &  & optimized prompt & 0.00 & 0.00 & 0.00 & 0.00 & - \\
 &  & opt.+ICL & 22.00 & 17.50 & 2.50 & 0.50 & - \\ \cmidrule{2-8}
 & Multimodal & optimized prompt & - & - & 0.00 & 0.00 & - \\ \midrule
\multirow{6}{*}{DeepSeek-R1-7B} & Unstructured Note & prompt & - & - & 55.00 & 57.50 & 1.50 \\ \cmidrule{2-8}
 & \multirow{3}{*}{Structured EHR} & base prompt & 16.00 & 70.00 & 48.50 & 92.50 & - \\
 &  & optimized prompt & 12.00 & 14.50 & 0.00 & 0.00 & - \\
 &  & opt.+ICL & 84.00 & 76.00 & 53.00 & 55.00 & - \\ \cmidrule{2-8}
 & Multimodal & optimized prompt & - & - & 1.00 & 0.50 & - \\ \midrule
\multirow{6}{*}{DeepSeek-R1} & Unstructured Note & prompt & - & - & 11.00 & 27.00 & 15.00 \\ \cmidrule{2-8}
 & \multirow{3}{*}{Structured EHR} & base prompt & 1.00 & 1.50 & 0.00 & 0.50 & - \\
 &  & optimized prompt & 0.50 & 2.00 & 0.00 & 0.00 & - \\
 &  & opt.+ICL & 0.50 & 3.50 & 1.00 & 1.00 & - \\ \cmidrule{2-8}
 & Multimodal & optimized prompt & - & - & 4.00 & 6.50 & - \\ \midrule
 \multirow{6}{*}{DeepSeek-V3.1-Think} & Unstructured Note & prompt & - & - & 0.00 & 0.00 & 0.00 \\ \cmidrule{2-8}
 & \multirow{3}{*}{Structured EHR} & base prompt & 2.00 & 1.00 & 0.50 & 0.50 & - \\
 &  & optimized prompt & 3.00 & 0.50 & 0.50 & 0.00 & - \\
 &  & opt.+ICL & 1.50 & 0.50 & 0.00 & 0.00 & - \\ \cmidrule{2-8}
 & Multimodal & optimized prompt & - & - & 0.00 & 2.00 & - \\ \midrule
\multirow{6}{*}{o3-mini-high} & Unstructured Note & prompt & - & - & 4.00 & 0.00 & 0.00 \\ \cmidrule{2-8}
 & \multirow{3}{*}{Structured EHR} & base prompt & 60.50 & 15.00 & 0.50 & 3.50 & - \\
 &  & optimized prompt & 0.00 & 0.00 & 0.00 & 0.00 & - \\
 &  & opt.+ICL & 18.50 & 0.00 & 0.00 & 1.00 & - \\ \cmidrule{2-8}
 & Multimodal & optimized prompt & - & - & 0.00 & 0.00 & - \\ \midrule
\multirow{6}{*}{GPT-5} & Unstructured Note & prompt & - & - & 0.00 & 0.00 & 0.00 \\ \cmidrule{2-8}
 & \multirow{3}{*}{Structured EHR} & base prompt & 0.00 & 0.00 & 0.00 & 0.00 & - \\
 &  & optimized prompt & 0.00 & 0.00 & 0.00 & 0.00 & - \\
 &  & opt.+ICL & 0.00 & 0.00 & 0.00 & 0.00 & - \\ \cmidrule{2-8}
 & Multimodal & optimized prompt & - & - & 0.00 & 0.00 & - \\
 
\bottomrule
\end{tabular}
}
\end{table}

\begin{itemize}
    \item \textbf{Model capability is a primary factor.} There is a clear correlation between a model's general capability and its reliability. State-of-the-art models such as GPT-5, DeepSeek-V3.1, and Gemma-3-4B (with optimized prompts) demonstrated very low to zero failure rates across nearly all tasks. This indicates their superior instruction-following ability and robustness in handling complex clinical data formats.
    
    \item \textbf{Smaller or less advanced models struggle more.} Conversely, models such as OpenBioLLM-8B and Qwen2.5-7B exhibited higher failure rates, particularly on the unstructured clinical note tasks (e.g., 49.0\% for OpenBioLLM and 47.5\% for Qwen2.5-7B on MIMIC-IV Outcome). This suggests that parsing narrative clinical text while adhering to a structured output format is a significant challenge for these models.

    \item \textbf{Prompt design significantly impacts reliability.} The choice of prompt had a substantial effect on failure rates, especially for structured EHR data. For instance, for the MIMIC-IV Readmission task, using an ``optimized prompt'' drastically reduced the failure rate for HuatuoGPT-o1-7B from 89.0\% (``base prompt'') to 0.0\%. Similarly, for Gemma-3-4B, the rate dropped from 72.0\% to 0.0\%. This demonstrates the critical importance of effective prompt engineering. However, increasing prompt complexity with in-context learning (``opt.+ICL'') sometimes had a negative effect. For instance, HuatuoGPT-o1-7B's failure rate on the TJH Outcome task increased from 0.0\% (``optimized prompt'') to 22.0\% (``opt.+ICL''), suggesting that overly complex prompts can sometimes confuse certain models.
    
    \item \textbf{Specific models exhibit unique failure modes.} DeepSeek-R1-7B showed exceptionally high failure rates in several settings (e.g., 92.5\% on MIMIC-IV Readmission with the base prompt, and 84.0\% on TJH Outcome with ICL). Our qualitative analysis suggests this is not necessarily a failure to reason about the task, but rather an issue of excessive verbosity. The model's detailed reasoning process was often so lengthy that it exceeded the maximum token limit for the output, preventing the final prediction from being generated. This is a distinct failure mode related to output control rather than comprehension.

    \item \textbf{Task difficulty influences failure rates.} Certain task and prompt combinations appeared to be more challenging for the models. For example, the MIMIC-IV Readmission task using the ``base prompt'' resulted in very high failure rates across multiple models (e.g., Qwen2.5-7B: 84.0\%, Gemma-3-4B: 72.0\%, HuatuoGPT-o1-7B: 89.0\%). This may point to ambiguities in the base prompt for this specific task that were successfully addressed by the optimized prompt.
\end{itemize}

\section{Data Contamination and Timeline Analysis}\label{sec:appendix_contamination}

To ensure the integrity of our benchmark, it is critical to consider the potential for data contamination, where a model might have been trained on or exposed to the benchmark's test data. This is particularly relevant for large-scale proprietary models with unknown training corpora. To provide full transparency, we have compiled a detailed timeline of the release dates for the specific dataset versions used in our study alongside the training data cutoff dates for the evaluated models (Table~\ref{tab:dataset_model_timelines}).

Furthermore, two key factors significantly mitigate the risk of contamination in our study:

\begin{itemize}
\item \textbf{Access-Controlled Nature of MIMIC Datasets:} The MIMIC family of datasets is not publicly available on the open web. Access requires credentialed login, completion of mandatory ethics training (the CITI program), and adherence to a strict data use agreement under PhysioNet Credentialed Health Data License (\url{https://physionet.org/news/post/llm-responsible-use}). This controlled access makes it highly improbable that these datasets were included in the large-scale web crawls typically used for pre-training large language models.
\item \textbf{Mismatch of Data Modality and Task Formulation.} General-purpose LLMs are primarily pre-trained on vast quantities of unstructured, natural language text (e.g., books, articles, websites). They also have not been explicitly trained to perform non-generative prediction tasks. Our evaluation on structured EHR or clinical notes data requires the model to interpret a format it was not designed for and to perform a discriminative task, not a generative one. Furthermore, our specific prompting strategies and required JSON output formats (including reasoning) are highly idiosyncratic to our study. It is improbable that the models have encountered and memorized this exact task format during pre-training. Their strong performance is therefore more indicative of genuine in-context reasoning and generalization ability rather than memorization.
\end{itemize}

In conclusion, while it is impossible to definitively rule out any data leakage for language models, the combination of stringent data access controls for MIMIC, the fundamental mismatch in data modality and task design provides a multi-faceted argument that our zero-shot results are a valid reflection of the models' reasoning capabilities, not an artifact of data contamination.

\begin{table*}[!ht]
\centering
\caption{\textit{Timelines of datasets and models used in the benchmark.} This table details the release dates for the specific versions of the datasets utilized in our study and the training data cutoff and release dates for the evaluated models. }
\label{tab:dataset_model_timelines}
\centering
\footnotesize
\resizebox{1.0\textwidth}{!}{
\begin{tabular}{lllcc}
\toprule
\textbf{Category} & \textbf{Name} & \textbf{Version / Identifier} & \textbf{Release Date} & \textbf{Estimated Data Cutoff Date} \\ \midrule
\multirow{4}{*}{Datasets}
 & MIMIC-III & v1.4 & 2016-09 & - \\
 & MIMIC-IV & v3.1 & 2024-11 & - \\
 & MIMIC-IV-Note & v2.2 & 2023-01 & - \\
 & TJH & N/A & 2020-05 & - \\ \midrule

\multirow{6}{*}{Proprietary LLMs}
 & DeepSeek-R1 & \texttt{deepseek-r1-250528} & 2025-05 & 2025-03 \\
 & DeepSeek-V3.1 & \texttt{deepseek-v3.1} & 2025-08 & 2025-07 \\
 & DeepSeek-V3.1-Think & \texttt{deepseek-v3.1-think} & 2025-08 & 2025-07 \\
 & GPT-4o & \texttt{gpt-4o-2025-04-10} & 2025-04 & 2024-06 \\
 & o3-mini-high & \texttt{o3-mini-2025-01-31} & 2025-01 & 2023-10 \\ 
 & GPT-5 (High) & \texttt{gpt-5-2025-08-07} & 2025-08 & 2024-09 \\ \midrule 

\multirow{5}{*}{Open-Source LLMs}
 & OpenBioLLM-8B & \texttt{OpenBioLLM-Llama3-8B} & 2024-05 & 2023-12 \\
 & Qwen2.5-7B & \texttt{Qwen2.5-7B-Instruct} & 2024-12 & 2023-12 \\
 & Gemma-3-4B & \texttt{gemma-3-4b-it} & 2025-03 & 2024-08 \\
 & HuatuoGPT-o1-7B & \texttt{HuatuoGPT-o1-7B} & 2024-12 & 2023-12 \\ 
 & DeepSeek-R1-7B & \texttt{DeepSeek-R1-Distill-Qwen-7B} & 2025-01 & 2024-07 \\ \midrule

\multirow{5}{*}{BERT-style Models}
 & BERT & \texttt{bert-base-uncased} & 2018-10 & 2018-10 \\
 & ClinicalBERT & \texttt{medicalai/ClinicalBERT} & 2019-04 & 2012-12 \\
 & BioBERT & \texttt{pritamdeka/BioBert-PubMed200kRCT} & 2019-09 & 2019-01 \\
 & GatorTron & \texttt{UFNLP/gatortron-base} & 2022-02 & 2021-12 \\
 & Clinical-Longformer & \texttt{yikuan8/Clinical-Longformer} & 2023-01 & 2012-12 \\
\bottomrule
\end{tabular}
}
\end{table*}

\clearpage
\section{Supplementary Tables}
\renewcommand{\thetable}{S\arabic{table}}
\setcounter{table}{0}

\begin{table}[!ht]
\footnotesize
\centering
\caption{\textit{Statistics of the TJH dataset, MIMIC-IV EHR (structured EHR data), and MIMIC-IV Note (clinical notes) datasets.} ``Re.'' stands for Readmission, indicating patients who are readmitted to the ICU within 30 days of discharge, while ``No Re.'' represents patients who are not readmitted. ``LOS'' denotes ``length-of-stay''.}
\label{tab:dataset_stats}
\resizebox{\linewidth}{!}{
\begin{tabular}{lcccccccccccccccc}
\toprule
\multicolumn{1}{c}{\multirow{2}{*}{\textbf{Dataset}}}          
& \multicolumn{3}{c}{\textbf{TJH}} 
& \multicolumn{3}{c}{\textbf{MIMIC-III Note}} 
& \multicolumn{5}{c}{\textbf{MIMIC-IV EHR}} 
& \multicolumn{5}{c}{\textbf{MIMIC-IV Note}} \\
\cmidrule(lr){2-4} \cmidrule(lr){5-7} \cmidrule(lr){8-12} \cmidrule(lr){13-17}
                & Total & Alive & Dead
                & Total & Alive & Dead
                & Total & Alive & Dead & Re. & No Re. 
                & Total & Alive & Dead & Re. & No Re. \\
\midrule
& \multicolumn{16}{c}{\textit{Test Set Statistics}} \\
\midrule
\# Patients     & 200 & 109 & 91 & 200 & 180 & 20 & 200 & 183 & 17 & 53 & 147 & 200 & 183 & 17 & 53 & 147 \\
\# Total visits & 967 & 601 & 366 & 200 & 180 & 20 & 801 & 717 & 84 & 274 & 527 & 200 & 183 & 17 & 53 & 147 \\
\# Avg. visits  & 4.8 & 5.5 & 4.0 & 1.0 & 1.0 & 1.0 & 4.0 & 3.9 & 4.9 & 5.2 & 3.6 & 1.0 & 1.0 & 1.0 & 1.0 & 1.0 \\
Avg. LOS        & 7.1 & 7.8 & 5.8 & - & - & - & - & - & - & - & - & - & - & - & - & - \\
\midrule
& \multicolumn{16}{c}{\textit{Training Set Statistics}} \\
\midrule
\# Patients     & 140 & 75 & 65 & 8750 & 7887 & 863 & 8750 & 8028 & 722 & 2112 & 6638 & 8750 & 8028 & 722 & 2112 & 6638 \\
\# Total visits & 641 & 395 & 246 & 8750 & 7887 & 863 & 33423 & 30117 & 3306 & 10448 & 22975 & 8750 & 8028 & 722 & 2112 & 6638 \\
\# Avg. visits  & 4.6 & 5.3 & 3.8 & 1.0 & 1.0 & 1.0 & 3.8 & 3.8 & 4.6 & 4.9 & 3.5 & 1.0 & 1.0 & 1.0 & 1.0 & 1.0 \\
Avg. LOS        & 8.6 & 9.0 & 8.0 & & & & - & - & - & - & - & - & - & - & - & - \\
\midrule
& \multicolumn{16}{c}{\textit{Validation Set Statistics}} \\
\midrule
\# Patients     & 21 & 11 & 10 & 1250 & 1127 & 123 & 1250 & 1147 & 103 & 305 & 945 & 1250 & 1147 & 103 & 305 & 945 \\
\# Total visits & 96 & 54 & 42 & 1250 & 1127 & 123 & 4685 & 4176 & 509 & 1522 & 3163 & 1250 & 1147 & 103 & 305 & 945 \\
\# Avg. visits  & 4.6 & 4.9 & 4.2 & 1.0 & 1.0 & 1.0 & 3.7 & 3.6 & 4.9 & 5.0 & 3.3 & 1.0 & 1.0 & 1.0 & 1.0 & 1.0 \\
Avg. LOS        & 6.5 & 8.1 & 4.4 & - & - & - & - & - & - & - & - & - & - & - & - & - \\
\bottomrule
\end{tabular}
}
\end{table}
\clearpage

\captionof{table}{\textit{The detailed task descriptions for various predictive tasks, including in-hospital mortality, 30-day readmission and length-of-stay.} These task descriptions modify the instruction part of our designed prompt templates.}
\label{tab:task_description_and_response}
\begin{center}
    \begin{tcolorbox}[colback=lightbluebg!30!white, colframe=blueframe, breakable, 
        title=The detailed task descriptions for various predictive tasks
    ]
    \begin{VerbatimWrap}
# (1) In-hospital mortality prediction (for unstructured clinical notes): Your primary task is to assess the provided ICU discharge summary to determine the likelihood of the patient not surviving their current hospital stay (in-hospital mortality).

# (2) 30-day readmission prediction (for unstructured clinical notes): Your primary task is to assess the provided ICU discharge summary to predict the probability of unplanned hospital readmission OR death within 30 days following hospital discharge.

# (3) In-hospital mortality prediction (for structured EHR): Your primary task is to assess the provided medical data and analyze the health records from ICU visits to determine the likelihood of the patient not surviving their hospital stay.

# (4) 30-day readmission prediction (for structured EHR): Your primary task is to analyze the medical data to predict the probability of readmission within 30 days post-discharge. Include cases where a patient passes away within 30 days from the discharge date as readmissions.

# (5) Length-of-stay prediction (for structured EHR): Your primary task is to analyze the medical data to predict the length of stay (LOS) in the hospital. The LOS is defined as the number of days from admission to discharge, including any days spent in the ICU.

\end{VerbatimWrap}

    \end{tcolorbox}
\end{center}
\clearpage

\captionof{table}{\textit{The prompt template for the mortality prediction task on the MIMIC-III dataset with clinical notes data.}}
\label{tab:mimic3_note_mortality}
\begin{center}
    \begin{tcolorbox}[
        colback=lightbluebg!30!white, colframe=blueframe, 
        breakable, 
        title=The prompt template for the mortality prediction task on the MIMIC-III dataset with clinical notes data
    ]
        \begin{VerbatimWrap}
You are an experienced critical care physician working in an Intensive Care Unit (ICU). You are skilled in interpreting ICU admission summaries of patients, and predicting clinical outcomes based on the patient's status at the time of admission and their overall hospital course.

Your primary task is to assess the provided ICU admission summary to determine the likelihood of the patient not surviving their current hospital stay (in-hospital mortality).

Please first perform a step-by-step analysis of the patient data presented in the summary note. Then, provide a final assessment of the likelihood of in-hospital mortality.

Your final output must be a JSON object containing two keys:
1.  `"think"`: A string containing your detailed step-by-step clinical reasoning (under 500 words). Focus on the factors influencing the mortality risk based on the summary note.
2.  `"answer"`: A floating-point number between 0.0 and 1.0 representing the predicted probability of in-hospital mortality (higher value means higher likelihood of death).

Example Format: ```json { "think": "The patient was admitted with severe sepsis, required prolonged ventilation. Although weaned off pressors, renal function remains poor and discharge from ICU is to the palliative care ward. Key risk factors include persistent organ dysfunction and goals of care discussion outcomes. Overall assessment suggests a very high risk of not surviving this hospitalization.", "answer": 0.90 }
\end{VerbatimWrap}
    \end{tcolorbox}
\end{center}
\clearpage

\captionof{table}{\textit{The prompt template for the mortality prediction task on the MIMIC-IV dataset with clinical notes data.}}
\label{tab:mimic4_note_mortality}
\begin{center}
    \begin{tcolorbox}[
        colback=lightbluebg!30!white, colframe=blueframe, 
        breakable, 
        title=The prompt template for the mortality prediction task on the MIMIC-IV dataset with clinical notes data
    ]
        \begin{VerbatimWrap}
You are an experienced critical care physician working in an Intensive Care Unit (ICU). You are skilled in interpreting ICU discharge summaries of patients, and predicting clinical outcomes based on the patient's status at the time of discharge and their overall hospital course.

Your primary task is to assess the provided ICU discharge summary to determine the likelihood of the patient not surviving their current hospital stay (in-hospital mortality).

Please first perform a step-by-step analysis of the patient data presented in the discharge summary. Then, provide a final assessment of the likelihood of in-hospital mortality.

Your final output must be a JSON object containing two keys:
1.  `"think"`: A string containing your detailed step-by-step clinical reasoning (under 500 words). Focus on the factors influencing the mortality risk based on the discharge note.
2.  `"answer"`: A floating-point number between 0.0 and 1.0 representing the predicted probability of in-hospital mortality (higher value means higher likelihood of death).

Example Format: ```json { "think": "The patient was admitted with severe sepsis, required prolonged ventilation. Although weaned off pressors, renal function remains poor and discharge from ICU is to the palliative care ward. Key risk factors include persistent organ dysfunction and goals of care discussion outcomes. Overall assessment suggests a very high risk of not surviving this hospitalization.", "answer": 0.90 }```
\end{VerbatimWrap}
    \end{tcolorbox}
\end{center}
\clearpage

\captionof{table}{\textit{The prompt template for the readmission prediction task on the MIMIC-IV dataset with clinical notes data.}}
\label{tab:mimic4_note_readmission}
\begin{center} 
    \begin{tcolorbox}[
        colback=lightbluebg!30!white, colframe=blueframe, 
        breakable, 
        title=The prompt template for the readmission prediction task on the MIMIC-IV dataset with clinical notes data,
    ]
        \begin{VerbatimWrap}
You are an experienced critical care physician working in an Intensive Care Unit (ICU). You are skilled in interpreting ICU discharge summaries of patients, and predicting clinical outcomes based on the patient's status at the time of discharge and their overall hospital course.

Your primary task is to assess the provided ICU discharge summary to predict the probability of unplanned hospital readmission OR death within 30 days following hospital discharge.

Please first perform a step-by-step analysis of the patient data presented in the discharge summary. Then, provide a final assessment of the predicted probability of 30-day readmission or mortality.

Your final output must be a JSON object containing two keys:
1.  `"think"`: A string containing your detailed step-by-step clinical reasoning (under 500 words). Focus on the factors influencing the 30-day post-discharge risk.
2.  `"answer"`: A floating-point number between 0.0 and 1.0 representing the predicted probability of 30-day readmission or death (higher value means higher likelihood of the event).

Example Format: ```json { "think": "Patient discharged after prolonged ICU stay for exacerbation of severe COPD. Discharged on home oxygen and multiple new medications. While stable at discharge, patient has poor baseline function, multiple comorbidities (CHF, CKD), and a history of frequent admissions. Limited social support noted. High risk for decompensation or needing further acute care within 30 days.", "answer": 0.65 }```
\end{VerbatimWrap}
    \end{tcolorbox}
\end{center}
\clearpage

\captionof{table}{\textit{The base-setting prompt template for the mortality prediction task on the TJH dataset with structured EHR data.}}
\label{tab:tjh_base_full_ehr_prompt}
\begin{center} 
    \begin{tcolorbox}[colback=lightbluebg!30!white, colframe=blueframe, breakable, 
        title=The base-setting prompt template for the mortality prediction task on the TJH dataset with structured EHR data
    ]
        \begin{VerbatimWrap}
You are an experienced doctor specializing in COVID-19 treatment, skilled in interpreting longitudinal patient data and predicting clinical outcomes.

I will provide you with longitudinal medical information for a patient. The data covers 5 visits/time points that occurred at 2020-01-23, 2020-01-30, 2020-02-04, 2020-02-05, 2020-02-06.
Each clinical feature is presented as a list of values, corresponding to these time points. Missing values are represented as `NaN` for numerical values and "unknown" for categorical values. Note that units and reference ranges are provided alongside relevant features.

Patient Background:
- Sex: female
- Age: 70.0 years

Your Task:
Your primary task is to assess the provided medical data and analyze the health records from ICU visits to determine the likelihood of the patient not surviving their hospital stay.

Instructions & Output Format:
Please first perform a step-by-step analysis of the patient data, considering trends, abnormal values relative to reference ranges, and their clinical significance for survival. Then, provide a final assessment of the likelihood of not surviving the hospital stay.

Your final output must be a JSON object containing two keys:
1.  `"think"`: A string containing your detailed step-by-step clinical reasoning (under 500 words).
2.  `"answer"`: A floating-point number between 0 and 1 representing the predicted probability of mortality (higher value means higher likelihood of death).

Example Format: ```json { "think": "The patient presents with worsening X, stable Y, and improved Z. Factor A is a major risk indicator... Overall assessment suggests a high risk.", "answer": 0.85 }```

Handling Uncertainty:
In situations where the provided data is clearly insufficient or too ambiguous to make a reasonable prediction, respond with the exact phrase: `I do not know`

Now, please analyze and predict for the following patient:

Clinical Features Over Time:
- Hypersensitive cardiac troponinI: [NaN, NaN, NaN, NaN, NaN]
- hemoglobin: [109.0, 112.0, NaN, 126.0, NaN]
- Serum chloride: [99.1, 102.9, NaN, 102.2, NaN]
- Prothrombin time: [13.6, NaN, NaN, NaN, NaN]
- procalcitonin: [0.06, 0.06, NaN, NaN, NaN]
- eosinophils(%): [0.0, 0.2, NaN, 0.1, NaN]
- Interleukin 2 receptor: [591.0, NaN, NaN, NaN, NaN]
- Alkaline phosphatase: [47.0, 61.0, NaN, 69.0, NaN]
- albumin: [34.9, 34.0, NaN, 38.4, NaN]
- basophil(%): [0.0, 0.2, NaN, 0.1, NaN]
- Interleukin 10: [8.1, NaN, NaN, NaN, NaN]
- Total bilirubin: [7.5, 7.7, NaN, 12.6, NaN]
- Platelet count: [169.0, 275.0, NaN, 238.0, NaN]
- monocytes(%): [5.9, 6.4, NaN, 6.1, NaN]
- antithrombin: [84.0, NaN, NaN, NaN, NaN]
- Interleukin 8: [21.9, NaN, NaN, NaN, NaN]
- indirect bilirubin: [3.8, 3.7, NaN, 7.5, NaN]
- Red blood cell distribution width: [12.8, 12.6, NaN, NaN, NaN]
- neutrophils(%): [75.0, 60.1, NaN, 66.4, NaN]
- total protein: [68.3, 62.2, NaN, 68.2, NaN]
- Quantification of Treponema pallidum antibodies: [0.07, NaN, NaN, NaN, NaN]
- Prothrombin activity: [94.0, NaN, NaN, NaN, NaN]
- HBsAg: [0.01, NaN, NaN, NaN, NaN]
- mean corpuscular volume: [94.6, 93.6, NaN, 94.4, NaN]
- hematocrit: [33.2, 33.7, NaN, 35.7, NaN]
- White blood cell count: [3.72, 5.31, NaN, 7.68, NaN]
- Tumor necrosis factorα: [10.4, NaN, NaN, NaN, NaN]
- mean corpuscular hemoglobin concentration: [328.0, 332.0, NaN, 353.0, NaN]
- fibrinogen: [5.33, NaN, NaN, NaN, NaN]
- Interleukin 1β: [5.0, NaN, NaN, NaN, NaN]
- Urea: [2.9, 3.7, NaN, 4.22, NaN]
- lymphocyte count: [0.71, 1.76, NaN, 2.1, NaN]
- PH value: [NaN, NaN, NaN, NaN, NaN]
- Red blood cell count: [3.51, 3.6, NaN, 3.78, NaN]
- Eosinophil count: [0.0, 0.01, NaN, 0.01, NaN]
- Corrected calcium: [2.17, 2.25, NaN, 2.32, NaN]
- Serum potassium: [3.34, 3.34, NaN, 3.9, NaN]
- glucose: [9.02, 9.42, NaN, NaN, NaN]
- neutrophils count: [2.79, 3.19, NaN, 5.09, NaN]
- Direct bilirubin: [3.7, 4.0, NaN, 5.1, NaN]
- Mean platelet volume: [11.6, 10.7, NaN, 9.7, NaN]
- ferritin: [567.2, NaN, NaN, NaN, NaN]
- RBC distribution width SD: [44.4, 42.7, NaN, NaN, NaN]
- Thrombin time: [16.7, NaN, NaN, NaN, NaN]
- (%)lymphocyte: [19.1, 33.1, NaN, 27.3, NaN]
- HCV antibody quantification: [0.06, NaN, NaN, NaN, NaN]
- DD dimer: [0.98, NaN, NaN, NaN, NaN]
- Total cholesterol: [3.28, 3.49, NaN, 4.47, NaN]
- aspartate aminotransferase: [30.0, 30.0, NaN, 20.0, NaN]
- Uric acid: [151.0, 135.0, NaN, 221.1, NaN]
- HCO3: [22.3, 24.6, NaN, 25.3, NaN]
- calcium: [2.07, 2.13, NaN, 2.29, NaN]
- Aminoterminal brain natriuretic peptide precursor(NTproBNP): [NaN, NaN, NaN, NaN, NaN]
- Lactate dehydrogenase: [328.0, 269.0, NaN, 226.0, NaN]
- platelet large cell ratio: [37.8, 30.0, NaN, 21.4, NaN]
- Interleukin 6: [47.82, NaN, NaN, NaN, NaN]
- Fibrin degradation products: [4.0, NaN, NaN, NaN, NaN]
- monocytes count: [0.22, 0.34, NaN, 0.47, NaN]
- PLT distribution width: [13.9, 12.5, NaN, 10.1, NaN]
- globulin: [33.4, 28.2, NaN, 29.8, NaN]
- γglutamyl transpeptidase: [21.0, 54.0, NaN, 53.0, NaN]
- International standard ratio: [1.04, NaN, NaN, NaN, NaN]
- basophil count(#): [0.0, 0.01, NaN, 0.01, NaN]
- mean corpuscular hemoglobin: [31.1, 31.1, NaN, 33.3, NaN]
- Activation of partial thromboplastin time: [34.8, NaN, NaN, NaN, NaN]
- High sensitivity Creactive protein: [42.3, 3.6, NaN, NaN, NaN]
- HIV antibody quantification: [0.1, NaN, NaN, NaN, NaN]
- serum sodium: [135.7, 140.4, NaN, 143.2, NaN]
- thrombocytocrit: [0.2, 0.29, NaN, 0.23, NaN]
- ESR: [66.0, 29.0, NaN, NaN, NaN]
- glutamicpyruvic transaminase: [19.0, 84.0, NaN, 67.0, NaN]
- eGFR: [77.2, 90.0, NaN, 84.6, NaN]
- creatinine: [69.0, 58.0, NaN, 64.0, NaN]
\end{VerbatimWrap}
    \end{tcolorbox}
\end{center}
\clearpage

\captionof{table}{\textit{The optimized-setting prompt template for the mortality prediction task on the TJH dataset with structured EHR data.}}
\label{tab:tjh_optimized_full_ehr_prompt}
\begin{center} 
    \begin{tcolorbox}[colback=lightbluebg!30!white, colframe=blueframe, breakable, 
        title=The optimized-setting prompt template for the mortality prediction task on the TJH dataset with structured EHR data
    ]
        \begin{VerbatimWrap}
You are an experienced doctor specializing in COVID-19 treatment, skilled in interpreting longitudinal patient data and predicting clinical outcomes.

I will provide you with longitudinal medical information for a patient. The data covers 5 visits/time points that occurred at 2020-01-23, 2020-01-30, 2020-02-04, 2020-02-05, 2020-02-06.
Each clinical feature is presented as a list of values, corresponding to these time points. Missing values are represented as `NaN` for numerical values and "unknown" for categorical values. Note that units and reference ranges are provided alongside relevant features.

Patient Background:
- Sex: female
- Age: 70.0 years

Your Task:
Your primary task is to assess the provided medical data and analyze the health records from ICU visits to determine the likelihood of the patient not surviving their hospital stay.

Instructions & Output Format:
Please first perform a step-by-step analysis of the patient data, considering trends, abnormal values relative to reference ranges, and their clinical significance for survival. Then, provide a final assessment of the likelihood of not surviving the hospital stay.

Your final output must be a JSON object containing two keys:
1.  `"think"`: A string containing your detailed step-by-step clinical reasoning (under 500 words).
2.  `"answer"`: A floating-point number between 0 and 1 representing the predicted probability of mortality (higher value means higher likelihood of death).

Example Format: ```json { "think": "The patient presents with worsening X, stable Y, and improved Z. Factor A is a major risk indicator... Overall assessment suggests a high risk.", "answer": 0.85 }```

Handling Uncertainty:
In situations where the provided data is clearly insufficient or too ambiguous to make a reasonable prediction, respond with the exact phrase: `I do not know`

Now, please analyze and predict for the following patient:

Clinical Features Over Time:
- Hypersensitive cardiac troponinI (Unit: ng/L. Reference range: less than 14.): [NaN, NaN, NaN, NaN, NaN]
- hemoglobin (Unit: g/L. Reference range: 140 - 180 for men, 120 - 160 for women.): [109.0, 112.0, NaN, 126.0, NaN]
- Serum chloride (Unit: mmol/L. Reference range: 96 - 106.): [99.1, 102.9, NaN, 102.2, NaN]
- Prothrombin time (Unit: seconds. Reference range: 13.1 - 14.125.): [13.6, NaN, NaN, NaN, NaN]
- procalcitonin (Unit: ng/mL. Reference range: less than 0.05.): [0.06, 0.06, NaN, NaN, NaN]
- eosinophils(%) (Unit: %. Reference range: 1 - 6.): [0.0, 0.2, NaN, 0.1, NaN]
- Interleukin 2 receptor (Unit: pg/mL. Reference range: less than 625.): [591.0, NaN, NaN, NaN, NaN]
- Alkaline phosphatase (Unit: IU/L. Reference range: 44 - 147.): [47.0, 61.0, NaN, 69.0, NaN]
- albumin (Unit: g/dL. Reference range: 3.5 - 5.5.): [34.9, 34.0, NaN, 38.4, NaN]
- basophil(%) (Unit: %. Reference range: 0.5 - 1.): [0.0, 0.2, NaN, 0.1, NaN]
- Interleukin 10 (Unit: pg/mL. Reference range: less than 9.8.): [8.1, NaN, NaN, NaN, NaN]
- Total bilirubin (Unit: µmol/L. Reference range: 5.1 - 17.): [7.5, 7.7, NaN, 12.6, NaN]
- Platelet count (Unit: x 10^9/L. Reference range: 150 - 450.): [169.0, 275.0, NaN, 238.0, NaN]
- monocytes(%) (Unit: %. Reference range: 2 - 10.): [5.9, 6.4, NaN, 6.1, NaN]
- antithrombin (Unit: %. Reference range: 80 - 120.): [84.0, NaN, NaN, NaN, NaN]
- Interleukin 8 (Unit: pg/mL. Reference range: less than 62.): [21.9, NaN, NaN, NaN, NaN]
- indirect bilirubin (Unit: μmol/L. Reference range: 3.4 - 12.0.): [3.8, 3.7, NaN, 7.5, NaN]
- Red blood cell distribution width (Unit: %. Reference range: 11.5 - 14.5 for men, 12.2 - 16.1 for women.): [12.8, 12.6, NaN, NaN, NaN]
- neutrophils(%) (Unit: %. Reference range: 45 - 70.): [75.0, 60.1, NaN, 66.4, NaN]
- total protein (Unit: g/L. Reference range: 60 - 83.): [68.3, 62.2, NaN, 68.2, NaN]
- Quantification of Treponema pallidum antibodies (Unit: /. Reference range: less than 1.0.): [0.07, NaN, NaN, NaN, NaN]
- Prothrombin activity (Unit: %. Reference range: 70 - 130.): [94.0, NaN, NaN, NaN, NaN]
- HBsAg (Unit: IU/mL. Reference range: 0.0 - 0.01.): [0.01, NaN, NaN, NaN, NaN]
- mean corpuscular volume (Unit: fL. Reference range: 80 - 100.): [94.6, 93.6, NaN, 94.4, NaN]
- hematocrit (Unit: %. Reference range: 40 - 54 for men, 36 - 48 for women.): [33.2, 33.7, NaN, 35.7, NaN]
- White blood cell count (Unit: x 10^9/L. Reference range: 4.5 - 11.0.): [3.72, 5.31, NaN, 7.68, NaN]
- Tumor necrosis factorα (Unit: pg/mL. Reference range: less than 8.1.): [10.4, NaN, NaN, NaN, NaN]
- mean corpuscular hemoglobin concentration (Unit: g/L. Reference range: 320 - 360.): [328.0, 332.0, NaN, 353.0, NaN]
- fibrinogen (Unit: g/L. Reference range: 2 - 4.): [5.33, NaN, NaN, NaN, NaN]
- Interleukin 1β (Unit: pg/mL. Reference range: less than 6.5.): [5.0, NaN, NaN, NaN, NaN]
- Urea (Unit: mmol/L. Reference range: 1.8 - 7.1.): [2.9, 3.7, NaN, 4.22, NaN]
- lymphocyte count (Unit: x 10^9/L. Reference range: 1.0 - 4.8.): [0.71, 1.76, NaN, 2.1, NaN]
- PH value (Unit: /. Reference range: 7.35 - 7.45.): [NaN, NaN, NaN, NaN, NaN]
- Red blood cell count (Unit: x 10^12/L. Reference range: 4.5 - 5.5 for men, 4.0 - 5.0 for women.): [3.51, 3.6, NaN, 3.78, NaN]
- Eosinophil count (Unit: x 10^9/L. Reference range: 0.02 - 0.5.): [0.0, 0.01, NaN, 0.01, NaN]
- Corrected calcium (Unit: mmol/L. Reference range: 2.12 - 2.57.): [2.17, 2.25, NaN, 2.32, NaN]
- Serum potassium (Unit: mmol/L. Reference range: 3.5 - 5.0.): [3.34, 3.34, NaN, 3.9, NaN]
- glucose (Unit: mmol/L. Reference range: 3.9 - 5.6.): [9.02, 9.42, NaN, NaN, NaN]
- neutrophils count (Unit: x 10^9/L. Reference range: 2.0 - 8.0.): [2.79, 3.19, NaN, 5.09, NaN]
- Direct bilirubin (Unit: µmol/L. Reference range: 1.7 - 5.1.): [3.7, 4.0, NaN, 5.1, NaN]
- Mean platelet volume (Unit: fL. Reference range: 7.4 - 11.4.): [11.6, 10.7, NaN, 9.7, NaN]
- ferritin (Unit: ng/mL. Reference range: 24 - 336 for men, 11 - 307 for women.): [567.2, NaN, NaN, NaN, NaN]
- RBC distribution width SD (Unit: fL. Reference range: 40.0 - 55.0.): [44.4, 42.7, NaN, NaN, NaN]
- Thrombin time (Unit: seconds. Reference range: 12 - 19.): [16.7, NaN, NaN, NaN, NaN]
- (%)lymphocyte (Unit: %. Reference range: 20 - 40.): [19.1, 33.1, NaN, 27.3, NaN]
- HCV antibody quantification (Unit: IU/mL. Reference range: 0.04 - 0.08.): [0.06, NaN, NaN, NaN, NaN]
- DD dimer (Unit: mg/L. Reference range: 0 - 0.5.): [0.98, NaN, NaN, NaN, NaN]
- Total cholesterol (Unit: mmol/L. Reference range: less than 5.17.): [3.28, 3.49, NaN, 4.47, NaN]
- aspartate aminotransferase (Unit: U/L. Reference range: 8 - 33.): [30.0, 30.0, NaN, 20.0, NaN]
- Uric acid (Unit: µmol/L. Reference range: 240 - 510 for men, 160 - 430 for women.): [151.0, 135.0, NaN, 221.1, NaN]
- HCO3 (Unit: mmol/L. Reference range: 22 - 29.): [22.3, 24.6, NaN, 25.3, NaN]
- calcium (Unit: mmol/L. Reference range: 2.13 - 2.55.): [2.07, 2.13, NaN, 2.29, NaN]
- Aminoterminal brain natriuretic peptide precursor(NTproBNP) (Unit: pg/mL. Reference range: 0 - 125.): [NaN, NaN, NaN, NaN, NaN]
- Lactate dehydrogenase (Unit: U/L. Reference range: 140 - 280.): [328.0, 269.0, NaN, 226.0, NaN]
- platelet large cell ratio (Unit: %. Reference range: 15 - 35.): [37.8, 30.0, NaN, 21.4, NaN]
- Interleukin 6 (Unit: pg/mL. Reference range: 0 - 7.): [47.82, NaN, NaN, NaN, NaN]
- Fibrin degradation products (Unit: μg/mL. Reference range: 0 - 10.): [4.0, NaN, NaN, NaN, NaN]
- monocytes count (Unit: x 10^9/L. Reference range: 0.32 - 0.58.): [0.22, 0.34, NaN, 0.47, NaN]
- PLT distribution width (Unit: fL. Reference range: 9.2 - 16.7.): [13.9, 12.5, NaN, 10.1, NaN]
- globulin (Unit: g/L. Reference range: 23 - 35.): [33.4, 28.2, NaN, 29.8, NaN]
- γglutamyl transpeptidase (Unit: U/L. Reference range: 7 - 47 for men, 5 - 25 for women.): [21.0, 54.0, NaN, 53.0, NaN]
- International standard ratio (Unit: ratio. Reference range: 0.8 - 1.2.): [1.04, NaN, NaN, NaN, NaN]
- basophil count(#) (Unit: x 10^9/L. Reference range: 0.01 - 0.02.): [0.0, 0.01, NaN, 0.01, NaN]
- mean corpuscular hemoglobin (Unit: pg. Reference range: 27 - 31.): [31.1, 31.1, NaN, 33.3, NaN]
- Activation of partial thromboplastin time (Unit: seconds. Reference range: 22 - 35.): [34.8, NaN, NaN, NaN, NaN]
- High sensitivity Creactive protein (Unit: mg/L. Reference range: 3 - 10.): [42.3, 3.6, NaN, NaN, NaN]
- HIV antibody quantification (Unit: IU/mL. Reference range: 0.08 - 0.11.): [0.1, NaN, NaN, NaN, NaN]
- serum sodium (Unit: mmol/L. Reference range: 135 - 145.): [135.7, 140.4, NaN, 143.2, NaN]
- thrombocytocrit (Unit: %. Reference range: 0.22 - 0.24.): [0.2, 0.29, NaN, 0.23, NaN]
- ESR (Unit: mm/hr. Reference range: less than 15  for men, less than 20 for women.): [66.0, 29.0, NaN, NaN, NaN]
- glutamicpyruvic transaminase (Unit: U/L. Reference range: 0 - 35.): [19.0, 84.0, NaN, 67.0, NaN]
- eGFR (Unit: mL/min/1.73m². Reference range: more than 90.): [77.2, 90.0, NaN, 84.6, NaN]
- creatinine (Unit: µmol/L. Reference range: 61.9 - 114.9 for men, 53 - 97.2 for women.): [69.0, 58.0, NaN, 64.0, NaN]
\end{VerbatimWrap}
    \end{tcolorbox}
\end{center}
\clearpage

\captionof{table}{\textit{The optimized-setting with in-context learning prompt template for the mortality prediction task on the TJH dataset with structured EHR data.}}
\label{tab:tjh_optimized_icl_full_ehr_prompt}
\begin{center} 
    \begin{tcolorbox}[colback=lightbluebg!30!white, colframe=blueframe, breakable, 
        title=The optimized-setting with in-context learning prompt template for the mortality prediction task on the TJH dataset with structured EHR data
    ]
        \begin{VerbatimWrap}
You are an experienced doctor specializing in COVID-19 treatment, skilled in interpreting longitudinal patient data and predicting clinical outcomes.

I will provide you with longitudinal medical information for a patient. The data covers 5 visits/time points that occurred at 2020-01-23, 2020-01-30, 2020-02-04, 2020-02-05, 2020-02-06.
Each clinical feature is presented as a list of values, corresponding to these time points. Missing values are represented as `NaN` for numerical values and "unknown" for categorical values. Note that units and reference ranges are provided alongside relevant features.

Patient Background:
- Sex: female
- Age: 70.0 years

Your Task:
Your primary task is to assess the provided medical data and analyze the health records from ICU visits to determine the likelihood of the patient not surviving their hospital stay.

Instructions & Output Format:
Please first perform a step-by-step analysis of the patient data, considering trends, abnormal values relative to reference ranges, and their clinical significance for survival. Then, provide a final assessment of the likelihood of not surviving the hospital stay.

Your final output must be a JSON object containing two keys:
1.  `"think"`: A string containing your detailed step-by-step clinical reasoning (under 500 words).
2.  `"answer"`: A floating-point number between 0 and 1 representing the predicted probability of mortality (higher value means higher likelihood of death).

Example Format: ```json { "think": "The patient presents with worsening X, stable Y, and improved Z. Factor A is a major risk indicator... Overall assessment suggests a high risk.", "answer": 0.85 }```

Handling Uncertainty:
In situations where the provided data is clearly insufficient or too ambiguous to make a reasonable prediction, respond with the exact phrase: `I do not know`

Example:
Input information of a patient:
The patient is a male, aged 52.0 years.
The patient had 5 visits that occurred at 2020-02-09, 2020-02-10, 2020-02-13, 2020-02-14, 2020-02-17.
Details of the features for each visit are as follows:
- Hypersensitive cardiac troponinI (Unit: ng/L. Reference range: less than 14.): [1.9, 1.9, 1.9, 1.9, 1.9]
- hemoglobin (Unit: g/L. Reference range: 140 - 180 for men, 120 - 160 for women.): [139.0, 139.0, 142.0, 142.0, 142.0]
- Serum chloride (Unit: mmol/L. Reference range: 96 - 106.): [103.7, 103.7, 104.2, 104.2, 104.2]
...... (other features omitted for brevity)

Response:
```json { "think": "The patient is a 52-year-old male. Key labs like Troponin I are consistently normal and low. Hemoglobin is borderline low for a male but stable. Serum chloride is within normal limits and stable. Assuming other unlisted vital signs and labs are also stable or within normal limits, the overall picture suggests a relatively stable condition without indicators of severe organ damage or rapid decline commonly associated with high mortality risk in this context. The risk appears low.", "answer": 0.25 } ```

Now, please analyze and predict for the following patient:

Clinical Features Over Time:
- Hypersensitive cardiac troponinI (Unit: ng/L. Reference range: less than 14.): [NaN, NaN, NaN, NaN, NaN]
- hemoglobin (Unit: g/L. Reference range: 140 - 180 for men, 120 - 160 for women.): [109.0, 112.0, NaN, 126.0, NaN]
- Serum chloride (Unit: mmol/L. Reference range: 96 - 106.): [99.1, 102.9, NaN, 102.2, NaN]
- Prothrombin time (Unit: seconds. Reference range: 13.1 - 14.125.): [13.6, NaN, NaN, NaN, NaN]
- procalcitonin (Unit: ng/mL. Reference range: less than 0.05.): [0.06, 0.06, NaN, NaN, NaN]
- eosinophils(%) (Unit: %. Reference range: 1 - 6.): [0.0, 0.2, NaN, 0.1, NaN]
- Interleukin 2 receptor (Unit: pg/mL. Reference range: less than 625.): [591.0, NaN, NaN, NaN, NaN]
- Alkaline phosphatase (Unit: IU/L. Reference range: 44 - 147.): [47.0, 61.0, NaN, 69.0, NaN]
- albumin (Unit: g/dL. Reference range: 3.5 - 5.5.): [34.9, 34.0, NaN, 38.4, NaN]
- basophil(%) (Unit: %. Reference range: 0.5 - 1.): [0.0, 0.2, NaN, 0.1, NaN]
- Interleukin 10 (Unit: pg/mL. Reference range: less than 9.8.): [8.1, NaN, NaN, NaN, NaN]
- Total bilirubin (Unit: µmol/L. Reference range: 5.1 - 17.): [7.5, 7.7, NaN, 12.6, NaN]
- Platelet count (Unit: x 10^9/L. Reference range: 150 - 450.): [169.0, 275.0, NaN, 238.0, NaN]
- monocytes(%) (Unit: %. Reference range: 2 - 10.): [5.9, 6.4, NaN, 6.1, NaN]
- antithrombin (Unit: %. Reference range: 80 - 120.): [84.0, NaN, NaN, NaN, NaN]
- Interleukin 8 (Unit: pg/mL. Reference range: less than 62.): [21.9, NaN, NaN, NaN, NaN]
- indirect bilirubin (Unit: μmol/L. Reference range: 3.4 - 12.0.): [3.8, 3.7, NaN, 7.5, NaN]
- Red blood cell distribution width (Unit: %. Reference range: 11.5 - 14.5 for men, 12.2 - 16.1 for women.): [12.8, 12.6, NaN, NaN, NaN]
- neutrophils(%) (Unit: %. Reference range: 45 - 70.): [75.0, 60.1, NaN, 66.4, NaN]
- total protein (Unit: g/L. Reference range: 60 - 83.): [68.3, 62.2, NaN, 68.2, NaN]
- Quantification of Treponema pallidum antibodies (Unit: /. Reference range: less than 1.0.): [0.07, NaN, NaN, NaN, NaN]
- Prothrombin activity (Unit: %. Reference range: 70 - 130.): [94.0, NaN, NaN, NaN, NaN]
- HBsAg (Unit: IU/mL. Reference range: 0.0 - 0.01.): [0.01, NaN, NaN, NaN, NaN]
- mean corpuscular volume (Unit: fL. Reference range: 80 - 100.): [94.6, 93.6, NaN, 94.4, NaN]
- hematocrit (Unit: %. Reference range: 40 - 54 for men, 36 - 48 for women.): [33.2, 33.7, NaN, 35.7, NaN]
- White blood cell count (Unit: x 10^9/L. Reference range: 4.5 - 11.0.): [3.72, 5.31, NaN, 7.68, NaN]
- Tumor necrosis factorα (Unit: pg/mL. Reference range: less than 8.1.): [10.4, NaN, NaN, NaN, NaN]
- mean corpuscular hemoglobin concentration (Unit: g/L. Reference range: 320 - 360.): [328.0, 332.0, NaN, 353.0, NaN]
- fibrinogen (Unit: g/L. Reference range: 2 - 4.): [5.33, NaN, NaN, NaN, NaN]
- Interleukin 1β (Unit: pg/mL. Reference range: less than 6.5.): [5.0, NaN, NaN, NaN, NaN]
- Urea (Unit: mmol/L. Reference range: 1.8 - 7.1.): [2.9, 3.7, NaN, 4.22, NaN]
- lymphocyte count (Unit: x 10^9/L. Reference range: 1.0 - 4.8.): [0.71, 1.76, NaN, 2.1, NaN]
- PH value (Unit: /. Reference range: 7.35 - 7.45.): [NaN, NaN, NaN, NaN, NaN]
- Red blood cell count (Unit: x 10^12/L. Reference range: 4.5 - 5.5 for men, 4.0 - 5.0 for women.): [3.51, 3.6, NaN, 3.78, NaN]
- Eosinophil count (Unit: x 10^9/L. Reference range: 0.02 - 0.5.): [0.0, 0.01, NaN, 0.01, NaN]
- Corrected calcium (Unit: mmol/L. Reference range: 2.12 - 2.57.): [2.17, 2.25, NaN, 2.32, NaN]
- Serum potassium (Unit: mmol/L. Reference range: 3.5 - 5.0.): [3.34, 3.34, NaN, 3.9, NaN]
- glucose (Unit: mmol/L. Reference range: 3.9 - 5.6.): [9.02, 9.42, NaN, NaN, NaN]
- neutrophils count (Unit: x 10^9/L. Reference range: 2.0 - 8.0.): [2.79, 3.19, NaN, 5.09, NaN]
- Direct bilirubin (Unit: µmol/L. Reference range: 1.7 - 5.1.): [3.7, 4.0, NaN, 5.1, NaN]
- Mean platelet volume (Unit: fL. Reference range: 7.4 - 11.4.): [11.6, 10.7, NaN, 9.7, NaN]
- ferritin (Unit: ng/mL. Reference range: 24 - 336 for men, 11 - 307 for women.): [567.2, NaN, NaN, NaN, NaN]
- RBC distribution width SD (Unit: fL. Reference range: 40.0 - 55.0.): [44.4, 42.7, NaN, NaN, NaN]
- Thrombin time (Unit: seconds. Reference range: 12 - 19.): [16.7, NaN, NaN, NaN, NaN]
- (%)lymphocyte (Unit: %. Reference range: 20 - 40.): [19.1, 33.1, NaN, 27.3, NaN]
- HCV antibody quantification (Unit: IU/mL. Reference range: 0.04 - 0.08.): [0.06, NaN, NaN, NaN, NaN]
- DD dimer (Unit: mg/L. Reference range: 0 - 0.5.): [0.98, NaN, NaN, NaN, NaN]
- Total cholesterol (Unit: mmol/L. Reference range: less than 5.17.): [3.28, 3.49, NaN, 4.47, NaN]
- aspartate aminotransferase (Unit: U/L. Reference range: 8 - 33.): [30.0, 30.0, NaN, 20.0, NaN]
- Uric acid (Unit: µmol/L. Reference range: 240 - 510 for men, 160 - 430 for women.): [151.0, 135.0, NaN, 221.1, NaN]
- HCO3 (Unit: mmol/L. Reference range: 22 - 29.): [22.3, 24.6, NaN, 25.3, NaN]
- calcium (Unit: mmol/L. Reference range: 2.13 - 2.55.): [2.07, 2.13, NaN, 2.29, NaN]
- Aminoterminal brain natriuretic peptide precursor(NTproBNP) (Unit: pg/mL. Reference range: 0 - 125.): [NaN, NaN, NaN, NaN, NaN]
- Lactate dehydrogenase (Unit: U/L. Reference range: 140 - 280.): [328.0, 269.0, NaN, 226.0, NaN]
- platelet large cell ratio (Unit: %. Reference range: 15 - 35.): [37.8, 30.0, NaN, 21.4, NaN]
- Interleukin 6 (Unit: pg/mL. Reference range: 0 - 7.): [47.82, NaN, NaN, NaN, NaN]
- Fibrin degradation products (Unit: μg/mL. Reference range: 0 - 10.): [4.0, NaN, NaN, NaN, NaN]
- monocytes count (Unit: x 10^9/L. Reference range: 0.32 - 0.58.): [0.22, 0.34, NaN, 0.47, NaN]
- PLT distribution width (Unit: fL. Reference range: 9.2 - 16.7.): [13.9, 12.5, NaN, 10.1, NaN]
- globulin (Unit: g/L. Reference range: 23 - 35.): [33.4, 28.2, NaN, 29.8, NaN]
- γglutamyl transpeptidase (Unit: U/L. Reference range: 7 - 47 for men, 5 - 25 for women.): [21.0, 54.0, NaN, 53.0, NaN]
- International standard ratio (Unit: ratio. Reference range: 0.8 - 1.2.): [1.04, NaN, NaN, NaN, NaN]
- basophil count(#) (Unit: x 10^9/L. Reference range: 0.01 - 0.02.): [0.0, 0.01, NaN, 0.01, NaN]
- mean corpuscular hemoglobin (Unit: pg. Reference range: 27 - 31.): [31.1, 31.1, NaN, 33.3, NaN]
- Activation of partial thromboplastin time (Unit: seconds. Reference range: 22 - 35.): [34.8, NaN, NaN, NaN, NaN]
- High sensitivity Creactive protein (Unit: mg/L. Reference range: 3 - 10.): [42.3, 3.6, NaN, NaN, NaN]
- HIV antibody quantification (Unit: IU/mL. Reference range: 0.08 - 0.11.): [0.1, NaN, NaN, NaN, NaN]
- serum sodium (Unit: mmol/L. Reference range: 135 - 145.): [135.7, 140.4, NaN, 143.2, NaN]
- thrombocytocrit (Unit: %. Reference range: 0.22 - 0.24.): [0.2, 0.29, NaN, 0.23, NaN]
- ESR (Unit: mm/hr. Reference range: less than 15  for men, less than 20 for women.): [66.0, 29.0, NaN, NaN, NaN]
- glutamicpyruvic transaminase (Unit: U/L. Reference range: 0 - 35.): [19.0, 84.0, NaN, 67.0, NaN]
- eGFR (Unit: mL/min/1.73m². Reference range: more than 90.): [77.2, 90.0, NaN, 84.6, NaN]
- creatinine (Unit: µmol/L. Reference range: 61.9 - 114.9 for men, 53 - 97.2 for women.): [69.0, 58.0, NaN, 64.0, NaN]
\end{VerbatimWrap}
    \end{tcolorbox}
\end{center}
\clearpage

\captionof{table}{\textit{The base-setting prompt template for the mortality prediction task on the MIMIC-IV dataset with structured EHR data.}}
\label{tab:mimic4_base_full_ehr_prompt}
\begin{center} 
    \begin{tcolorbox}[colback=lightbluebg!30!white, colframe=blueframe, breakable, 
        title=The base-setting prompt template for the mortality prediction task on the MIMIC-IV dataset with structured EHR data
    ]
        \begin{VerbatimWrap}
You are an experienced critical care physician working in an Intensive Care Unit (ICU), skilled in interpreting complex longitudinal patient data and predicting clinical outcomes.
System Prompt: You are an experienced critical care physician working in an Intensive Care Unit (ICU), skilled in interpreting complex longitudinal patient data and predicting clinical outcomes.

User Prompt: I will provide you with longitudinal medical information for a patient. The data covers 3 visits that occurred at 2113-01-31, 2113-02-01, 2113-02-02.
Each clinical feature is presented as a list of values, corresponding to these visits. Missing values are represented as `NaN` for numerical values and "unknown" for categorical values. Note that units and reference ranges are provided alongside relevant features.

Patient Background:
- Sex: male
- Age: 50 years

Your Task:
Your primary task is to assess the provided medical data and analyze the health records from ICU visits to determine the likelihood of the patient not surviving their hospital stay.

Instructions & Output Format:
Please first perform a step-by-step analysis of the patient data, considering trends, abnormal values relative to reference ranges, and their clinical significance for survival. Then, provide a final assessment of the likelihood of not surviving the hospital stay.

Your final output must be a JSON object containing two keys:
1.  `"think"`: A string containing your detailed step-by-step clinical reasoning (under 500 words).
2.  `"answer"`: A floating-point number between 0 and 1 representing the predicted probability of mortality (higher value means higher likelihood of death).

Example Format: ```json { "think": "The patient presents with worsening X, stable Y, and improved Z. Factor A is a major risk indicator... Overall assessment suggests a high risk.", "answer": 0.85 }```

Handling Uncertainty:
In situations where the provided data is clearly insufficient or too ambiguous to make a reasonable prediction, respond with the exact phrase: `I do not know`.

Now, please analyze and predict for the following patient:

Clinical Features Over Time:
- Capillary refill rate: [0.0, 0.0, 0.0]
- Glascow coma scale eye opening: [Spontaneously, Spontaneously, To Speech]
- Glascow coma scale motor response: [Obeys Commands, Obeys Commands, Obeys Commands]
- Glascow coma scale total: [0.0, 0.0, 0.0]
- Glascow coma scale verbal response: [Oriented, Oriented, No Response]
- Diastolic blood pressure: [55.0, 74.0, 73.0]
- Fraction inspired oxygen: [80.0, 50.0, 70.0]
- Glucose: [119.0, 118.0, 127.0]
- Heart Rate: [86.0, 110.0, 118.0]
- Height: [157.0, NaN, NaN]
- Mean blood pressure: [67.0, 85.0, 102.0]
- Oxygen saturation: [96.0, 100.0, 100.0]
- Respiratory rate: [17.0, 26.0, 15.0]
- Systolic blood pressure: [105.0, 124.0, 156.0]
- Temperature: [37.89, 37.61, 37.17]
- Weight: [80.92, 80.92, 80.92]
- pH: [7.48, 7.47, 7.51]
\end{VerbatimWrap}
    \end{tcolorbox}
\end{center}
\clearpage

\captionof{table}{\textit{The optimized-setting prompt template for the mortality prediction task on the MIMIC-IV dataset with structured EHR data.}}
\label{tab:mimic4_optimized_full_ehr_prompt}
\begin{center} 
    \begin{tcolorbox}[colback=lightbluebg!30!white, colframe=blueframe, breakable, 
        title=The optimized-setting prompt template for the mortality prediction task on the MIMIC-IV dataset with structured EHR data
    ]
        \begin{VerbatimWrap}    
You are an experienced critical care physician working in an Intensive Care Unit (ICU), skilled in interpreting complex longitudinal patient data and predicting clinical outcomes.

I will provide you with longitudinal medical information for a patient. The data covers 3 visits that occurred at 2113-01-31, 2113-02-01, 2113-02-02.
Each clinical feature is presented as a list of values, corresponding to these visits. Missing values are represented as `NaN` for numerical values and "unknown" for categorical values. Note that units and reference ranges are provided alongside relevant features.

Patient Background:
- Sex: male
- Age: 50 years

Your Task:
Your primary task is to assess the provided medical data and analyze the health records from ICU visits to determine the likelihood of the patient not surviving their hospital stay.

Instructions & Output Format:
Please first perform a step-by-step analysis of the patient data, considering trends, abnormal values relative to reference ranges, and their clinical significance for survival. Then, provide a final assessment of the likelihood of not surviving the hospital stay.

Your final output must be a JSON object containing two keys:
1.  `"think"`: A string containing your detailed step-by-step clinical reasoning (under 500 words).
2.  `"answer"`: A floating-point number between 0 and 1 representing the predicted probability of mortality (higher value means higher likelihood of death).

Example Format: ```json { "think": "The patient presents with worsening X, stable Y, and improved Z. Factor A is a major risk indicator... Overall assessment suggests a high risk.", "answer": 0.85 }```

Handling Uncertainty:
In situations where the provided data is clearly insufficient or too ambiguous to make a reasonable prediction, respond with the exact phrase: `I do not know`.

Now, please analyze and predict for the following patient:

Clinical Features Over Time:
- Capillary refill rate (Unit: /. Reference range: /.): [0.0, 0.0, 0.0]
- Glascow coma scale eye opening (Unit: /. Reference range: /.): [Spontaneously, Spontaneously, To Speech]
- Glascow coma scale motor response (Unit: /. Reference range: /.): [Obeys Commands, Obeys Commands, Obeys Commands]
- Glascow coma scale total (Unit: /. Reference range: /.): [0.0, 0.0, 0.0]
- Glascow coma scale verbal response (Unit: /. Reference range: /.): [Oriented, Oriented, No Response]
- Diastolic blood pressure (Unit: mmHg. Reference range: less than 80.): [55.0, 74.0, 73.0]
- Fraction inspired oxygen (Unit: /. Reference range: more than 21.): [80.0, 50.0, 70.0]
- Glucose (Unit: mg/dL. Reference range: 70 - 100.): [119.0, 118.0, 127.0]
- Heart Rate (Unit: bpm. Reference range: 60 - 100.): [86.0, 110.0, 118.0]
- Height (Unit: cm. Reference range: /.): [157.0, NaN, NaN]
- Mean blood pressure (Unit: mmHg. Reference range: less than 100.): [67.0, 85.0, 102.0]
- Oxygen saturation (Unit: %. Reference range: 95 - 100.): [96.0, 100.0, 100.0]
- Respiratory rate (Unit: breaths per minute. Reference range: 15 - 18.): [17.0, 26.0, 15.0]
- Systolic blood pressure (Unit: mmHg. Reference range: less than 120.): [105.0, 124.0, 156.0]
- Temperature (Unit: degrees Celsius. Reference range: 36.1 - 37.2.): [37.89, 37.61, 37.17]
- Weight (Unit: kg. Reference range: /.): [80.92, 80.92, 80.92]
- pH (Unit: /. Reference range: 7.35 - 7.45.): [7.48, 7.47, 7.51]
\end{VerbatimWrap}
    \end{tcolorbox}
\end{center}
\clearpage

\captionof{table}{\textit{The optimized-setting with in-context learning prompt template for the mortality prediction task on the MIMIC-IV dataset with structured EHR data.}}
\label{tab:mimic4_optimized_icl_full_ehr_prompt}
\begin{center} 
    \begin{tcolorbox}[colback=lightbluebg!30!white, colframe=blueframe, breakable, 
        title=The optimized-setting with in-context learning prompt template for the mortality prediction task on the MIMIC-IV dataset with structured EHR data
    ]
        \begin{VerbatimWrap}
You are an experienced critical care physician working in an Intensive Care Unit (ICU), skilled in interpreting complex longitudinal patient data and predicting clinical outcomes.

I will provide you with longitudinal medical information for a patient. The data covers 3 visits that occurred at 2113-01-31, 2113-02-01, 2113-02-02.
Each clinical feature is presented as a list of values, corresponding to these visits. Missing values are represented as `NaN` for numerical values and "unknown" for categorical values. Note that units and reference ranges are provided alongside relevant features.

Patient Background:
- Sex: male
- Age: 50 years

Your Task:
Your primary task is to assess the provided medical data and analyze the health records from ICU visits to determine the likelihood of the patient not surviving their hospital stay.

Instructions & Output Format:
Please first perform a step-by-step analysis of the patient data, considering trends, abnormal values relative to reference ranges, and their clinical significance for survival. Then, provide a final assessment of the likelihood of not surviving the hospital stay.

Your final output must be a JSON object containing two keys:
1.  `"think"`: A string containing your detailed step-by-step clinical reasoning (under 500 words).
2.  `"answer"`: A floating-point number between 0 and 1 representing the predicted probability of mortality (higher value means higher likelihood of death).

Example Format: ```json { "think": "The patient presents with worsening X, stable Y, and improved Z. Factor A is a major risk indicator... Overall assessment suggests a high risk.", "answer": 0.85 }```

Handling Uncertainty:
In situations where the provided data is clearly insufficient or too ambiguous to make a reasonable prediction, respond with the exact phrase: `I do not know`.

Example:
Input information of a patient:
The patient is a female, aged 52 years.
The patient had 4 visits that occurred at 0, 1, 2, 3.
Details of the features for each visit are as follows:
- Capillary refill rate (Unit: /. Reference range: /.): ["unknown", "unknown", "unknown", "unknown"]
- Glascow coma scale eye opening (Unit: /. Reference range: /.): ["Spontaneously", "Spontaneously", "Spontaneously", "Spontaneously"]
- Glascow coma scale motor response (Unit: /. Reference range: /.): ["Obeys Commands", "Obeys Commands", "Obeys Commands", "Obeys Commands"]
...... (other features omitted for brevity)

Response:
```json { "think": "Patient is 52 years old. GCS components indicate full alertness and responsiveness (spontaneous eye opening, obeys commands) consistently across the recorded time points. While capillary refill is unknown, the neurological status appears stable and good. Assuming other vital signs and labs (not shown) are not critically deranged, the current data suggests a lower risk of mortality.", "answer": 0.3 } ```

Now, please analyze and predict for the following patient:

Clinical Features Over Time:
- Capillary refill rate (Unit: /. Reference range: /.): [0.0, 0.0, 0.0]
- Glascow coma scale eye opening (Unit: /. Reference range: /.): [Spontaneously, Spontaneously, To Speech]
- Glascow coma scale motor response (Unit: /. Reference range: /.): [Obeys Commands, Obeys Commands, Obeys Commands]
- Glascow coma scale total (Unit: /. Reference range: /.): [0.0, 0.0, 0.0]
- Glascow coma scale verbal response (Unit: /. Reference range: /.): [Oriented, Oriented, No Response]
- Diastolic blood pressure (Unit: mmHg. Reference range: less than 80.): [55.0, 74.0, 73.0]
- Fraction inspired oxygen (Unit: /. Reference range: more than 21.): [80.0, 50.0, 70.0]
- Glucose (Unit: mg/dL. Reference range: 70 - 100.): [119.0, 118.0, 127.0]
- Heart Rate (Unit: bpm. Reference range: 60 - 100.): [86.0, 110.0, 118.0]
- Height (Unit: cm. Reference range: /.): [157.0, NaN, NaN]
- Mean blood pressure (Unit: mmHg. Reference range: less than 100.): [67.0, 85.0, 102.0]
- Oxygen saturation (Unit: %. Reference range: 95 - 100.): [96.0, 100.0, 100.0]
- Respiratory rate (Unit: breaths per minute. Reference range: 15 - 18.): [17.0, 26.0, 15.0]
- Systolic blood pressure (Unit: mmHg. Reference range: less than 120.): [105.0, 124.0, 156.0]
- Temperature (Unit: degrees Celsius. Reference range: 36.1 - 37.2.): [37.89, 37.61, 37.17]
- Weight (Unit: kg. Reference range: /.): [80.92, 80.92, 80.92]
- pH (Unit: /. Reference range: 7.35 - 7.45.): [7.48, 7.47, 7.51]
\end{VerbatimWrap}
    \end{tcolorbox}
\end{center}
\clearpage

\begin{table}[H]
\footnotesize
\centering
\caption{\textit{Results of fairness analysis for the mortality prediction task on the TJH dataset.}}
\label{tab:fairness_tjh_mortality}
\resizebox{0.8\textwidth}{!}{
\begin{tabular}{c|c|c|cccc|cccc}
\toprule
\multicolumn{2}{c|}{\multirow{2}{*}{\textbf{Methods}}} & \multirow{2}{*}{\textbf{Setting}} & \multicolumn{4}{c|}{\textbf{Age}} & \multicolumn{4}{c}{\textbf{Gender}} \\
\multicolumn{2}{c|}{} & & \textbf{DI} & \textbf{SPD} & \textbf{AOD} & \textbf{EOD} & \textbf{DI} & \textbf{SPD} & \textbf{AOD} & \textbf{EOD} \\ \midrule

\multirow{57}{*}{Structured EHR} & \multirow{2}{*}{CatBoost} & 10 shot & 7.8000 & 0.5667 & 0.0362 & -0.0230 & 1.7619 & 0.2471 & -0.0069 & -0.0286 \\
 &  & full shot & 9.0000 & 0.5333 & -0.0099 & -0.0575 & 1.8738 & 0.2480 & -0.0179 & -0.0714 \\ \cmidrule{2-11}
 & \multirow{2}{*}{DT} & 10 shot & 2.0714 & 0.3214 & 0.0390 & 0.1006 & 1.6966 & 0.2542 & 0.0902 & 0.0476 \\
 &  & full shot & 7.2857 & 0.5238 & 0.0017 & -0.0920 & 1.8153 & 0.2424 & -0.0130 & -0.1143 \\ \cmidrule{2-11}
 & \multirow{2}{*}{RF} & 10 shot & 7.7143 & 0.5595 & 0.0267 & -0.0230 & 1.7374 & 0.2392 & -0.0158 & -0.0286 \\
 &  & full shot & 7.3714 & 0.5310 & -0.0036 & -0.0460 & 1.8420 & 0.2503 & -0.0112 & -0.0571 \\ \cmidrule{2-11}
 & \multirow{2}{*}{XGBoost} & 10 shot & 1.5000 & 0.2333 & 0.0793 & 0.0201 & 100000000.0000 & 1.0000 & 1.0000 & 1.0000 \\
 &  & full shot & 6.5143 & 0.4595 & -0.0648 & -0.1494 & 1.6780 & 0.1924 & -0.0661 & -0.1857 \\ \cmidrule{2-11}
 & \multirow{2}{*}{GRU} & 10 shot & 1.0376 & 0.0357 & 0.0079 & 0.0000 & 1.0257 & 0.0247 & 0.0104 & 0.0000 \\
 &  & full shot & 1.2205 & 0.1690 & 0.0401 & 0.0000 & 1.0893 & 0.0749 & 0.0150 & 0.0000 \\ \cmidrule{2-11}
 & \multirow{2}{*}{LSTM} & 10 shot & 5.5714 & 0.0762 & 0.0695 & 0.1379 & 3.5238 & 0.0682 & 0.0543 & 0.1095 \\
 &  & full shot & 6.4286 & 0.0905 & 0.0810 & 0.1609 & 1.7619 & 0.0412 & 0.0066 & 0.0143 \\ \cmidrule{2-11}
 & \multirow{2}{*}{Transformer} & 10 shot & 714285.7142 & 0.0071 & 0.0057 & 0.0115 & 793650.7936 & 0.0079 & 0.0071 & 0.0143 \\
 &  & full shot & 714285.7142 & 0.0071 & 0.0057 & 0.0115 & 793650.7936 & 0.0079 & 0.0071 & 0.0143 \\ \cmidrule{2-11}
 & \multirow{2}{*}{RNN} & 10 shot & 9.8571 & 0.4429 & 0.0363 & -0.0029 & 1.5270 & 0.1424 & -0.0724 & -0.1429 \\
 &  & full shot & 9.4286 & 0.4214 & 0.0191 & -0.0374 & 1.7965 & 0.1830 & -0.0010 & 0.0000 \\ \cmidrule{2-11}
 & \multirow{2}{*}{AdaCare} & 10 shot & 3.0857 & 0.3476 & -0.1287 & -0.2069 & 1.8206 & 0.2218 & 0.0205 & -0.0095 \\
 &  & full shot & 4.7143 & 0.4952 & -0.0115 & -0.0460 & 1.5812 & 0.2042 & -0.0400 & -0.0571 \\ \cmidrule{2-11}
 & \multirow{2}{*}{AICare} & 10 shot & 11.2286 & 0.8524 & 0.6457 & 0.4770 & 1.8331 & 0.3715 & 0.2263 & 0.1286 \\
 &  & full shot & 11.8571 & 0.5429 & 0.1278 & 0.1236 & 1.9381 & 0.2535 & 0.0329 & 0.0143 \\ \cmidrule{2-11}
 & \multirow{2}{*}{ConCare} & 10 shot & 4.6154 & 0.7833 & 0.6518 & 0.5000 & 1.5024 & 0.2919 & 0.1658 & 0.0333 \\
 &  & full shot & 1428571.4285 & 0.0143 & 0.0115 & 0.0230 & 1587301.5872 & 0.0159 & 0.0143 & 0.0286 \\ \cmidrule{2-11}
 & \multirow{2}{*}{GRASP} & 10 shot & 2.1122 & 0.2595 & -0.1669 & -0.2874 & 1.2875 & 0.1010 & -0.0682 & -0.1095 \\
 &  & full shot & 3.4714 & 0.4119 & -0.0659 & -0.1379 & 1.3921 & 0.1431 & -0.0636 & -0.0476 \\ \cmidrule{2-11}
 & \multirow{3}{*}{OpenBioLLM-8B} & base prompt & 1.0208 & 0.0190 & -0.0145 & -0.0805 & 1.0933 & 0.0819 & 0.0905 & 0.0857 \\
 &  & optimized prompt & 0.9828 & -0.0167 & -0.0371 & -0.0345 & 0.9926 & -0.0071 & -0.0199 & -0.0429 \\
 &  & opt.+ICL & 1.0255 & 0.0119 & -0.0123 & 0.0057 & 0.9206 & -0.0397 & -0.0644 & -0.0857 \\ \cmidrule{2-11}
 & \multirow{3}{*}{Qwen2.5-7B} & base prompt & 1.0962 & 0.0833 & 0.0054 & 0.0000 & 1.0105 & 0.0097 & -0.0238 & 0.0000 \\
 &  & optimized prompt & 1.0753 & 0.0690 & 0.0258 & 0.0000 & 1.0554 & 0.0517 & 0.0293 & 0.0000 \\
 &  & opt.+ICL & 1.4143 & 0.2762 & 0.1031 & 0.0000 & 1.1849 & 0.1424 & 0.0622 & 0.0000 \\ \cmidrule{2-11}
 & \multirow{3}{*}{Gemma-3-4B} & base prompt & 1.0097 & 0.0095 & 0.1193 & 0.2385 & 1.0057 & 0.0056 & 0.0167 & 0.0333 \\
 &  & optimized prompt & 0.9929 & -0.0071 & -0.0094 & 0.0000 & 0.9921 & -0.0079 & -0.0089 & 0.0000 \\
 &  & opt.+ICL & 1.0024 & 0.0024 & -0.0063 & -0.0115 & 0.9762 & -0.0238 & -0.0250 & -0.0143 \\ \cmidrule{2-11}
 & \multirow{3}{*}{DeepSeek-V3.1} & base prompt & 2.3673 & 0.4786 & 0.1329 & -0.0345 & 1.4242 & 0.2293 & 0.0672 & -0.0429 \\
 &  & optimized prompt & 1.9745 & 0.4548 & 0.1819 & 0.0000 & 1.2568 & 0.1735 & 0.0568 & 0.0000 \\
 &  & opt.+ICL & 1.4286 & 0.2786 & 0.0968 & -0.0115 & 1.1540 & 0.1186 & 0.0372 & -0.0143 \\ \cmidrule{2-11}
 & \multirow{3}{*}{GPT-4o} & base prompt & 3.9740 & 0.5452 & 0.0790 & 0.0000 & 1.7011 & 0.2748 & 0.0495 & 0.0000 \\
 &  & optimized prompt & 1.3902 & 0.2667 & 0.1036 & 0.0000 & 1.2375 & 0.1798 & 0.0984 & 0.0000 \\
 &  & opt.+ICL & 2.4156 & 0.5190 & 0.1883 & 0.0000 & 1.3615 & 0.2149 & 0.0687 & 0.0000 \\ \cmidrule{2-11}
 & \multirow{3}{*}{HuatuoGPT-o1-7B} & base prompt & 1.1739 & 0.1333 & 0.0077 & -0.0460 & 0.9911 & -0.0077 & -0.0592 & -0.0571 \\
 &  & optimized prompt & 1.4774 & 0.3024 & 0.1189 & -0.0230 & 1.1240 & 0.0972 & 0.0206 & -0.0286 \\
 &  & opt.+ICL & 1.2967 & 0.1929 & -0.0016 & -0.0575 & 1.1859 & 0.1306 & 0.0416 & -0.0095 \\ \cmidrule{2-11}
 & \multirow{3}{*}{DeepSeek-R1-7B} & base prompt & 1.0026 & 0.0024 & 0.0779 & 0.2155 & 0.9816 & -0.0174 & -0.0229 & 0.0048 \\
 &  & optimized prompt & 1.0208 & 0.0190 & -0.0292 & -0.0345 & 1.0431 & 0.0390 & 0.0215 & 0.0190 \\
 &  & opt.+ICL & 1.0485 & 0.0452 & 0.0148 & -0.0230 & 1.0554 & 0.0517 & 0.0329 & -0.0286 \\ \cmidrule{2-11}
 & \multirow{3}{*}{DeepSeek-R1} & base prompt & 2.2919 & 0.4952 & 0.1700 & 0.0000 & 1.3182 & 0.1935 & 0.0504 & 0.0000 \\
 &  & optimized prompt & 1.9286 & 0.4333 & 0.1536 & 0.0000 & 1.2585 & 0.1712 & 0.0483 & 0.0000 \\
 &  & opt.+ICL & 1.6387 & 0.3619 & 0.1378 & 0.0000 & 1.3390 & 0.2291 & 0.1193 & 0.0000 \\ \cmidrule{2-11}
 & \multirow{3}{*}{DeepSeek-V3.1-Think} & base prompt & 1.8000 & 0.4000 & 0.1395 & -0.0115 & 1.1746 & 0.1227 & 0.0129 & -0.0143 \\
 &  & optimized prompt & 1.4323 & 0.2738 & 0.0738 & 0.0000 & 1.2763 & 0.1941 & 0.0915 & 0.0000 \\
 &  & opt.+ICL & 1.5595 & 0.3357 & 0.1294 & 0.0000 & 1.1641 & 0.1242 & 0.0359 & 0.0000 \\ \cmidrule{2-11}
 & \multirow{3}{*}{o3-mini-high} & base prompt & 1.1345 & 0.1143 & 0.0332 & 0.0000 & 1.0678 & 0.0605 & 0.0219 & 0.0000 \\
 &  & optimized prompt & 1.6008 & 0.3405 & 0.1095 & 0.0000 & 1.3826 & 0.2482 & 0.1292 & 0.0000 \\
 &  & opt.+ICL & 1.1078 & 0.0952 & 0.0342 & 0.0000 & 1.0537 & 0.0493 & 0.0209 & 0.0000 \\ \cmidrule{2-11}
 & \multirow{3}{*}{GPT-5} & base prompt & 3.6593 & 0.5762 & 0.1461 & 0.0000 & 1.5546 & 0.2548 & 0.0559 & 0.0000 \\
 &  & optimized prompt & 2.1429 & 0.4762 & 0.1710 & 0.0000 & 1.5102 & 0.2896 & 0.1412 & 0.0000 \\
 &  & opt.+ICL & 1.8286 & 0.4143 & 0.1546 & 0.0000 & 1.3064 & 0.2029 & 0.0841 & 0.0000 \\ \bottomrule

\end{tabular}
}
\end{table}
\begin{table}[H]
\footnotesize
\centering
\caption{\textit{Results of fairness analysis for the mortality prediction task on the MIMIC-III dataset.}}
\label{tab:fairness_mimic3_mortality}
\resizebox{0.8\textwidth}{!}{
\begin{tabular}{c|c|c|cccc|cccc|cccc}
\toprule
\multicolumn{2}{c|}{\multirow{2}{*}{\textbf{Methods}}} & \multirow{2}{*}{\textbf{Setting}} & \multicolumn{4}{c|}{\textbf{Age}} & \multicolumn{4}{c|}{\textbf{Gender}} & \multicolumn{4}{c}{\textbf{Race}} \\
\multicolumn{2}{c|}{} & & \textbf{DI} & \textbf{SPD} & \textbf{AOD} & \textbf{EOD} & \textbf{DI} & \textbf{SPD} & \textbf{AOD} & \textbf{EOD} & \textbf{DI} & \textbf{SPD} & \textbf{AOD} & \textbf{EOD} \\ \midrule

\multirow{38}{*}{Unstructured Note} & \multirow{2}{*}{BERT} & freeze & 0.0000 & 0.0000 & 0.0000 & 0.0000 & 0.0000 & 0.0000 & 0.0000 & 0.0000 & 0.0000 & 0.0000 & 0.0000 & 0.0000 \\
 &  & finetune & 0.0000 & 0.0000 & 0.0000 & 0.0000 & 0.0000 & 0.0000 & 0.0000 & 0.0000 & 0.0000 & 0.0000 & 0.0000 & 0.0000 \\ \cmidrule{2-15}
 & \multirow{2}{*}{Clinical-Longformer} & freeze & 0.0000 & 0.0000 & 0.0000 & 0.0000 & 909090.8750 & 0.0091 & 0.0455 & 0.0909 & 0.0000 & 0.0000 & 0.0000 & 0.0000 \\
 &  & finetune & 0.0000 & 0.0000 & 0.0000 & 0.0000 & 3.2727 & 0.0253 & 0.1352 & 0.2727 & 0.0000 & 0.0000 & 0.0000 & 0.0000 \\ \cmidrule{2-15}
 & \multirow{2}{*}{BioBERT} & freeze & 0.0000 & 0.0000 & 0.0000 & 0.0000 & 0.0000 & 0.0000 & 0.0000 & 0.0000 & 0.0000 & 0.0000 & 0.0000 & 0.0000 \\
 &  & finetune & 0.0000 & 0.0000 & 0.0000 & 0.0000 & 2727272.7270 & 0.0273 & 0.0960 & 0.1818 & 0.0000 & 0.0000 & 0.0000 & 0.0000 \\
  \cmidrule{2-15}
 & \multirow{2}{*}{GatorTron} & freeze & 0.0000 & 0.0000 & 0.0000 & 0.0000 & 0.5455 & -0.0152 & 0.0230 & 0.0707 & 0.0000 & 0.0000 & 0.0000 & 0.0000 \\
 &  & finetune & 0.0000 & 0.0000 & 0.0000 & 0.0000 & 1.4727 & 0.0263 & 0.2076 & 0.4343 & 0.0000 & 0.0000 & 0.0000 & 0.0000 \\
  \cmidrule{2-15}
 & \multirow{2}{*}{ClinicalBERT} & freeze & 0.0000 & 0.0000 & 0.0000 & 0.0000 & 0.0000 & 0.0000 & 0.0000 & 0.0000 & 0.0000 & 0.0000 & 0.0000 & 0.0000 \\
 &  & finetune & 0.0000 & 0.0000 & 0.0000 & 0.0000 & 2.0455 & 0.0465 & 0.1874 & 0.3636 & 0.0000 & 0.0000 & 0.0000 & 0.0000 \\
 \cmidrule{2-15}
 & \multirow{2}{*}{GPT-2} & freeze & 0.0000 & 0.0000 & 0.0000 & 0.0000 & 0.0000 & 0.0000 & 0.0000 & 0.0000 & 0.0000 & 0.0000 & 0.0000 & 0.0000 \\
 &  & finetune & 0.0000 & 0.0000 & 0.0000 & 0.0000 & 3.2727 & 0.0253 & 0.1352 & 0.2727 & 0.0000 & 0.0000 & 0.0000 & 0.0000 \\
  \cmidrule{2-15}
 & \multirow{2}{*}{BioGPT} & freeze & 0.0000 & 0.0000 & 0.0000 & 0.0000 & 5454545.5000 & 0.0545 & 0.1515 & 0.2727 & 0.0000 & 0.0000 & 0.0000 & 0.0000 \\
 &  & finetune & 0.0000 & 0.0000 & 0.0000 & 0.0000 & 1.3091 & 0.0172 & 0.1218 & 0.2525 & 0.0000 & 0.0000 & 0.0000 & 0.0000 \\
 \cmidrule{2-15}
 & \multirow{2}{*}{meditron} & freeze & 0.0000 & 0.0000 & 0.0000 & 0.0000 & 0.0000 & -0.0111 & -0.0062 & 0.0000 & 0.0000 & 0.0000 & 0.0000 & 0.0000 \\
 &  & finetune & 0.0000 & 0.0000 & 0.0000 & 0.0000 & 0.9205 & -0.0071 & 0.1397 & 0.3232 & 0.0000 & 0.0000 & 0.0000 & 0.0000 \\
 \cmidrule{2-15}
 & \multirow{2}{*}{BioMistral} & freeze & 0.0000 & 0.0000 & 0.0000 & 0.0000 & 0.8182 & -0.0040 & 0.0382 & 0.0909 & 0.0000 & 0.0000 & 0.0000 & 0.0000 \\
 &  & finetune & 0.0000 & 0.0000 & 0.0000 & 0.0000 & 2727272.7270 & 0.0273 & 0.1364 & 0.2727 & 0.0000 & 0.0000 & 0.0000 & 0.0000 \\
 \cmidrule{2-15}
 & \multirow{3}{*}{OpenBioLLM-8B} & freeze & 0.0000 & 0.0000 & 0.0000 & 0.0000 & 0.8182 & -0.0020 & -0.0011 & 0.0000 & 0.0000 & 0.0000 & 0.0000 & 0.0000 \\
 &  & finetune & 0.0000 & 0.0000 & 0.0000 & 0.0000 & 2727272.7270 & 0.0273 & 0.1364 & 0.2727 & 0.0000 & 0.0000 & 0.0000 & 0.0000 \\
 &  & prompt & 0.0000 & 0.0000 & 0.0000 & 0.0000 & 0.9818 & -0.0182 & -0.0101 & 0.0000 & 0.0000 & 0.0000 & 0.0000 & 0.0000 \\ \cmidrule{2-15}
 & \multirow{3}{*}{Qwen2.5-7B} & freeze & 0.0000 & 0.0000 & 0.0000 & 0.0000 & 0.8182 & -0.0040 & 0.0382 & 0.0909 & 0.0000 & 0.0000 & 0.0000 & 0.0000 \\
 &  & finetune & 0.0000 & 0.0000 & 0.0000 & 0.0000 & 1.8409 & 0.0374 & 0.2138 & 0.4343 & 0.0000 & 0.0000 & 0.0000 & 0.0000 \\
 &  & prompt & 0.0000 & 0.0000 & 0.0000 & 0.0000 & 0.9511 & -0.0434 & -0.0241 & 0.0000 & 0.0000 & 0.0000 & 0.0000 & 0.0000 \\ \cmidrule{2-15}
 & \multirow{2}{*}{Gemma-3-4B} & freeze & 0.0000 & 0.0000 & 0.0000 & 0.0000 & 5454545.5000 & 0.0545 & 0.1515 & 0.2727 & 0.0000 & 0.0000 & 0.0000 & 0.0000 \\
 &  & prompt & 0.0000 & 0.0000 & 0.0000 & 0.0000 & 0.9875 & -0.0121 & -0.0067 & 0.0000 & 0.0000 & 0.0000 & 0.0000 & 0.0000 \\ \cmidrule{2-15}
 & \multirow{1}{*}{DeepSeek-V3.1} & prompt & 0.0000 & 0.0000 & 0.0000 & 0.0000 & 0.8416 & -0.0616 & -0.1150 & -0.1818 & 0.0000 & 0.0000 & 0.0000 & 0.0000 \\ \cmidrule{2-15}
 & \multirow{1}{*}{GPT-4o} & prompt & 0.0000 & 0.0000 & 0.0000 & 0.0000 & 0.8364 & -0.0818 & -0.0769 & -0.0707 & 0.0000 & 0.0000 & 0.0000 & 0.0000 \\ \cmidrule{2-15}
 & \multirow{3}{*}{HuatuoGPT-o1-7B} & freeze & 0.0000 & 0.0000 & 0.0000 & 0.0000 & 0.0000 & -0.0111 & -0.0062 & 0.0000 & 0.0000 & 0.0000 & 0.0000 & 0.0000 \\
 &  & finetune & 0.0000 & 0.0000 & 0.0000 & 0.0000 & 1.9091 & 0.0303 & 0.2189 & 0.4545 & 0.0000 & 0.0000 & 0.0000 & 0.0000 \\
 &  & prompt & 0.0000 & 0.0000 & 0.0000 & 0.0000 & 0.8935 & -0.0899 & -0.0903 & -0.0909 & 0.0000 & 0.0000 & 0.0000 & 0.0000 \\ \cmidrule{2-15}
 & \multirow{3}{*}{DeepSeek-R1-7B} & freeze & 0.0000 & 0.0000 & 0.0000 & 0.0000 & 0.0000 & 0.0000 & 0.0000 & 0.0000 & 0.0000 & 0.0000 & 0.0000 & 0.0000 \\
 &  & finetune & 0.0000 & 0.0000 & 0.0000 & 0.0000 & 0.9205 & -0.0071 & 0.0993 & 0.2323 & 0.0000 & 0.0000 & 0.0000 & 0.0000 \\
 &  & prompt & 0.0000 & 0.0000 & 0.0000 & 0.0000 & 0.9403 & -0.0444 & -0.0157 & 0.0202 & 0.0000 & 0.0000 & 0.0000 & 0.0000 \\ \cmidrule{2-15}
 & \multirow{1}{*}{DeepSeek-R1} & prompt & 0.0000 & 0.0000 & 0.0000 & 0.0000 & 0.9614 & -0.0172 & -0.0095 & 0.0000 & 0.0000 & 0.0000 & 0.0000 & 0.0000 \\ \cmidrule{2-15}
 & \multirow{1}{*}{DeepSeek-V3.1-Think} & prompt & 0.0000 & 0.0000 & 0.0000 & 0.0000 & 0.9273 & -0.0242 & -0.0045 & 0.0202 & 0.0000 & 0.0000 & 0.0000 & 0.0000 \\ \cmidrule{2-15}
 & \multirow{1}{*}{o3-mini-high} & prompt & 0.0000 & 0.0000 & 0.0000 & 0.0000 & 0.9474 & -0.0111 & -0.0780 & -0.1616 & 0.0000 & 0.0000 & 0.0000 & 0.0000 \\ \cmidrule{2-15}
 & \multirow{1}{*}{GPT-5} & prompt & 0.0000 & 0.0000 & 0.0000 & 0.0000 & 1.0041 & 0.0010 & 0.0185 & 0.0404 & 0.0000 & 0.0000 & 0.0000 & 0.0000 \\
\bottomrule

\end{tabular}
}
\end{table}
\begin{table}[H]
\footnotesize
\centering
\caption{\textit{Results of fairness analysis for the mortality prediction task on the MIMIC-IV dataset.}}
\label{tab:fairness_mimic4_mortality}
\resizebox{0.8\textwidth}{!}{
\begin{tabular}{c|c|c|cccc|cccc|cccc}
\toprule
\multicolumn{2}{c|}{\multirow{2}{*}{\textbf{Methods}}} & \multirow{2}{*}{\textbf{Setting}} & \multicolumn{4}{c|}{\textbf{Age}} & \multicolumn{4}{c|}{\textbf{Gender}} & \multicolumn{4}{c}{\textbf{Race}} \\
\multicolumn{2}{c|}{} & & 
\textbf{DI} & \textbf{SPD} & \textbf{AOD} & \textbf{EOD} & \textbf{DI} & \textbf{SPD} & \textbf{AOD} & \textbf{EOD} & \textbf{DI} & \textbf{SPD} & \textbf{AOD} & \textbf{EOD} \\ \midrule
\multirow{47}{*}{Unstructured Note} & \multirow{2}{*}{BERT} & freeze & 0.0000 & 0.0000 & 0.0000 & 0.0000 & 0.0000 & 0.0000 & 0.0000 & 0.0000 & 0.0000 & 0.0000 & 0.0000 & 0.0000 \\
 &  & finetune & 0.0000 & 0.0000 & 0.0000 & 0.0000 & 0.0000 & 0.0000 & 0.0000 & 0.0000 & 0.0000 & 0.0000 & 0.0000 & 0.0000 \\ \cmidrule{2-15}
 & \multirow{2}{*}{Clinical-Longformer} & freeze & 581395.3488 & 0.0058 & 0.0312 & 0.0625 & 869565.2173 & 0.0087 & 0.0500 & 0.1000 & 0.0000 & -0.0196 & -0.0556 & -0.1111 \\
 &  & finetune & 1.6279 & 0.0224 & 0.2379 & 0.5000 & 1.2935 & 0.0138 & 0.0388 & 0.0714 & 0.4107 & -0.0578 & -0.0796 & -0.1806 \\ \cmidrule{2-15}
 & \multirow{2}{*}{BioBERT} & freeze & 0.0000 & 0.0000 & 0.0000 & 0.0000 & 0.0000 & 0.0000 & 0.0000 & 0.0000 & 0.0000 & 0.0000 & 0.0000 & 0.0000 \\
 &  & finetune & 1162790.6976 & 0.0116 & 0.0345 & 0.0625 & 1739130.4346 & 0.0174 & 0.0548 & 0.1000 & 0.3423 & -0.0129 & -0.0520 & -0.1111 \\ \cmidrule{2-15}
 & \multirow{2}{*}{GatorTron} & freeze & 2906976.7440 & 0.0291 & 0.1002 & 0.1875 & 0.1848 & -0.0384 & -0.1057 & -0.1857 & 0.0856 & -0.0717 & -0.0724 & -0.0972 \\
 &  & finetune & 0.5698 & -0.0307 & -0.2998 & -0.5625 & 0.9239 & -0.0036 & -0.0810 & -0.1714 & 0.2738 & -0.0712 & -0.0867 & -0.1806 \\ \cmidrule{2-15}
 & \multirow{2}{*}{ClinicalBERT} & freeze & 0.0000 & 0.0000 & 0.0000 & 0.0000 & 0.0000 & 0.0000 & 0.0000 & 0.0000 & 0.0000 & 0.0000 & 0.0000 & 0.0000 \\
 &  & finetune & 1.1395 & 0.0050 & 0.1161 & 0.2500 & 1.2319 & 0.0082 & -0.0350 & -0.0857 & 0.3423 & -0.0516 & -0.1054 & -0.2083 \\ \cmidrule{2-15}
 & \multirow{2}{*}{GPT-2} & freeze & 0.0000 & 0.0000 & 0.0000 & 0.0000 & 0.0000 & 0.0000 & 0.0000 & 0.0000 & 0.0000 & 0.0000 & 0.0000 & 0.0000 \\
 &  & finetune & 1162790.6976 & 0.0116 & 0.0625 & 0.1250 & 1739130.4346 & 0.0174 & 0.1000 & 0.2000 & 0.0000 & -0.0392 & -0.1111 & -0.2222 \\ \cmidrule{2-15}
 & \multirow{2}{*}{BioGPT} & freeze & 2325581.3952 & 0.0233 & 0.0409 & 0.0625 & 0.2464 & -0.0266 & -0.0795 & -0.1429 & 1.0268 & 0.0005 & 0.0577 & 0.1250 \\
 &  & finetune & 1.7907 & 0.0282 & 0.2411 & 0.5000 & 2.2174 & 0.0430 & 0.0548 & 0.0714 & 0.4792 & -0.0511 & -0.0761 & -0.1806 \\ \cmidrule{2-15}
 & \multirow{2}{*}{meditron} & freeze & 581395.3488 & 0.0058 & 0.0312 & 0.0625 & 869565.2173 & 0.0087 & 0.0500 & 0.1000 & 0.0000 & -0.0196 & -0.0556 & -0.1111 \\
 &  & finetune & 0.0000 & 0.0000 & 0.0000 & 0.0000 & 0.0000 & 0.0000 & 0.0000 & 0.0000 & 0.0000 & 0.0000 & 0.0000 & 0.0000 \\ \cmidrule{2-15}
 & \multirow{2}{*}{BioMistral} & freeze & 0.4884 & -0.0183 & 0.0472 & 0.1250 & 0.2464 & -0.0266 & -0.0342 & -0.0429 & 0.3423 & -0.0258 & -0.0014 & 0.0139 \\
 &  & finetune & 0.6105 & -0.1113 & 0.1724 & 0.5000 & 1.1333 & 0.0235 & 0.1548 & 0.3143 & 0.6582 & -0.0871 & -0.1075 & -0.1806 \\ \cmidrule{2-15}
 & \multirow{3}{*}{OpenBioLLM-8B} & freeze & 0.6512 & -0.0125 & -0.0057 & 0.0000 & 2.9565 & 0.0230 & 0.0126 & 0.0000 & 0.5134 & -0.0191 & -0.0132 & 0.0000 \\
 &  & finetune & 1744186.0464 & 0.0174 & 0.0657 & 0.1250 & 1.4783 & 0.0056 & -0.0167 & -0.0429 & 0.6846 & -0.0062 & 0.0105 & 0.0139 \\
 &  & prompt & 0.9942 & -0.0058 & -0.0032 & 0.0000 & 0.9913 & -0.0087 & -0.0048 & 0.0000 & 0.9933 & -0.0067 & -0.0035 & -0.0000 \\ \cmidrule{2-15}
 & \multirow{3}{*}{Qwen2.5-7B} & freeze & 2325581.3952 & 0.0233 & 0.0128 & 0.0000 & 0.0000 & -0.0471 & -0.0256 & 0.0000 & 0.3423 & -0.0258 & -0.0167 & 0.0000 \\
 &  & finetune & 2906976.7440 & 0.0291 & 0.1282 & 0.2500 & 1.1087 & 0.0026 & -0.0381 & -0.0857 & 0.2282 & -0.0454 & -0.1006 & -0.2083 \\
 &  & prompt & 1.0921 & 0.0789 & 0.0388 & 0.0000 & 1.0841 & 0.0742 & 0.0403 & 0.0000 & 1.0050 & 0.0046 & 0.0086 & -0.0000 \\ \cmidrule{2-15}
 & \multirow{2}{*}{Gemma-3-4B} & freeze & 1162790.6976 & 0.0116 & 0.0625 & 0.1250 & 0.7391 & -0.0031 & -0.0214 & -0.0429 & 0.3423 & -0.0129 & 0.0069 & 0.0139 \\
 &  & prompt & 1.0069 & 0.0066 & 0.0025 & 0.0000 & 1.0311 & 0.0297 & 0.0161 & 0.0000 & 1.0705 & 0.0650 & 0.0405 & -0.0000 \\ \cmidrule{2-15}
 & \multirow{1}{*}{DeepSeek-V3.1} & prompt & 1.5465 & 0.0781 & 0.0150 & 0.0000 & 0.8130 & -0.0440 & -0.0262 & 0.0000 & 0.9646 & -0.0076 & 0.0578 & -0.0000 \\ \cmidrule{2-15}
 & \multirow{1}{*}{GPT-4o} & prompt & 1.3566 & 0.0764 & 0.0164 & 0.0000 & 0.9165 & -0.0246 & -0.0154 & 0.0000 & 0.7226 & -0.0979 & -0.0008 & -0.0000 \\ \cmidrule{2-15}
 & \multirow{3}{*}{HuatuoGPT-o1-7B} & freeze & 1162790.6976 & 0.0116 & 0.0064 & 0.0000 & 0.0000 & -0.0235 & -0.0128 & 0.0000 & 0.3423 & -0.0129 & -0.0084 & 0.0000 \\
 &  & finetune & 1162790.6976 & 0.0116 & 0.0625 & 0.1250 & 0.7391 & -0.0031 & -0.0214 & -0.0429 & 0.3423 & -0.0129 & 0.0069 & 0.0139 \\
 &  & prompt & 1.1567 & 0.1063 & 0.0481 & 0.0000 & 1.0676 & 0.0501 & 0.0267 & 0.0000 & 0.9128 & -0.0719 & -0.0241 & -0.0000 \\ \cmidrule{2-15}
 & \multirow{3}{*}{DeepSeek-R1-7B} & freeze & 581395.3488 & 0.0058 & 0.0032 & 0.0000 & 0.0000 & -0.0118 & -0.0064 & 0.0000 & 0.0000 & -0.0196 & -0.0119 & 0.0000 \\
 &  & finetune & 0.8140 & -0.0066 & 0.1097 & 0.2500 & 1.4783 & 0.0113 & 0.0769 & 0.1571 & 0.1711 & -0.0650 & -0.1125 & -0.2083 \\
 &  & prompt & 1.1318 & 0.0988 & 0.0463 & 0.0000 & 1.0761 & 0.0609 & 0.0328 & 0.0000 & 0.9568 & -0.0372 & -0.0089 & -0.0000 \\ \cmidrule{2-15}
 & \multirow{1}{*}{DeepSeek-R1} & prompt & 1.9942 & 0.1420 & 0.0502 & 0.0000 & 1.1263 & 0.0312 & 0.0150 & 0.0000 & 0.7248 & -0.0917 & 0.0041 & -0.0000 \\ \cmidrule{2-15}
 & \multirow{1}{*}{DeepSeek-V3.1-Think} & prompt & 1.6822 & 0.0731 & -0.0170 & -0.0625 & 0.9362 & -0.0113 & 0.0566 & 0.1429 & 0.8215 & -0.0350 & 0.0885 & 0.1111 \\ \cmidrule{2-15}
 & \multirow{1}{*}{o3-mini-high} & prompt & 1.6744 & 0.1686 & 0.0684 & 0.0000 & 0.9783 & -0.0087 & -0.0064 & 0.0000 & 1.0097 & 0.0038 & 0.0499 & -0.0000 \\ \cmidrule{2-15}
 & \multirow{1}{*}{GPT-5} & prompt & 1.3837 & 0.0548 & 0.0021 & 0.0000 & 0.9130 & -0.0174 & -0.0117 & 0.0000 & 0.7416 & -0.0608 & 0.0281 & -0.0000 \\ \midrule
 
\multirow{57}{*}{Structured EHR} 
 & \multirow{2}{*}{DT} & 10 shot & 0.7513 & -0.1154 & 0.1258 & 0.3889 & 0.8959 & -0.0404 & -0.0274 & -0.0114 & 1.0459 & 0.0162 & -0.0184 & -0.0667 \\
 &  & full shot & 2.2791 & 0.0457 & 0.0760 & 0.1111 & 1.4783 & 0.0281 & 0.0002 & -0.0341 & 0.9413 & -0.0046 & -0.0851 & -0.2000 \\ \cmidrule{2-15}
 & \multirow{2}{*}{RF} & 10 shot & 1.3204 & 0.1030 & 0.2666 & 0.4444 & 0.7391 & -0.1258 & -0.3239 & -0.5682 & 1.3087 & 0.1029 & -0.0082 & -0.1667 \\
 &  & full shot & 0.0000 & 0.0000 & 0.0000 & 0.0000 & 0.0000 & 0.0000 & 0.0000 & 0.0000 & 0.0000 & 0.0000 & 0.0000 & 0.0000 \\ \cmidrule{2-15}
 & \multirow{2}{*}{XGBoost} & 10 shot & 0.7764 & -0.1038 & 0.0587 & 0.2222 & 0.9407 & -0.0230 & -0.1397 & -0.2841 & 1.1678 & 0.0559 & 0.0282 & 0.0222 \\
 &  & full shot & 0.0000 & 0.0000 & 0.0000 & 0.0000 & 0.0000 & 0.0000 & 0.0000 & 0.0000 & 0.0000 & 0.0000 & 0.0000 & 0.0000 \\ \cmidrule{2-15}
 & \multirow{2}{*}{GRU} & 10 shot & 2.6047 & 0.0573 & -0.3254 & -0.7222 & 1.0559 & 0.0046 & 0.0532 & 0.1136 & 1.5973 & 0.0351 & 0.1457 & 0.2444 \\
 &  & full shot & 1.7907 & 0.0282 & -0.2680 & -0.5556 & 0.5280 & -0.0389 & -0.0422 & -0.0455 & 0.2445 & -0.1037 & -0.1384 & -0.2667 \\ \cmidrule{2-15}
 & \multirow{2}{*}{LSTM} & 10 shot & 1.0581 & 0.0042 & 0.0788 & 0.1667 & 1.1087 & 0.0077 & -0.0671 & -0.1591 & 2.2248 & 0.0480 & 0.0882 & 0.1222 \\
 &  & full shot & 0.9767 & -0.0008 & -0.3333 & -0.6667 & 0.9855 & -0.0005 & -0.0057 & -0.0114 & 0.2567 & -0.0583 & -0.0333 & -0.0667 \\ \cmidrule{2-15}
 & \multirow{2}{*}{Transformer} & 10 shot & 0.4070 & -0.0847 & 0.0075 & 0.1111 & 0.9855 & -0.0010 & -0.0159 & -0.0341 & 0.6161 & -0.0376 & -0.0147 & 0.0111 \\
 &  & full shot & 0.1628 & -0.0299 & -0.4722 & -0.9444 & 0.7391 & -0.0031 & -0.0170 & -0.0341 & 0.0000 & -0.0392 & -0.1000 & -0.2000 \\ \cmidrule{2-15}
 & \multirow{2}{*}{RNN} & 10 shot & 3.8527 & 0.3056 & 0.4693 & 0.6667 & 1.1469 & 0.0501 & 0.0325 & 0.0114 & 1.2408 & 0.0755 & 0.2767 & 0.4889 \\
 &  & full shot & 1.3023 & 0.0108 & -0.3023 & -0.6111 & 0.9239 & -0.0036 & 0.0333 & 0.0795 & 0.1711 & -0.0975 & -0.1853 & -0.3778 \\ \cmidrule{2-15}
 & \multirow{2}{*}{AdaCare} & 10 shot & 1.5058 & 0.2168 & 0.4571 & 0.7222 & 0.8283 & -0.1171 & -0.0190 & 0.1023 & 0.9734 & -0.0167 & -0.1078 & -0.2444 \\
 &  & full shot & 0.8953 & -0.0075 & 0.2195 & 0.5000 & 0.4620 & -0.0506 & -0.1454 & -0.2614 & 0.3993 & -0.0707 & -0.0293 & -0.0556 \\ \cmidrule{2-15}
 & \multirow{2}{*}{AICare} & 10 shot & 70930232.5540 & 0.7093 & 0.7395 & 0.7778 & 1.1395 & 0.0788 & -0.0634 & -0.2386 & 1.0497 & 0.0292 & -0.0278 & -0.1333 \\
 &  & full shot & 1.1395 & 0.0050 & -0.3056 & -0.6111 & 1.2319 & 0.0082 & 0.0398 & 0.0795 & 0.2054 & -0.0779 & -0.0833 & -0.1667 \\ \cmidrule{2-15}
 & \multirow{2}{*}{ConCare} & 10 shot & 1.1599 & 0.1370 & 0.0708 & 0.0000 & 1.0186 & 0.0179 & 0.0099 & 0.0000 & 1.0199 & 0.0191 & 0.0137 & -0.0000 \\
 &  & full shot & 2.7674 & 0.0631 & -0.1504 & -0.3333 & 0.5913 & -0.0481 & -0.0780 & -0.1136 & 0.6846 & -0.0371 & 0.1067 & 0.1778 \\ \cmidrule{2-15}
 & \multirow{2}{*}{GRASP} & 10 shot & 1.0039 & 0.0033 & -0.0284 & -0.0556 & 0.9788 & -0.0184 & 0.0457 & 0.1250 & 0.9957 & -0.0037 & -0.0416 & -0.1111 \\
 &  & full shot & 1.6279 & 0.0224 & -0.2713 & -0.5556 & 0.6159 & -0.0271 & -0.1324 & -0.2614 & 0.4107 & -0.0578 & -0.0206 & -0.0556 \\ \cmidrule{2-15}
 & \multirow{3}{*}{OpenBioLLM-8B} & base prompt & 1.0446 & 0.0382 & -0.0397 & -0.1250 & 0.9920 & -0.0072 & -0.0947 & -0.2000 & 1.0116 & 0.0103 & -0.1032 & -0.2500 \\
 &  & optimized prompt & 1.0446 & 0.0382 & -0.0117 & -0.0625 & 1.0151 & 0.0133 & -0.0383 & -0.1000 & 0.9540 & -0.0424 & -0.0752 & -0.1250 \\
 &  & opt.+ICL & 1.0024 & 0.0017 & -0.1218 & -0.2500 & 1.0244 & 0.0164 & -0.0626 & -0.1571 & 1.0268 & 0.0179 & 0.0051 & -0.0278 \\ \cmidrule{2-15}
 & \multirow{3}{*}{Qwen2.5-7B} & base prompt & 1.1247 & 0.0980 & 0.0190 & -0.0625 & 0.9531 & -0.0419 & 0.0418 & 0.1429 & 0.9524 & -0.0429 & 0.0287 & 0.1111 \\
 &  & optimized prompt & 1.0081 & 0.0075 & -0.0543 & -0.1250 & 1.0329 & 0.0302 & -0.0742 & -0.2000 & 1.0801 & 0.0707 & 0.0313 & -0.0139 \\
 &  & opt.+ICL & 0.9360 & -0.0640 & -0.0353 & 0.0000 & 0.9855 & -0.0138 & -0.0077 & 0.0000 & 0.9779 & -0.0212 & -0.0081 & -0.0000 \\ \cmidrule{2-15}
 & \multirow{3}{*}{Gemma-3-4B} & base prompt & 1.0370 & 0.0357 & 0.0185 & 0.0000 & 0.9913 & -0.0087 & -0.0048 & 0.0000 & 1.0200 & 0.0196 & 0.0119 & -0.0000 \\
 &  & optimized prompt & 0.9826 & -0.0174 & -0.0377 & -0.0625 & 0.9943 & -0.0056 & -0.0484 & -0.1000 & 1.0338 & 0.0325 & 0.0639 & 0.1111 \\
 &  & opt.+ICL & 1.0009 & 0.0008 & -0.0288 & -0.0625 & 0.9796 & -0.0199 & -0.0562 & -0.1000 & 1.0059 & 0.0057 & -0.0529 & -0.1250 \\ \cmidrule{2-15}
 & \multirow{3}{*}{DeepSeek-V3.1} & base prompt & 1.2047 & 0.1462 & 0.0432 & -0.0625 & 1.0870 & 0.0696 & 0.1026 & 0.1429 & 0.9078 & -0.0832 & -0.0952 & -0.1250 \\
 &  & optimized prompt & 1.4843 & 0.2940 & 0.1212 & -0.0625 & 1.0514 & 0.0430 & 0.0881 & 0.1429 & 1.0594 & 0.0490 & 0.0834 & 0.1111 \\
 &  & opt.+ICL & 1.1124 & 0.0963 & 0.0204 & -0.0625 & 1.0892 & 0.0798 & 0.1084 & 0.1429 & 1.0268 & 0.0247 & 0.0629 & 0.1111 \\ \cmidrule{2-15}
 & \multirow{3}{*}{GPT-4o} & base prompt & 1.8605 & 0.2151 & 0.0099 & -0.1875 & 1.0471 & 0.0199 & 0.2044 & 0.4286 & 0.8030 & -0.1004 & 0.0139 & 0.0972 \\
 &  & optimized prompt & 1.3721 & 0.1860 & 0.0581 & -0.0625 & 1.0348 & 0.0225 & 0.0764 & 0.1429 & 0.8162 & -0.1405 & -0.1147 & -0.1250 \\
 &  & opt.+ICL & 1.6279 & 0.2691 & 0.1296 & 0.0000 & 1.0348 & 0.0225 & 0.0114 & 0.0000 & 0.7872 & -0.1669 & -0.0712 & -0.0000 \\ \cmidrule{2-15}
 & \multirow{3}{*}{HuatuoGPT-o1-7B} & base prompt & 1.2081 & 0.1412 & 0.0393 & -0.0625 & 1.0803 & 0.0614 & 0.0980 & 0.1429 & 0.9024 & -0.0842 & -0.0926 & -0.1250 \\
 &  & optimized prompt & 1.1041 & 0.0855 & 0.0133 & -0.0625 & 1.0488 & 0.0425 & -0.0223 & -0.1000 & 0.9613 & -0.0357 & -0.0716 & -0.1250 \\
 &  & opt.+ICL & 0.9564 & -0.0374 & -0.0533 & -0.0625 & 1.0284 & 0.0230 & -0.0332 & -0.1000 & 0.8103 & -0.1823 & -0.1522 & -0.1250 \\ \cmidrule{2-15}
 & \multirow{3}{*}{DeepSeek-R1-7B} & base prompt & 1.0310 & 0.0299 & 0.0153 & 0.0000 & 0.9826 & -0.0174 & -0.0095 & 0.0000 & 0.9866 & -0.0134 & -0.0071 & -0.0000 \\
 &  & optimized prompt & 1.0456 & 0.0424 & 0.0210 & 0.0000 & 1.0220 & 0.0210 & 0.0114 & 0.0000 & 0.9789 & -0.0207 & -0.0094 & -0.0000 \\
 &  & opt.+ICL & 0.9709 & -0.0291 & -0.0441 & -0.0625 & 0.9767 & -0.0230 & -0.0579 & -0.1000 & 0.9664 & -0.0336 & -0.0767 & -0.1250 \\ \cmidrule{2-15}
 & \multirow{3}{*}{DeepSeek-R1} & base prompt & 1.3366 & 0.2284 & 0.0873 & -0.0625 & 1.0574 & 0.0486 & 0.0912 & 0.1429 & 0.9888 & -0.0099 & -0.0549 & -0.1250 \\
 &  & optimized prompt & 1.2558 & 0.1919 & 0.0976 & 0.0000 & 0.9516 & -0.0455 & -0.0251 & 0.0000 & 0.9627 & -0.0351 & -0.0139 & -0.0000 \\
 &  & opt.+ICL & 1.0809 & 0.0723 & 0.0363 & 0.0000 & 0.9825 & -0.0169 & -0.0093 & 0.0000 & 0.9919 & -0.0078 & -0.0010 & -0.0000 \\ \cmidrule{2-15}
 & \multirow{3}{*}{DeepSeek-V3.1-Think} & base prompt & 1.2424 & 0.1645 & 0.0521 & -0.0625 & 1.0435 & 0.0348 & 0.0835 & 0.1429 & 0.9051 & -0.0837 & -0.0939 & -0.1250 \\
 &  & optimized prompt & 1.4855 & 0.2774 & 0.1108 & -0.0625 & 0.9962 & -0.0031 & 0.0628 & 0.1429 & 0.9472 & -0.0445 & 0.0325 & 0.1111 \\
 &  & opt.+ICL & 1.2713 & 0.2035 & 0.1040 & 0.0000 & 0.9490 & -0.0486 & -0.0267 & 0.0000 & 0.9242 & -0.0744 & -0.0377 & -0.0000 \\ \cmidrule{2-15}
 & \multirow{3}{*}{o3-mini-high} & base prompt & 0.9079 & -0.0855 & -0.0495 & 0.0000 & 1.0410 & 0.0343 & 0.0183 & 0.0000 & 0.9301 & -0.0630 & -0.0256 & -0.0000 \\
 &  & optimized prompt & 1.6279 & 0.1570 & -0.1343 & -0.4375 & 0.9855 & -0.0056 & 0.1196 & 0.2714 & 0.9128 & -0.0359 & 0.0251 & 0.0694 \\
 &  & opt.+ICL & 1.1550 & 0.1163 & 0.0279 & -0.0625 & 1.0060 & 0.0051 & 0.0674 & 0.1429 & 0.8958 & -0.0961 & -0.0009 & 0.1111 \\ \cmidrule{2-15}
 & \multirow{3}{*}{GPT-5} & base prompt & 2.1163 & 0.2791 & 0.0732 & -0.1250 & 1.1182 & 0.0542 & 0.0480 & 0.0429 & 0.8144 & -0.1055 & -0.0371 & -0.0139 \\
 &  & optimized prompt & 1.9535 & 0.3065 & 0.0626 & -0.1875 & 1.3701 & 0.1785 & 0.1811 & 0.1857 & 0.9092 & -0.0570 & 0.0276 & 0.0972 \\
 &  & opt.+ICL & 1.4109 & 0.2201 & 0.0781 & -0.0625 & 0.9355 & -0.0486 & -0.0725 & -0.1000 & 0.8682 & -0.1059 & 0.0031 & 0.1111 \\
\bottomrule
\end{tabular}
}
\end{table}
\begin{table}[H]
\footnotesize
\centering
\caption{\textit{Results of fairness analysis for the 30-day readmission prediction task on the MIMIC-IV dataset.}}
\label{tab:fairness_mimic4_readmission}
\resizebox{0.8\textwidth}{!}{
\begin{tabular}{c|c|c|cccc|cccc|cccc}
\toprule
\multicolumn{2}{c|}{\multirow{2}{*}{\textbf{Methods}}} & \multirow{2}{*}{\textbf{Setting}} & \multicolumn{4}{c|}{\textbf{Age}} & \multicolumn{4}{c|}{\textbf{Gender}} & \multicolumn{4}{c}{\textbf{Race}} \\
\multicolumn{2}{c|}{} & & \textbf{DI} & \textbf{SPD} & \textbf{AOD} & \textbf{EOD} & \textbf{DI} & \textbf{SPD} & \textbf{AOD} & \textbf{EOD} & \textbf{DI} & \textbf{SPD} & \textbf{AOD} & \textbf{EOD} \\ \midrule
\multirow{47}{*}{Unstructured Note} & \multirow{2}{*}{BERT} & freeze & 9302325.5809 & 0.0930 & 0.1092 & 0.1458 & 1.6261 & 0.0368 & 0.0350 & 0.0229 & 0.7530 & -0.0242 & -0.0511 & -0.1117 \\
 &  & finetune & 1.8450 & 0.0905 & 0.0175 & -0.0667 & 1.9957 & 0.1171 & 0.1709 & 0.2643 & 1.0649 & 0.0114 & -0.0587 & -0.2179 \\ \cmidrule{2-15}
 & \multirow{2}{*}{Clinical-Longformer} & freeze & 1.1395 & 0.0050 & 0.0512 & 0.1458 & 1.2319 & 0.0082 & 0.0172 & 0.0229 & 1.0268 & 0.0011 & -0.0028 & -0.0147 \\
 &  & finetune & 1.9535 & 0.0681 & 0.0772 & 0.1333 & 1.0079 & 0.0010 & 0.0340 & 0.0771 & 1.4376 & 0.0429 & 0.0467 & 0.0476 \\
 \cmidrule{2-15}
 & \multirow{2}{*}{BioBERT} & freeze & 0.3256 & -0.0241 & -0.0073 & 0.0208 & 1.4783 & 0.0056 & -0.0085 & -0.0400 & 2013422.8187 & 0.0201 & 0.0219 & 0.0256 \\
 &  & finetune & 0.7733 & -0.0324 & -0.0839 & -0.1292 & 1.3859 & 0.0363 & 0.0823 & 0.1571 & 0.4450 & -0.1088 & -0.1652 & -0.2949 \\ \cmidrule{2-15}
 & \multirow{2}{*}{GatorTron} & freeze & 3488372.0928 & 0.0349 & 0.0497 & 0.0833 & 1.4783 & 0.0113 & 0.0310 & 0.0671 & 0.3423 & -0.0387 & -0.0243 & 0.0055 \\
 &  & finetune & 1.2209 & 0.0158 & 0.1000 & 0.2708 & 0.6570 & -0.0363 & -0.0143 & 0.0100 & 0.4890 & -0.0701 & -0.1244 & -0.2491 \\ \cmidrule{2-15}
 & \multirow{2}{*}{ClinicalBERT} & freeze & 5232558.1392 & 0.0523 & 0.0618 & 0.0833 & 5.9130 & 0.0578 & 0.0623 & 0.0671 & 0.6846 & -0.0186 & -0.0411 & -0.0916 \\
 &  & finetune & 6976744.1856 & 0.0698 & 0.0995 & 0.1667 & 1.0348 & 0.0020 & -0.0234 & -0.0929 & 1.7114 & 0.0279 & 0.0237 & 0.0110 \\ \cmidrule{2-15}
 & \multirow{2}{*}{GPT-2} & freeze & 4651162.7904 & 0.0465 & 0.0578 & 0.0833 & 2.2174 & 0.0286 & 0.0663 & 0.1429 & 1.0268 & 0.0011 & 0.0029 & 0.0055 \\
 &  & finetune & 11627906.9761 & 0.1163 & 0.1509 & 0.2292 & 1.3727 & 0.0307 & 0.0307 & 0.0143 & 0.7987 & -0.0237 & -0.0483 & -0.1062 \\ \cmidrule{2-15}
 & \multirow{2}{*}{BioGPT} & freeze & 13953488.3713 & 0.1395 & 0.1542 & 0.1875 & 1.7950 & 0.0655 & 0.0295 & -0.0571 & 1.3007 & 0.0295 & 0.0018 & -0.0604 \\
 &  & finetune & 2.8488 & 0.1321 & 0.1471 & 0.2167 & 1.3645 & 0.0558 & 0.0937 & 0.1443 & 1.4669 & 0.0641 & 0.0936 & 0.1502 \\ \cmidrule{2-15}
 & \multirow{2}{*}{meditron} & freeze & 5232558.1392 & 0.0523 & 0.0682 & 0.1042 & 1.4783 & 0.0169 & 0.0463 & 0.1029 & 1.1980 & 0.0078 & 0.0157 & 0.0311 \\
 &  & finetune & 0.0000 & 0.0000 & 0.0000 & 0.0000 & 0.0000 & 0.0000 & 0.0000 & 0.0000 & 0.0000 & 0.0000 & 0.0000 & 0.0000 \\ \cmidrule{2-15}
 & \multirow{2}{*}{BioMistral} & freeze & 2.2791 & 0.0457 & 0.0794 & 0.1458 & 0.6467 & -0.0332 & -0.0606 & -0.1286 & 0.6846 & -0.0309 & -0.0251 & -0.0147 \\
 &  & finetune & 2.9302 & 0.1379 & 0.0792 & 0.0375 & 1.2671 & 0.0440 & 0.0858 & 0.1400 & 0.6582 & -0.0871 & -0.0850 & -0.0897 \\ \cmidrule{2-15}
 & \multirow{3}{*}{OpenBioLLM-8B} & freeze & 5232558.1392 & 0.0523 & 0.0491 & 0.0417 & 1.4783 & 0.0169 & 0.0337 & 0.0714 & 0.6846 & -0.0186 & 0.0033 & 0.0513 \\
 &  & finetune & 1.6279 & 0.0897 & -0.0200 & -0.1417 & 1.2935 & 0.0552 & 0.0926 & 0.1357 & 0.4944 & -0.1784 & -0.2534 & -0.4267 \\
 &  & prompt & 0.9884 & -0.0116 & -0.0081 & 0.0000 & 1.0241 & 0.0235 & 0.0167 & 0.0000 & 1.0132 & 0.0129 & 0.0090 & 0.0000 \\ \cmidrule{2-15}
 & \multirow{3}{*}{Qwen2.5-7B} & freeze & 1.6279 & 0.0449 & 0.0228 & 0.0083 & 1.0676 & 0.0072 & 0.0140 & 0.0143 & 0.9128 & -0.0103 & -0.0088 & -0.0092 \\
 &  & finetune & 1.5194 & 0.0556 & 0.0125 & -0.0042 & 0.8975 & -0.0169 & 0.0689 & 0.2200 & 0.7188 & -0.0551 & -0.0792 & -0.1410 \\
 &  & prompt & 1.0206 & 0.0191 & 0.0008 & -0.0208 & 1.0518 & 0.0476 & 0.0470 & 0.0400 & 1.0641 & 0.0578 & 0.0625 & 0.0714 \\ \cmidrule{2-15}
 & \multirow{2}{*}{Gemma-3-4B} & freeze & 0.8682 & -0.0141 & 0.0248 & 0.0833 & 1.0163 & 0.0015 & -0.0026 & -0.0086 & 2.9094 & 0.0749 & 0.0529 & 0.0055 \\
 &  & prompt & 1.0394 & 0.0365 & 0.0193 & 0.0000 & 1.0129 & 0.0123 & 0.0103 & 0.0000 & 1.0268 & 0.0253 & 0.0178 & 0.0000 \\ \cmidrule{2-15}
 & \multirow{1}{*}{DeepSeek-V3.1} & prompt & 1.1802 & 0.1030 & 0.1533 & 0.2750 & 1.2935 & 0.1657 & 0.1760 & 0.1686 & 1.1152 & 0.0700 & 0.0164 & -0.1081 \\ \cmidrule{2-15}
 & \multirow{1}{*}{GPT-4o} & prompt & 1.2209 & 0.1262 & 0.0001 & -0.1667 & 1.1224 & 0.0777 & 0.0907 & 0.0929 & 1.0268 & 0.0179 & 0.0108 & -0.0110 \\ \cmidrule{2-15}
 & \multirow{3}{*}{HuatuoGPT-o1-7B} & freeze & 1.3837 & 0.0548 & 0.0613 & 0.0917 & 0.9130 & -0.0174 & -0.0058 & 0.0057 & 1.2836 & 0.0445 & 0.0305 & -0.0037 \\
 &  & finetune & 1.5194 & 0.0556 & 0.0189 & 0.0167 & 1.1703 & 0.0240 & 0.0616 & 0.1043 & 0.7188 & -0.0551 & -0.0710 & -0.1154 \\
 &  & prompt & 1.0803 & 0.0631 & 0.0128 & -0.0417 & 1.0870 & 0.0696 & 0.0550 & 0.0043 & 1.0602 & 0.0484 & 0.0186 & -0.0513 \\ \cmidrule{2-15}
 & \multirow{3}{*}{DeepSeek-R1-7B} & freeze & 0.3256 & -0.0241 & -0.0919 & -0.2000 & 0.3696 & -0.0148 & 0.0012 & 0.0357 & 0.1711 & -0.0325 & -0.0447 & -0.0714 \\
 &  & finetune & 2.7674 & 0.0631 & 0.0388 & 0.0292 & 1.4783 & 0.0338 & 0.0454 & 0.0500 & 0.6846 & -0.0371 & -0.0796 & -0.1777 \\
 &  & prompt & 1.0433 & 0.0341 & -0.0073 & -0.0417 & 0.9577 & -0.0353 & 0.0078 & 0.0800 & 0.9552 & -0.0378 & -0.0098 & 0.0458 \\ \cmidrule{2-15}
 & \multirow{1}{*}{DeepSeek-R1} & prompt & 0.8739 & -0.0855 & -0.1743 & -0.3417 & 1.0861 & 0.0496 & 0.0112 & -0.0557 & 1.3834 & 0.1804 & 0.1794 & 0.1813 \\ \cmidrule{2-15}
 & \multirow{1}{*}{DeepSeek-V3.1-Think} & prompt & 1.2820 & 0.1611 & 0.1744 & 0.2125 & 1.2045 & 0.1299 & 0.0888 & -0.0143 & 1.1766 & 0.1108 & 0.1104 & 0.1062 \\ \cmidrule{2-15}
 & \multirow{1}{*}{o3-mini-high} & prompt & 1.3598 & 0.2184 & 0.0990 & -0.0625 & 1.0689 & 0.0527 & 0.0572 & 0.0443 & 0.9234 & -0.0646 & -0.0667 & -0.0769 \\ \cmidrule{2-15}
 & \multirow{1}{*}{GPT-5} & prompt & 1.4109 & 0.2201 & 0.1787 & 0.1583 & 0.9620 & -0.0281 & 0.0049 & 0.0443 & 0.9638 & -0.0270 & -0.0401 & -0.0769 \\ \midrule
\multirow{57}{*}{Structured EHR} 
 & \multirow{2}{*}{CatBoost} & 10 shot & 1.1757 & 0.0565 & 0.2270 & 0.4651 & 0.8696 & -0.0522 & -0.1048 & -0.2121 & 1.1477 & 0.0492 & 0.0802 & 0.1373 \\
 &  & full shot & 0.0000 & 0.0000 & 0.0000 & 0.0000 & 0.0000 & 0.0000 & 0.0000 & 0.0000 & 0.0000 & 0.0000 & 0.0000 & 0.0000 \\ \cmidrule{2-15}
 & \multirow{2}{*}{DT} & 10 shot & 0.7513 & -0.1154 & 0.1121 & 0.4186 & 0.8959 & -0.0404 & -0.0673 & -0.1212 & 1.0459 & 0.0162 & 0.0380 & 0.0784 \\
 &  & full shot & 1.5465 & 0.0390 & 0.0802 & 0.1395 & 0.6719 & -0.0425 & -0.0910 & -0.1856 & 0.6846 & -0.0433 & -0.0451 & -0.0490 \\ \cmidrule{2-15}
 & \multirow{2}{*}{RF} & 10 shot & 1.3204 & 0.1030 & 0.2657 & 0.4884 & 0.7391 & -0.1258 & -0.1998 & -0.3447 & 1.3087 & 0.1029 & 0.0862 & 0.0539 \\
 &  & full shot & 0.0000 & 0.0000 & 0.0000 & 0.0000 & 0.0000 & 0.0000 & 0.0000 & 0.0000 & 0.0000 & 0.0000 & 0.0000 & 0.0000 \\ \cmidrule{2-15}
 & \multirow{2}{*}{XGBoost} & 10 shot & 0.7764 & -0.1038 & -0.0423 & 0.0388 & 0.9407 & -0.0230 & -0.0717 & -0.1629 & 1.1678 & 0.0559 & 0.0535 & 0.0490 \\
 &  & full shot & 0.0000 & 0.0000 & 0.0000 & 0.0000 & 0.0000 & 0.0000 & 0.0000 & 0.0000 & 0.0000 & 0.0000 & 0.0000 & 0.0000 \\ \cmidrule{2-15}
 & \multirow{2}{*}{GRU} & 10 shot & 2.6047 & 0.0573 & -0.0349 & -0.1240 & 1.0559 & 0.0046 & 0.0323 & 0.0682 & 1.5973 & 0.0351 & 0.0476 & 0.0686 \\
 &  & full shot & 2.7674 & 0.0631 & 0.0155 & 0.0155 & 0.5913 & -0.0481 & -0.0310 & -0.0303 & 0.4279 & -0.0897 & -0.1678 & -0.3186 \\ \cmidrule{2-15}
 & \multirow{2}{*}{LSTM} & 10 shot & 1.0581 & 0.0042 & -0.0820 & -0.1705 & 1.1087 & 0.0077 & -0.0261 & -0.1023 & 2.2248 & 0.0480 & 0.0745 & 0.1225 \\
 &  & full shot & 2.4419 & 0.0515 & -0.0155 & -0.0543 & 0.5749 & -0.0450 & -0.0445 & -0.0682 & 0.4401 & -0.0769 & -0.1512 & -0.2941 \\ \cmidrule{2-15}
 & \multirow{2}{*}{Transformer} & 10 shot & 0.4070 & -0.0847 & 0.0130 & 0.1628 & 0.9855 & -0.0010 & 0.0434 & 0.1174 & 0.6161 & -0.0376 & 0.0083 & 0.0931 \\
 &  & full shot & 0.8140 & -0.0066 & -0.1085 & -0.2171 & 3.6956 & 0.0317 & 0.0814 & 0.1629 & 0.6846 & -0.0124 & -0.0245 & -0.0490 \\ \cmidrule{2-15}
 & \multirow{2}{*}{RNN} & 10 shot & 3.8527 & 0.3056 & 0.3935 & 0.5349 & 1.1469 & 0.0501 & 0.0987 & 0.1742 & 1.2408 & 0.0755 & 0.0893 & 0.1127 \\
 &  & full shot & 1.9535 & 0.0341 & -0.0426 & -0.1008 & 0.8623 & -0.0097 & 0.0391 & 0.1098 & 0.2139 & -0.1233 & -0.2413 & -0.4657 \\ \cmidrule{2-15}
 & \multirow{2}{*}{AdaCare} & 10 shot & 1.5058 & 0.2168 & 0.4305 & 0.7209 & 0.8283 & -0.1171 & -0.1399 & -0.1894 & 0.9734 & -0.0167 & -0.0071 & 0.0098 \\
 &  & full shot & 1.0581 & 0.0042 & 0.0957 & 0.2558 & 0.8447 & -0.0128 & 0.0366 & 0.1098 & 0.2995 & -0.1099 & -0.2326 & -0.4657 \\ \cmidrule{2-15}
 & \multirow{2}{*}{AICare} & 10 shot & 70930232.5540 & 0.7093 & 0.7519 & 0.8372 & 1.1395 & 0.0788 & 0.0355 & -0.0682 & 1.0497 & 0.0292 & 0.0354 & 0.0441 \\
 &  & full shot & 2.7674 & 0.0631 & 0.0078 & -0.0078 & 0.9239 & -0.0072 & 0.0408 & 0.1023 & 0.4279 & -0.0897 & -0.2173 & -0.4608 \\ \cmidrule{2-15}
 & \multirow{2}{*}{ConCare} & 10 shot & 1.1599 & 0.1370 & 0.0761 & 0.0000 & 1.0186 & 0.0179 & 0.0128 & 0.0000 & 1.0199 & 0.0191 & 0.0126 & 0.0000 \\
 &  & full shot & 4.2326 & 0.1154 & 0.0736 & 0.0853 & 0.6863 & -0.0517 & -0.0194 & 0.0076 & 0.5819 & -0.0820 & -0.1711 & -0.3431 \\ \cmidrule{2-15}
 & \multirow{2}{*}{GRASP} & 10 shot & 1.0039 & 0.0033 & -0.0440 & -0.0930 & 0.9788 & -0.0184 & 0.0229 & 0.0947 & 0.9957 & -0.0037 & -0.0430 & -0.1176 \\
 &  & full shot & 1.1395 & 0.0100 & -0.0316 & -0.0310 & 0.7391 & -0.0246 & -0.0157 & -0.0265 & 0.4401 & -0.0769 & -0.1800 & -0.3775 \\ \cmidrule{2-15}
 & \multirow{3}{*}{OpenBioLLM-8B} & base prompt & 1.4884 & 0.1221 & 0.1106 & 0.0708 & 0.9537 & -0.0169 & -0.0016 & 0.0457 & 0.8728 & -0.0499 & -0.0445 & -0.0293 \\
 &  & optimized prompt & 1.2698 & 0.0482 & 0.0278 & -0.0125 & 0.8870 & -0.0266 & -0.0503 & -0.0971 & 0.8162 & -0.0468 & -0.0434 & -0.0348 \\
 &  & opt.+ICL & 1.3837 & 0.0548 & 0.0357 & 0.0083 & 0.8213 & -0.0379 & -0.0445 & -0.0614 & 1.2836 & 0.0445 & -0.0331 & -0.2033 \\ \cmidrule{2-15}
 & \multirow{3}{*}{Qwen2.5-7B} & base prompt & 1.0988 & 0.0847 & 0.0402 & -0.0208 & 0.9793 & -0.0194 & 0.0016 & 0.0400 & 0.9841 & -0.0150 & -0.0177 & -0.0256 \\
 &  & optimized prompt & 1.1085 & 0.0814 & 0.0624 & 0.0333 & 1.1258 & 0.0962 & 0.0371 & -0.0943 & 1.0268 & 0.0216 & -0.0270 & -0.1337 \\
 &  & opt.+ICL & 0.9593 & -0.0407 & -0.0282 & 0.0000 & 1.0220 & 0.0210 & 0.0161 & 0.0000 & 0.9789 & -0.0207 & -0.0138 & 0.0000 \\ \cmidrule{2-15}
 & \multirow{3}{*}{Gemma-3-4B} & base prompt & 1.0000 & 0.0000 & 0.0000 & 0.0000 & 1.0000 & 0.0000 & 0.0000 & 0.0000 & 1.0000 & 0.0000 & 0.0000 & 0.0000 \\
 &  & optimized prompt & 1.0000 & 0.0000 & 0.0000 & 0.0000 & 1.0000 & 0.0000 & 0.0000 & 0.0000 & 1.0000 & 0.0000 & 0.0000 & 0.0000 \\
 &  & opt.+ICL & 1.0000 & 0.0000 & 0.0000 & 0.0000 & 1.0000 & 0.0000 & 0.0000 & 0.0000 & 1.0000 & 0.0000 & 0.0000 & 0.0000 \\ \cmidrule{2-15}
 & \multirow{3}{*}{DeepSeek-V3.1} & base prompt & 1.0250 & 0.0241 & 0.0919 & 0.2000 & 0.9943 & -0.0056 & -0.0153 & -0.0357 & 0.9799 & -0.0201 & -0.0219 & -0.0256 \\
 &  & optimized prompt & 1.0000 & 0.0000 & 0.0000 & 0.0000 & 1.0000 & 0.0000 & 0.0000 & 0.0000 & 1.0000 & 0.0000 & 0.0000 & 0.0000 \\
 &  & opt.+ICL & 1.0000 & 0.0000 & 0.0000 & 0.0000 & 1.0000 & 0.0000 & 0.0000 & 0.0000 & 1.0000 & 0.0000 & 0.0000 & 0.0000 \\ \cmidrule{2-15}
 & \multirow{3}{*}{GPT-4o} & base prompt & 1.0353 & 0.0316 & -0.0064 & -0.0625 & 0.9399 & -0.0573 & -0.0482 & -0.0314 & 0.9698 & -0.0284 & -0.0129 & 0.0201 \\
 &  & optimized prompt & 0.9767 & -0.0233 & -0.0225 & -0.0208 & 1.0063 & 0.0061 & -0.0069 & -0.0357 & 0.9995 & -0.0005 & -0.0084 & -0.0256 \\
 &  & opt.+ICL & 0.9884 & -0.0116 & -0.0208 & -0.0417 & 1.0031 & 0.0031 & 0.0021 & 0.0043 & 1.0132 & 0.0129 & 0.0229 & 0.0458 \\ \cmidrule{2-15}
 & \multirow{3}{*}{HuatuoGPT-o1-7B} & base prompt & 1.0614 & 0.0548 & 0.1072 & 0.2000 & 0.9764 & -0.0225 & -0.0248 & -0.0357 & 0.9983 & -0.0016 & -0.0086 & -0.0256 \\
 &  & optimized prompt & 2.3953 & 0.3488 & 0.3301 & 0.3292 & 0.9907 & -0.0051 & -0.0333 & -0.1171 & 1.0024 & 0.0013 & -0.0347 & -0.1190 \\
 &  & opt.+ICL & 0.9587 & -0.0266 & -0.0854 & -0.1750 & 1.0580 & 0.0348 & 0.0496 & 0.0786 & 0.9060 & -0.0626 & -0.1347 & -0.2930 \\ \cmidrule{2-15}
 & \multirow{3}{*}{DeepSeek-R1-7B} & base prompt & 1.0519 & 0.0482 & 0.0992 & 0.1792 & 0.9679 & -0.0317 & -0.0208 & 0.0043 & 0.9597 & -0.0403 & -0.0438 & -0.0513 \\
 &  & optimized prompt & 1.0663 & 0.0473 & 0.0949 & 0.1708 & 0.9779 & -0.0169 & -0.0148 & -0.0100 & 0.9183 & -0.0657 & -0.0888 & -0.1392 \\
 &  & opt.+ICL & 0.9517 & -0.0449 & -0.0627 & -0.0833 & 0.9054 & -0.0890 & -0.0807 & -0.0671 & 0.9822 & -0.0161 & -0.0428 & -0.1026 \\ \cmidrule{2-15}
 & \multirow{3}{*}{DeepSeek-R1} & base prompt & 0.9826 & -0.0174 & -0.0185 & -0.0208 & 1.0152 & 0.0148 & -0.0012 & -0.0357 & 0.9799 & -0.0201 & -0.0219 & -0.0256 \\
 &  & optimized prompt & 1.0000 & 0.0000 & 0.0000 & 0.0000 & 1.0000 & 0.0000 & 0.0000 & 0.0000 & 1.0000 & 0.0000 & 0.0000 & 0.0000 \\
 &  & opt.+ICL & 1.0000 & 0.0000 & 0.0000 & 0.0000 & 1.0000 & 0.0000 & 0.0000 & 0.0000 & 1.0000 & 0.0000 & 0.0000 & 0.0000 \\ \cmidrule{2-15}
 & \multirow{3}{*}{DeepSeek-V3.1-Think} & base prompt & 1.0000 & 0.0000 & 0.0000 & 0.0000 & 1.0000 & 0.0000 & 0.0000 & 0.0000 & 1.0000 & 0.0000 & 0.0000 & 0.0000 \\
 &  & optimized prompt & 1.0000 & 0.0000 & 0.0000 & 0.0000 & 1.0000 & 0.0000 & 0.0000 & 0.0000 & 1.0000 & 0.0000 & 0.0000 & 0.0000 \\
 &  & opt.+ICL & 0.9942 & -0.0058 & -0.0104 & -0.0208 & 0.9913 & -0.0087 & -0.0179 & -0.0357 & 0.9933 & -0.0067 & -0.0128 & -0.0256 \\ \cmidrule{2-15}
 & \multirow{3}{*}{o3-mini-high} & base prompt & 1.0263 & 0.0216 & 0.0398 & 0.0750 & 0.8735 & -0.1146 & -0.1085 & -0.0986 & 0.9356 & -0.0568 & -0.0646 & -0.0824 \\
 &  & optimized prompt & 1.0242 & 0.0208 & -0.0296 & -0.1042 & 0.9628 & -0.0332 & -0.0537 & -0.1029 & 0.9056 & -0.0888 & -0.1008 & -0.1282 \\
 &  & opt.+ICL & 1.1192 & 0.1022 & 0.1242 & 0.1583 & 0.9645 & -0.0343 & -0.0335 & -0.0314 & 1.0341 & 0.0315 & 0.0279 & 0.0201 \\ \cmidrule{2-15}
 & \multirow{3}{*}{GPT-5} & base prompt & 1.0000 & 0.0000 & 0.0000 & 0.0000 & 1.0000 & 0.0000 & 0.0000 & 0.0000 & 1.0000 & 0.0000 & 0.0000 & 0.0000 \\
 &  & optimized prompt & 1.0000 & 0.0000 & 0.0000 & 0.0000 & 1.0000 & 0.0000 & 0.0000 & 0.0000 & 1.0000 & 0.0000 & 0.0000 & 0.0000 \\
 &  & opt.+ICL & 1.0000 & 0.0000 & 0.0000 & 0.0000 & 1.0000 & 0.0000 & 0.0000 & 0.0000 & 1.0000 & 0.0000 & 0.0000 & 0.0000 \\
\bottomrule
\end{tabular}
}
\end{table}
\begin{table}[H]
\centering
\caption{\textit{Results of fairness analysis for mortality and readmission prediction tasks on the MIMIC-IV dataset utilizing multimodal EHR data.}}
\label{tab:fairness_multimodal_mimic4}
\resizebox{\textwidth}{!}{
\begin{tabular}{c|c|c|cccc|cccc|cccc}
\toprule
\multirow{2}{*}{\textbf{Tasks}} & \multirow{2}{*}{\textbf{Methods}} & \multirow{2}{*}{\textbf{Setting}} & \multicolumn{4}{c|}{\textbf{Age}} & \multicolumn{4}{c|}{\textbf{Gender}} & \multicolumn{4}{c}{\textbf{Race}} \\
 & & & 
\textbf{DI} & \textbf{SPD} & \textbf{AOD} & \textbf{EOD} & \textbf{DI} & \textbf{SPD} & \textbf{AOD} & \textbf{EOD} & \textbf{DI} & \textbf{SPD} & \textbf{AOD} & \textbf{EOD} \\ \midrule
\multirow{22}{*}{Mortality} & \multirow{1}{*}{OpenBioLLM-8B} & optimized prompt & 1.0018 & 0.0017 & -0.0575 & -0.1250 & 1.0698 & 0.0624 & 0.0537 & 0.0429 & 1.0417 & 0.0376 & 0.0123 & -0.0139 \\ \cmidrule{2-15}
 & \multirow{1}{*}{Qwen2.5-7B} & optimized prompt & 1.0874 & 0.0781 & 0.0395 & 0.0000 & 1.0572 & 0.0532 & 0.0289 & 0.0000 & 1.0864 & 0.0779 & 0.0489 & -0.0000 \\ \cmidrule{2-15}
 & \multirow{1}{*}{Gemma-3-4B} & optimized prompt & 1.0189 & 0.0183 & 0.4904 & 1.0000 & 0.9652 & -0.0348 & -0.0643 & -0.1000 & 0.9995 & -0.0005 & 0.0449 & 0.1111 \\ \cmidrule{2-15}
 & \multirow{1}{*}{DeepSeek-V3.1} & optimized prompt & 1.1477 & 0.1055 & 0.0488 & 0.0000 & 1.1202 & 0.0905 & 0.0489 & 0.0000 & 1.0018 & 0.0014 & 0.0162 & -0.0000 \\ \cmidrule{2-15}
 & \multirow{1}{*}{GPT-4o} & optimized prompt & 1.3735 & 0.2135 & 0.1036 & 0.0000 & 1.1525 & 0.1059 & 0.0571 & 0.0000 & 1.0934 & 0.0659 & 0.0580 & -0.0000 \\ \cmidrule{2-15}
 & \multirow{1}{*}{HuatuoGPT-o1-7B} & optimized prompt & 1.0069 & 0.0066 & 0.0025 & 0.0000 & 0.9885 & -0.0113 & -0.0062 & 0.0000 & 1.0129 & 0.0124 & 0.0096 & -0.0000 \\ \cmidrule{2-15}
 & \multirow{1}{*}{DeepSeek-R1-7B} & optimized prompt & 1.0206 & 0.0191 & 0.0082 & 0.0000 & 0.9855 & -0.0138 & -0.0077 & 0.0000 & 1.0341 & 0.0315 & 0.0228 & -0.0000 \\ \cmidrule{2-15}
 & \multirow{1}{*}{DeepSeek-R1} & optimized prompt & 1.3114 & 0.2002 & 0.0986 & 0.0000 & 1.0326 & 0.0261 & 0.0137 & 0.0000 & 0.9861 & -0.0114 & 0.0079 & -0.0000 \\ \cmidrule{2-15}
 & \multirow{1}{*}{DeepSeek-V3.1-Think} & optimized prompt & 1.3939 & 0.2251 & 0.1100 & 0.0000 & 1.1148 & 0.0824 & 0.0443 & 0.0000 & 1.0359 & 0.0267 & 0.0342 & -0.0000 \\ \cmidrule{2-15}
 & \multirow{1}{*}{o3-mini-high} & optimized prompt & 2.0698 & 0.2674 & 0.0948 & -0.0625 & 1.0348 & 0.0164 & 0.0725 & 0.1429 & 0.7530 & -0.1453 & -0.1033 & -0.1250 \\ \cmidrule{2-15}
 & \multirow{1}{*}{GPT-5} & optimized prompt & 2.1705 & 0.3762 & 0.1852 & 0.0000 & 0.9945 & -0.0036 & -0.0029 & 0.0000 & 0.9193 & -0.0554 & -0.0046 & -0.0000 \\ \cmidrule{2-15}
 & \multirow{5}{*}{Tuning} & Add & 1.4651 & 0.0166 & -0.2745 & -0.5556 & 0.4928 & -0.0358 & -0.1372 & -0.2614 & 0.3423 & -0.0645 & -0.0242 & -0.0556 \\ \cmidrule{3-15}
 &  & Concat & 1.4651 & 0.0166 & -0.2745 & -0.5556 & 0.4928 & -0.0358 & -0.1372 & -0.2614 & 0.3423 & -0.0645 & -0.0242 & -0.0556 \\ \cmidrule{3-15}
 &  & Self-Attention & 2.1163 & 0.0399 & -0.2125 & -0.4444 & 0.4106 & -0.0624 & -0.1672 & -0.2955 & 0.3423 & -0.0903 & -0.0273 & -0.0444 \\ \cmidrule{3-15}
 &  & Cross-Attention & 2.2791 & 0.0457 & -0.1847 & -0.3889 & 0.4928 & -0.0537 & -0.1218 & -0.2045 & 0.3912 & -0.0836 & 0.0283 & 0.0667 \\

\midrule
\multirow{22}{*}{Readmission} & \multirow{1}{*}{OpenBioLLM-8B} & optimized prompt & 0.6033 & -0.2409 & -0.3919 & -0.6250 & 1.1087 & 0.0409 & 0.0500 & 0.0643 & 1.2685 & 0.0895 & 0.0336 & -0.0897 \\ \cmidrule{2-15}
 & \multirow{1}{*}{Qwen2.5-7B} & optimized prompt & 0.8921 & -0.0963 & -0.1014 & -0.0833 & 1.0751 & 0.0583 & 0.0245 & -0.0671 & 1.0101 & 0.0082 & 0.0052 & -0.0055 \\ \cmidrule{2-15}
 & \multirow{1}{*}{Gemma-3-4B} & optimized prompt & 1.0000 & 0.0000 & 0.0000 & 0.0000 & 1.0000 & 0.0000 & 0.0000 & 0.0000 & 1.0000 & 0.0000 & 0.0000 & 0.0000 \\ \cmidrule{2-15}
 & \multirow{1}{*}{DeepSeek-V3.1} & optimized prompt & 1.0250 & 0.0241 & 0.0792 & 0.1583 & 0.9943 & -0.0056 & -0.0157 & -0.0314 & 1.0063 & 0.0062 & 0.0101 & 0.0201 \\ \cmidrule{2-15}
 & \multirow{1}{*}{GPT-4o} & optimized prompt & 1.0581 & 0.0540 & 0.0250 & -0.0208 & 1.0186 & 0.0179 & 0.0014 & -0.0357 & 1.0199 & 0.0191 & 0.0051 & -0.0256 \\ \cmidrule{2-15}
 & \multirow{1}{*}{HuatuoGPT-o1-7B} & optimized prompt & 0.9858 & -0.0091 & -0.1388 & -0.3333 & 0.9992 & -0.0005 & -0.0341 & -0.1171 & 1.1067 & 0.0628 & 0.0369 & -0.0220 \\ \cmidrule{2-15}
 & \multirow{1}{*}{DeepSeek-R1-7B} & optimized prompt & 1.1628 & 0.0814 & -0.0418 & -0.2375 & 0.9805 & -0.0113 & -0.0636 & -0.1800 & 1.3542 & 0.1598 & 0.1765 & 0.2125 \\ \cmidrule{2-15}
 & \multirow{1}{*}{DeepSeek-R1} & optimized prompt & 0.9477 & -0.0523 & -0.0554 & -0.0625 & 1.0038 & 0.0036 & -0.0080 & -0.0314 & 0.9919 & -0.0078 & -0.0296 & -0.0769 \\ \cmidrule{2-15}
 & \multirow{1}{*}{DeepSeek-V3.1-Think} & optimized prompt & 1.0189 & 0.0183 & 0.0033 & -0.0208 & 1.0276 & 0.0266 & 0.0309 & 0.0400 & 1.0554 & 0.0521 & 0.0582 & 0.0714 \\ \cmidrule{2-15}
 & \multirow{1}{*}{o3-mini-high} & optimized prompt & 1.2209 & 0.1420 & 0.1082 & 0.0750 & 1.1148 & 0.0824 & 0.0782 & 0.0529 & 1.0005 & 0.0004 & -0.0549 & -0.1795 \\ \cmidrule{2-15}
 & \multirow{1}{*}{GPT-5} & optimized prompt & 1.0310 & 0.0299 & 0.0177 & 0.0000 & 1.0241 & 0.0235 & 0.0167 & 0.0000 & 1.0408 & 0.0392 & 0.0270 & 0.0000 \\ \cmidrule{2-15}
 & \multirow{5}{*}{Tuning} & Add & 1.4651 & 0.0166 & -0.0620 & -0.1240 & 0.7391 & -0.0153 & -0.0095 & -0.0189 & 0.1467 & -0.1171 & -0.2475 & -0.4951 \\ \cmidrule{3-15}
 &  & Concat & 2.7674 & 0.0631 & 0.0078 & -0.0078 & 0.9239 & -0.0072 & 0.0408 & 0.1023 & 0.5379 & -0.0634 & -0.1610 & -0.3480 \\ \cmidrule{3-15}
 &  & Self-Attention & 1.4651 & 0.0166 & -0.0620 & -0.1240 & 0.7391 & -0.0153 & -0.0095 & -0.0189 & 0.1467 & -0.1171 & -0.2475 & -0.4951 \\ \cmidrule{3-15}
 &  & Cross-Attention & 1.4651 & 0.0166 & -0.0620 & -0.1240 & 0.7391 & -0.0153 & -0.0095 & -0.0189 & 0.1467 & -0.1171 & -0.2475 & -0.4951 \\

\bottomrule
\end{tabular}
}
\end{table}

\captionof{table}{\textit{The prompt template for the mortality prediction task on the MIMIC-IV dataset with multimodal EHR data.}}
\label{tab:mimic4_multimodal_mortality_prompt}
\begin{center}
\begin{tcolorbox}[colback=lightbluebg!30!white, colframe=blueframe, breakable,
title=The prompt template for the mortality prediction task on the MIMIC-IV dataset with multimodal EHR data
]
\begin{VerbatimWrap}    
You are an experienced critical care physician working in an Intensive Care Unit (ICU), skilled in interpreting complex longitudinal patient data, including time-series vitals and labs, as well as unstructured clinical notes to predict clinical outcomes.

I will provide you with longitudinal medical information for a patient. The data covers 3 visits that occurred at 2113-01-31, 2113-02-01, 2113-02-02.
Each clinical feature is presented as a list of values, corresponding to these visits. Missing values are represented as `NaN` for numerical values and "unknown" for categorical values. Note that units and reference ranges are provided alongside relevant features.

Patient Background:
- Sex: male
- Age: 50 years

Your Task:
Your primary task is to assess the provided medical data, including both the time-series clinical features and the unstructured clinical note, to determine the likelihood of the patient not surviving their hospital stay.

Instructions & Output Format:
Please first perform a step-by-step analysis. In your reasoning, you must integrate insights from both the structured time-series data (considering trends and abnormal values) and the unstructured clinical note (considering patient history, context, and overall status). Then, provide a final assessment of the likelihood of not surviving the hospital stay.

Your final output must be a JSON object containing two keys:
1.  `"think"`: A string containing your detailed step-by-step clinical reasoning (under 500 words).
2.  `"answer"`: A floating-point number between 0 and 1 representing the predicted probability of mortality (higher value means higher likelihood of death).

Example Format: ```json { "think": "The patient presents with worsening X, stable Y, and improved Z. Factor A is a major risk indicator... Overall assessment suggests a high risk.", "answer": 0.85 }```

Handling Uncertainty:
In situations where the provided data is clearly insufficient or too ambiguous to make a reasonable prediction, respond with the exact phrase: `I do not know`

Now, please analyze and predict for the following patient:

Clinical Note: sex f service neurosurgery allergies prednisone latex lactose mirapex doxycycline attending . chief complaint intraventricular hemorrhage major surgical or invasive procedure r evd placed angio dural av fistula angio for embolization post meningeal branch replacement of r evd embolization of dural av fistula peg insertion history of present illness this is a year old female found down by her roommate in the bathroom on the ground with emesis. roommate stated that the patient may have taken some percocet. ems called and patient taken to . upon arrival patient was obtunded, received narcan with no improvement and was intubated for airway protection and underwent a head ct which revealed a hemorrhage......

Clinical Features Over Time:
- Capillary refill rate (Unit: /. Reference range: /.): [0.0, 0.0, 0.0]
- Glascow coma scale eye opening (Unit: /. Reference range: /.): [Spontaneously, Spontaneously, To Speech]
- Glascow coma scale motor response (Unit: /. Reference range: /.): [Obeys Commands, Obeys Commands, Obeys Commands]
- Glascow coma scale total (Unit: /. Reference range: /.): [0.0, 0.0, 0.0]
- Glascow coma scale verbal response (Unit: /. Reference range: /.): [Oriented, Oriented, No Response]
- Diastolic blood pressure (Unit: mmHg. Reference range: less than 80.): [55.0, 74.0, 73.0]
- Fraction inspired oxygen (Unit: /. Reference range: more than 21.): [80.0, 50.0, 70.0]
- Glucose (Unit: mg/dL. Reference range: 70 - 100.): [119.0, 118.0, 127.0]
- Heart Rate (Unit: bpm. Reference range: 60 - 100.): [86.0, 110.0, 118.0]
- Height (Unit: cm. Reference range: /.): [157.0, NaN, NaN]
- Mean blood pressure (Unit: mmHg. Reference range: less than 100.): [67.0, 85.0, 102.0]
- Oxygen saturation (Unit: %. Reference range: 95 - 100.): [96.0, 100.0, 100.0]
- Respiratory rate (Unit: breaths per minute. Reference range: 15 - 18.): [17.0, 26.0, 15.0]
- Systolic blood pressure (Unit: mmHg. Reference range: less than 120.): [105.0, 124.0, 156.0]
- Temperature (Unit: degrees Celsius. Reference range: 36.1 - 37.2.): [37.89, 37.61, 37.17]
- Weight (Unit: kg. Reference range: /.): [80.92, 80.92, 80.92]
- pH (Unit: /. Reference range: 7.35 - 7.45.): [7.48, 7.47, 7.51]
\end{VerbatimWrap}
\end{tcolorbox}
\end{center}
\clearpage

\captionof{table}{\textit{The prompt template for the readmission prediction task on the MIMIC-IV dataset with multimodal EHR data.}}
\label{tab:mimic4_multimodal_readmission_prompt}
\begin{center}
\begin{tcolorbox}[colback=lightbluebg!30!white, colframe=blueframe, breakable,
title=The prompt template for the readmission prediction task on the MIMIC-IV dataset with multimodal EHR data
]
\begin{VerbatimWrap}    
You are an experienced critical care physician working in an Intensive Care Unit (ICU), skilled in interpreting complex longitudinal patient data, including time-series vitals and labs, as well as unstructured clinical notes to predict clinical outcomes.

I will provide you with longitudinal medical information for a patient. The data covers 3 visits that occurred at 2113-01-31, 2113-02-01, 2113-02-02.
Each clinical feature is presented as a list of values, corresponding to these visits. Missing values are represented as `NaN` for numerical values and "unknown" for categorical values. Note that units and reference ranges are provided alongside relevant features.

Patient Background:
- Sex: male
- Age: 50 years

Your Task:
Your primary task is to analyze the provided medical data, including both the time-series clinical features and the unstructured clinical note, to predict the probability of readmission within 30 days post-discharge. Include cases where a patient passes away within 30 days from the discharge date as readmissions.

Instructions & Output Format:
Please first perform a step-by-step analysis. In your reasoning, you must integrate insights from both the structured time-series data (considering trends and abnormal values at discharge) and the unstructured clinical note (considering comorbidities, social situation, and discharge plan). Then, provide a final assessment of the likelihood of readmission within 30 days post-discharge.

Your final output must be a JSON object containing two keys:
1.  `"think"`: A string containing your detailed step-by-step clinical reasoning (under 500 words).
2.  `"answer"`: A floating-point number between 0 and 1 representing the predicted probability of readmission (higher value means higher likelihood of readmission).

Example Format: ```json { "think": "The patient's vital signs and labs stabilized before discharge. However, the admission summary notes a complex medication regimen and limited family support. This combination of clinical stability but social vulnerability suggests a moderate risk of readmission due to potential non-adherence or lack of care at home.", "answer": 0.40 }```

Handling Uncertainty:
In situations where the provided data is clearly insufficient or too ambiguous to make a reasonable prediction, respond with the exact phrase: `I do not know`

Now, please analyze and predict for the following patient:

Clinical Note: sex f service neurosurgery allergies prednisone latex lactose mirapex doxycycline attending . chief complaint intraventricular hemorrhage major surgical or invasive procedure r evd placed angio dural av fistula angio for embolization post meningeal branch replacement of r evd embolization of dural av fistula peg insertion history of present illness this is a year old female found down by her roommate in the bathroom on the ground with emesis. roommate stated that the patient may have taken some percocet. ems called and patient taken to . upon arrival patient was obtunded, received narcan with no improvement and was intubated for airway protection and underwent a head ct which revealed a hemorrhage......

Clinical Features Over Time:
- Capillary refill rate (Unit: /. Reference range: /.): [0.0, 0.0, 0.0]
- Glascow coma scale eye opening (Unit: /. Reference range: /.): [Spontaneously, Spontaneously, To Speech]
- Glascow coma scale motor response (Unit: /. Reference range: /.): [Obeys Commands, Obeys Commands, Obeys Commands]
- Glascow coma scale total (Unit: /. Reference range: /.): [0.0, 0.0, 0.0]
- Glascow coma scale verbal response (Unit: /. Reference range: /.): [Oriented, Oriented, No Response]
- Diastolic blood pressure (Unit: mmHg. Reference range: less than 80.): [55.0, 74.0, 73.0]
- Fraction inspired oxygen (Unit: /. Reference range: more than 21.): [80.0, 50.0, 70.0]
- Glucose (Unit: mg/dL. Reference range: 70 - 100.): [119.0, 118.0, 127.0]
- Heart Rate (Unit: bpm. Reference range: 60 - 100.): [86.0, 110.0, 118.0]
- Height (Unit: cm. Reference range: /.): [157.0, NaN, NaN]
- Mean blood pressure (Unit: mmHg. Reference range: less than 100.): [67.0, 85.0, 102.0]
- Oxygen saturation (Unit: %. Reference range: 95 - 100.): [96.0, 100.0, 100.0]
- Respiratory rate (Unit: breaths per minute. Reference range: 15 - 18.): [17.0, 26.0, 15.0]
- Systolic blood pressure (Unit: mmHg. Reference range: less than 120.): [105.0, 124.0, 156.0]
- Temperature (Unit: degrees Celsius. Reference range: 36.1 - 37.2.): [37.89, 37.61, 37.17]
- Weight (Unit: kg. Reference range: /.): [80.92, 80.92, 80.92]
- pH (Unit: /. Reference range: 7.35 - 7.45.): [7.48, 7.47, 7.51]
\end{VerbatimWrap}
\end{tcolorbox}
\end{center}

\end{document}